\documentclass[10pt,twocolumn,letterpaper]{article}

\usepackage{cvpr}              

\usepackage{graphicx}
\usepackage{amsmath}
\usepackage{amssymb}
\usepackage{booktabs}

\usepackage[utf8]{inputenc} 
\usepackage[T1]{fontenc}    
\usepackage{url}            
\usepackage{amsfonts}       
\usepackage{nicefrac}       
\usepackage{microtype}      
\usepackage{xcolor}         
\usepackage{multirow}
\usepackage{makecell}
\usepackage{caption}
\usepackage{subcaption}
\makeatletter
\@namedef{ver@everyshi.sty}{}
\makeatother
\usepackage{tikz}
\usepackage{bbm}
\usepackage{wrapfig}
\usepackage{mathtools,xparse}
\usepackage{pifont}
\usepackage{algorithmic}
\usepackage[ruled,norelsize,linesnumbered]{algorithm2e}
\usepackage{enumitem}
\usepackage{pifont}
\newcommand{\cmark}{\ding{51}}%
\newcommand{\xmark}{\ding{55}}%
\usepackage{array}

\usepackage[pagebackref,breaklinks,colorlinks]{hyperref}

\usepackage[capitalize]{cleveref}
\crefname{section}{Sec.}{Secs.}
\Crefname{section}{Section}{Sections}
\Crefname{table}{Table}{Tables}
\crefname{table}{Tab.}{Tabs.}

\def\cvprPaperID{11433} 
\def\confName{CVPR}
\def\confYear{2023}

\begin{document}

\title{Coreset Sampling from Open-Set for Fine-Grained Self-Supervised Learning}

\author{
Sungnyun Kim\thanks{equal contribution} \qquad Sangmin Bae\footnotemark[1] \qquad Se-Young Yun \vspace{3.5pt}\\
KAIST AI \\
{\tt\small \{ksn4397, bsmn0223, yunseyoung\}@kaist.ac.kr} 
}
\maketitle

\begin{abstract}

Deep learning in general domains has constantly been extended to domain-specific tasks requiring the recognition of fine-grained characteristics. However, real-world applications for fine-grained tasks suffer from two challenges: a high reliance on expert knowledge for annotation and necessity of a versatile model for various downstream tasks in a specific domain (e.g., prediction of categories, bounding boxes, or pixel-wise annotations). Fortunately, the recent self-supervised learning\,(SSL) is a promising approach to pretrain a model without annotations, serving as an effective initialization for any downstream tasks. Since SSL does not rely on the presence of annotation, in general, it utilizes the large-scale unlabeled dataset, referred to as an open-set. In this sense, we introduce a novel Open-Set Self-Supervised Learning problem under the assumption that a large-scale unlabeled open-set is available, as well as the fine-grained target dataset, during a pretraining phase. In our problem setup, it is crucial to consider the distribution mismatch between the open-set and target dataset. Hence, we propose SimCore algorithm to sample a coreset, the subset of an open-set that has a minimum distance to the target dataset in the latent space. We demonstrate that SimCore significantly improves representation learning performance through extensive experimental settings, including eleven fine-grained datasets and seven open-sets in various downstream tasks.
\let\thefootnote\relax\footnotetext{\!\!\url{https://github.com/sungnyun/openssl-simcore}}

\end{abstract}

\vspace{-3pt}
\section{Introduction}\label{sec:introduction}
\vspace{-2pt}

The success of deep learning in general computer vision tasks has encouraged its widespread applications to specific domains of industry and research\cite{el2021large, wei2021fine, luo2019cross}, such as facial recognition or vehicle identification.
We particularly focus on the visual recognition of fine-grained datasets, where the goal is to differentiate between hard-to-distinguish images.
However, real-world application for fine-grained tasks poses two challenges for practitioners and researchers developing algorithms.
First, it requires a number of experts for annotation, which incurs a large cost\cite{badge, chen2020simple, sohn2020fixmatch}.
For example, ordinary people do not have professional knowledge about aircraft types or fine-grained categories of birds. Therefore, a realistic presumption for a domain-specific fine-grained dataset is that there may be no or very few labeled samples.
Second, fine-grained datasets are often re-purposed or used for various tasks according to the user's demand, which motivates development of a versatile pretrained model.
%
One might ask, as a target task, that bird images be classified by species or even segmented into foreground and background.
A good initialization model can handle a variety of annotations for fine-grained datasets, such as multiple attributes\cite{birds, aircraft, liu2015faceattributes}, pixel-level annotations\cite{birds, pet}, or bounding boxes\cite{pet, aircraft, cars}.

\begin{figure*}[!t]
\centering
\vspace{-5pt}
\includegraphics[width=0.97\textwidth]{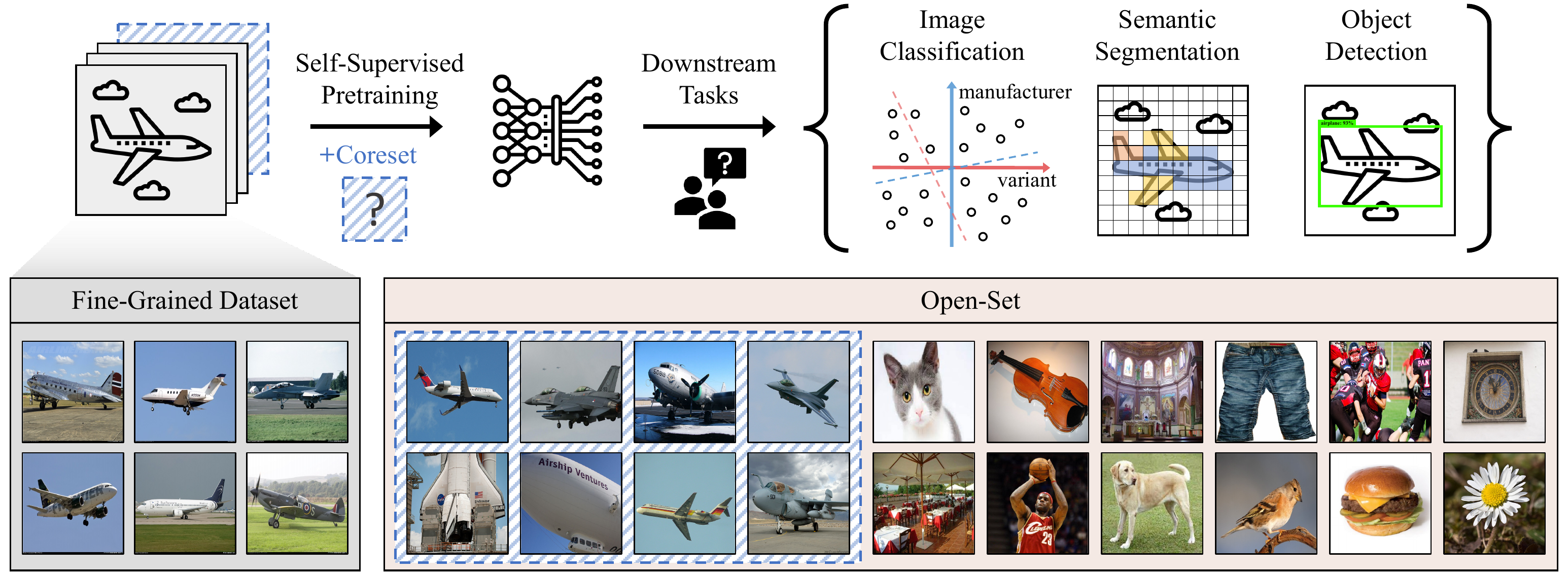}
\vspace{-3pt}
\caption{
Overview of an OpenSSL problem. For any downstream tasks, we pretrain an effective model with the fine-grained dataset via self-supervised learning (SSL). Here, the assumption for a large-scale unlabeled open-set in the pretraining phase is well-suited for a real-world scenario. The main goal is to find a coreset, highlighted by the blue box, among the open-set to enhance fine-grained SSL.}
\label{fig:concept}
\vspace{-2pt}
\end{figure*}

\begin{figure*}[!t]
\begin{minipage}[!t]{0.70\textwidth}
    \centering
    \includegraphics[width=\textwidth]{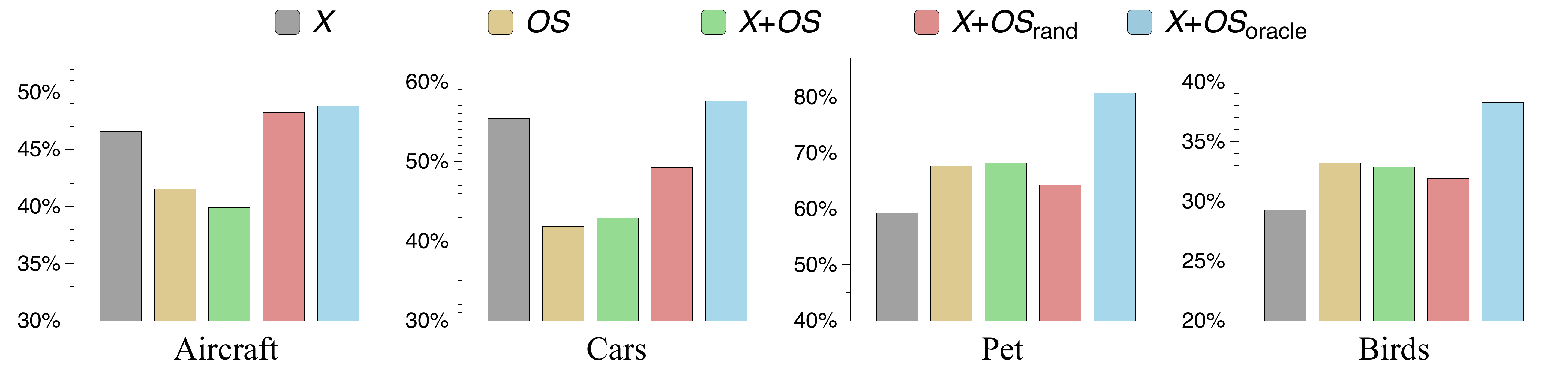}
    \label{fig:motivation_a}
\end{minipage}
\hfill
\begin{minipage}[!t]{0.29\textwidth}
    \vspace{-9pt}
    \small
    \raggedright
    \addtolength{\tabcolsep}{-2.5pt}
    \resizebox{\linewidth}{!}{
    \renewcommand{\arraystretch}{1.09}
    \begin{tabular}{|l|l|}
    \hline
    Target\,($X$) & Classes for $\textit{OS}_\text{oracle}$ (\#)  \\
    \hline
    Aircraft & airliner, warplane, ... (8) \\
    \hline
    Cars  & convertible, jeep, ... (10) \\
    \hline
    Pet & Persian cat, beagle, ... (24) \\
    \hline
    Birds & goldfinch, junco, ... (20) \\
    \hline
    \end{tabular}}
    \label{fig:motivation_b}
\end{minipage}
\vspace{-18pt}
\caption{Linear evaluation performance on the fine-grained target dataset. Each color corresponds to a pretraining dataset, while ``+'' means merging two datasets. The table on the right side shows the manually selected categories from an open-set\,(\textit{OS}), ImageNet-1k\cite{deng2009imagenet} in this case, according to each target dataset\,($X$).
Selected categories and exact numbers are detailed in Appendix\,\ref{appx:details_motivating_experiments}.
We followed the typical linear evaluation protocol\cite{chen2020simple, grill2020bootstrap} and used the SimCLR method\cite{chen2020simple} on ResNet50 encoder\cite{he2016deep}.}
\label{fig:motivation}
\vspace{-5pt}
\end{figure*}

Recently, self-supervised learning\,(SSL) \cite{chen2020simple, he2020momentum, grill2020bootstrap, caron2021emerging} has enabled learning how to represent the data even without annotations, such that the representations serve as an effective initialization for any future downstream tasks. Since labeling is not necessary, SSL generally utilizes an \textit{open-set}, or large-scale unlabeled dataset, which can be easily obtained by web crawling\cite{goyal2021self, tian2021divide}, for the pretraining.
In this paper, we introduce a novel Open-Set Self-Supervised Learning\,(OpenSSL) problem, where we can leverage the open-set as well as the training set of fine-grained target dataset. Refer to Figure\,\ref{fig:concept} for the overview of OpenSSL.

In the OpenSSL setup, since the open-set may contain instances irrelevant to the fine-grained dataset, we should consider the distribution mismatch. A large distribution mismatch might inhibit representation learning for the target task. 
For instance, in Figure\,\ref{fig:motivation}, SSL on the open-set\,(\textit{OS}) does not always outperform SSL on the fine-grained dataset\,($X$) because it depends on the semantic similarity between $X$ and \textit{OS}. This is in line with the previous observations\cite{ericsson2021well, el2021large, tian2021divide} that the performance of self-supervised representation on downstream tasks is correlated with similarity of pretraining and fine-tuning datasets.

To alleviate this mismatch issue, we could exploit a \textit{coreset}, a subset of an open-set, which shares similar semantics with the target dataset. As a motivating experiment, we manually selected the relevant classes from ImageNet\,($\textit{OS}_\text{oracle}$) that are supposed to be helpful according to each target dataset.
Interestingly, in Figure\,\ref{fig:motivation}, merging $\textit{OS}_\text{oracle}$ to $X$ shows a significant performance gain, and its superiority over merging the entire open-set\,($X$+\textit{OS}) or the randomly sampled subset\,($X$+$\textit{OS}_\text{rand}$) implies the necessity of a sampling algorithm for the coreset in the OpenSSL problem.

Therefore, we propose \textbf{SimCore}, a simple yet effective coreset sampling algorithm from an unlabeled open-set. 
Our main goal is to find a subset semantically similar to the target dataset.
We formulate the data subset selection problem to obtain a coreset that has a minimum distance to the target dataset in the latent space.
SimCore significantly improves performance in extensive experimental settings (eleven fine-grained datasets and seven open-sets), and shows consistent gains with different architectures, SSL losses, and downstream tasks. 
Our contributions are outlined as follows:
\vspace{-5pt}
\begin{itemize}[leftmargin=10pt]
\setlength\itemsep{-0.2em}
    \item We first propose a realistic OpenSSL task, assuming an unlabeled open-set available during the pretraining phase on the fine-grained dataset.
    \item We propose a coreset selection algorithm, SimCore, to leverage a subset semantically similar to the target dataset.
    \item Our extensive experiments with eleven fine-grained datasets and seven open-sets substantiate the significance of data selection in our OpenSSL problem.
\end{itemize}
\section{Related Works}\label{sec:related_works}

\begin{table*}[!t]
    \small
    \centering
    \addtolength{\tabcolsep}{-0.7pt}
    \resizebox{1.0\textwidth}{!}{
    \begin{tabular}{l|l|c|c|c|c|l}
        \toprule
         &  & Train & Train & & & \\
         \multirow{-2}{*}{Task [ref.]} & \multirow{-2}{*}{Problem Setting} & \!\!\!(Labeled)\!\!\! & \!\!\!(Unlabeled)\!\!\! & \multirow{-2}{*}{Test} & \multirow{-2}{*}{\!\!\!\!\!\!\!\!\!\!\!\!Definition of \texttt{OS}\,/\,\texttt{CS} \hfill} & \multirow{-2}{*}{Main Goal}  \\
         \midrule \midrule
         \thead[l]{Novel Class Discovery\\\cite{hsu2017learning, han2020automatically, zhong2021openmix}} & \thead[l]{test data consist of\\only novel classes} & \thead{seen} & - & \thead{novel} & - & \thead[l]{cluster novel classes\\in test dataset} \\
         \hline
         \thead[l]{Open-Set Recognition\\\cite{scheirer2012toward, bendale2016towards, chen2021adversarial, vaze2021open}} & \thead[l]{test set contains seen\\and novel classes} & \thead{seen} & - & \thead{\!seen\,+\!\\novel} & \thead[l]{[\texttt{OS}] test dataset containing\\seen and novel classes} & \thead[l]{reject instances from\\novel classes at test time} \\
         \hline
         \thead[l]{Webly Sup.\\\cite{chen2015webly, li2020mopro, sun2021webly}} & \thead[l]{train data contains web-\\crawled noisy samples} & \thead{partially\\noisy} & - & \thead{seen} & \thead[l]{[\texttt{OS}] web-crawled train dataset\\containing noisy samples} & \thead[l]{robustly train instances\\with corrupted labels} \\
         \hline
         \thead[l]{Open-Set Semi-Sup. \\\cite{saito2021openmatch, oliver2018realistic, chen2020semi, killamsetty2021retrieve, su2021realistic}} & \thead[l]{unlabeled train data\\contain novel classes} & \thead{seen} & \thead{seen\,+\\novel} & \thead{seen} & \thead[l]{[\texttt{OS}] training dataset contain-\\ing seen and novel classes} & \thead[l]{train a robust model while\\regularizing novel classes} \\
         \hline
         \thead[l]{Open-World Semi-Sup. \\\cite{cao2021open, boult2019learning, bendale2015towards}} & \thead[l]{train and test data\\contain novel classes} & \thead{seen} & \thead{seen\,+\\novel} & \thead{\!seen\,+\!\\novel} & \thead[l]{[\texttt{OS}] dataset containing\\seen and novel classes}& \thead[l]{discover novel classes and\\assign samples at test time} \\
         \hline
         \thead[l]{Open-Set Annotation\\\cite{ning2022active}} & \thead[l]{unlabeled data pool\\contains novel classes} & \thead{seen} & \thead{seen*\,+\\novel*} & \thead{seen} & \thead[l]{[\texttt{OS}] unlabeled data pool\\with seen and novel classes} & \thead[l]{aim to query seen classes\\from unlabeled data pool} \\
         \hline
         \thead[l]{Coreset Selection\\in AL\cite{coreset, wei2015submodularity}} & \thead[l]{query instances to be\\annotated 
         } & \thead{seen} & \thead{seen*} & \thead{seen} & \thead[l]{[\texttt{CS}] the most representative\\subset of unlabeled set} & \thead[l]{find a small subset\\competitive to whole dataset} \\
         \hline
         \thead[l]{Coreset Selection\\in CL\cite{aljundi2019gradient, yoon2021online, tiwari2022gcr}} & \thead[l]{continuously learn\\a sequence of tasks} & \thead{\!\!partially\!\!\\novel} & - & \thead{seen} & \thead[l]{[\texttt{CS}] the most representative\\instances at each task} & \thead[l]{promote task adaptation with\\less catastrophic forgetting} \\
         \hline
         \thead[l]{Hard Negative Mining\\in Self-Sup.\cite{robinson2020contrastive, wang2021understanding}} & \thead[l]{assume that hard\\negatives are helpful} & - & \thead{target} & \thead{target} & \thead[l]{[\texttt{CS}] the hardest contrastive\\pair instances for SSL} & \thead[l]{improve SSL performance \\using core-negative instances} \\
         \hline
         \thead[l]{\bf Open-Set Self-Sup.\\\bf [ours]} & \thead[l]{utilize open-set in \\pretraining, which may\!\\have irrelevant data} & - & \thead{target\,+\\\!irrelevant\!} & \thead{target} & \thead[l]{[\texttt{OS}] large-scale unlabeled set\\ \!\,[\texttt{CS}] subset of \texttt{OS} sharing the\\same semantics with target set} & \thead[l]{improve SSL performance\\on fine-grained datset via\\coreset sampling method} \\
        \bottomrule
        \end{tabular}}
    \caption{Comparisons of the OpenSSL problem with relevant literature. We focus on the problem setting and the definition of open-set\,(\texttt{OS}) or coreset\,(\texttt{CS}) in each field. AL and CL refer to active learning and continual learning, respectively, and * indicates the instances in the unlabeled data pool that are supposed to be annotated after the active selection.}
    \label{tab:related_works}
\end{table*}

\subsection{Self-Supervised Learning}

After Oord \etal\cite{oord2018representation} proposed an InfoNCE loss, contrastive learning algorithms began to show remarkable improvements in representation learning\cite{chen2020simple, he2020momentum, grill2020bootstrap, caron2020unsupervised, chen2021exploring, li2021efficient, caron2021emerging}. 
While a large-scale open-set enhances the generalization of learned representation \cite{jaiswal2020survey, tian2021divide, cole2022does}, recent literature has pointed out the distribution mismatch between pretraining and fine-tuning datasets \cite{el2021large, tian2021divide, goyal2021self}.
Particularly, El \etal\cite{el2021large} claimed that pretraining on ImageNet may not always be effective on the target task from different domains.
Tian \etal\cite{tian2021divide} found that pretraining with uncurated data, a more realistic scenario, deteriorates the target task performance.
Although the motivations coincide with ours, their proposed methods are focused on a single data scheme: a denoising autoencoder\cite{el2021large} only for fine-grained dataset, and distillation from expert models\cite{tian2021divide} or a novel architecture\cite{goyal2021self} for uncurated open-sets.
In contrast, we propose an explicit sampling strategy from an open-set, which becomes more effective by augmenting fine-grained dataset with well-matched samples, as well as achieving robustness to the distribution discrepancy or curation of the open-set.

\subsection{Coreset Selection from Open-Set}
In an OpenSSL problem, we denote an open-set as the additional unlabeled pretraining set that includes instances either from relevant or irrelevant domain to target dataset. The assumption of available open-set is also common in other research fields, such as open-set recognition\cite{scheirer2012toward, bendale2016towards, chen2021adversarial, vaze2021open}, webly supervised learning\cite{chen2015webly, li2020mopro, sun2021webly}, open-set\cite{oliver2018realistic, chen2020semi, saito2021openmatch,killamsetty2021retrieve} or open-world\cite{bendale2015towards, cao2021open, boult2019learning} semi-supervised learning, and open-set annotation\cite{ning2022active}, although its detailed meaning varies in each field. We summarize the details in Table\,\ref{tab:related_works}.
Especially, our OpenSSL task is to recognize coreset from a large-scale open-set, without exploiting any label information.
From this perspective, recent coreset selection approaches give us a good intuition.
Existing studies find the representative subset of the unlabeled set for active selection\cite{wei2015submodularity, coreset}, or find the subset of current task data to avoid catastrophic forgetting for continual learning\cite{aljundi2019gradient, yoon2021online, tiwari2022gcr}.
In the meantime, several works on self-supervised learning have developed a novel loss function that leverages hard negative samples, \ie, hard negative mining\cite{robinson2020contrastive, wang2021understanding}, which shares similar concepts to our coreset.
Our problem setup further requires an effective algorithm that takes into consideration of the distribution discrepancy in the open-set.
\section{Method}\label{sec:method}

\DeclarePairedDelimiter{\norm}{\lVert}{\rVert}

\subsection{OpenSSL Problem Formulation}
\label{subsec:openssl_problem_formulation}

SSL is a groundbreaking paradigm for learning inherent properties of data, while discarding irrelevant signals by discriminating perturbed samples.
Recent literature has proposed a contrastive loss function to encourage making representations from the same image similar and representations from different images dissimilar\cite{chen2020simple, he2020momentum, zbontar2021barlow}.

Given an input data $X=\{x_i\}_{i=1}^N$, we generate two copies of random augmented image, $\mathcal{A}(X) = \{\tilde{x}_i\}_{i=1}^{2N}$, where $\tilde{x}_i$ and $\tilde{x}_{N+i}$ are an augmented pair of each other.
Let $E_{\theta}$ be an encoder network. Then, with the augmented pairs, we generally formulate a contrastive loss as follows:
\vspace{-2pt}
\begin{equation}
    \mathcal{L}(X; E_{\theta}) = \frac{1}{2|X|} \sum_{\tilde{x}_i \in \mathcal{A}(X)} \ell_{\textsf{ssl}}(z_i, z_i^{+}; \{ z_i^{-}\})
\label{eq:contrastive}
\vspace{-3pt}
\end{equation}
where $z_i = E_{\theta}(\tilde{x}_i)$, $z_i^{+}$ denotes the representation from the augmented pair of $\tilde{x}_i$, and $\{z_i^{-}\}$ is the set of features from all the other samples.
Eq.\,\ref{eq:contrastive} forces $z_i$ to be closer with $z_i^{+}$ and farther from $\{z_i^{-}\}$.
Projection heads\cite{chen2020simple} or predictors\cite{grill2020bootstrap} are often used in measuring the similarities of representations.
In addition, non-contrastive methods \cite{grill2020bootstrap, caron2020unsupervised, caron2021emerging, chen2021exploring} have also been proposed by not utilizing any negative pair set, \ie, $\{ z_i^{-} \} = \emptyset$.

In the OpenSSL problem, $X$ corresponds to the training set of the fine-grained target dataset. Using Eq.\,\ref{eq:contrastive}, we can pretrain an encoder without any annotation of $X$.
Furthermore, we have an unlabeled open-set $\mathcal{U}$, which can be jointly used with $X$.
Rather than pretraining on a simple fusion of $X$ and $\mathcal{U}$, we are motivated to sample a relevant subset $\mathcal{S}$ from the open-set. 
We then pretrain the encoder with $\mathcal{L}(X \cup \mathcal{S};E_{\theta})$.

\subsection{Simple Coreset Sampling from Open-Set}
\label{subsec:simcore}
We introduce a simple coreset sampling algorithm, coined as \textbf{SimCore}.
Motivated by Figure\,\ref{fig:motivation}, selecting an appropriate coreset is a key for the OpenSSL problem. 
To this end, we build a set with the open-set samples that are the nearest neighbors of the target samples.

This is formulated by finding a subset $\mathcal{S}$ that maximizes the following objective function:
\vspace{-3pt}
\begin{equation}\label{eq:submodular}
    f(\mathcal{S}) = \sum_{x \in X} \max_{u \in \mathcal{S}} w(x, u), \text{\,where\,\,} \mathcal{S} \subseteq \mathcal{U},\,\, \mathcal{U} \cap X = \emptyset
\vspace{-4pt}
\end{equation}
while $w(x,u) = z_x^\top z_u$ estimates similarity of two representations, and $z$ is the normalized feature from the encoder $E_\theta$ pretrained on $X$ with small epochs.
From Eq.\,\ref{eq:submodular}, SimCore finds a subset that shares the most similar semantics with the target set.
This is reminiscent of the facility location function\cite{mirchandani1990discrete, wei2015submodularity}, $f_{\text{fac}}(\mathcal{S})=\sum_{x \in \mathcal{U}} \max_{u \in \mathcal{S}} w(x, u),\mathcal{S}\subseteq \mathcal{U}$. However, $f(\mathcal{S})$ is different from $f_{\text{fac}}(\mathcal{S})$, since target samples are the elements of $X$, which is disjoint with $\mathcal{S}$.

Meanwhile, the direct calculation of pairwise similarities requires the complexity of $\mathcal{O}(|X||\,\mathcal{U}|)$, which might be extremely large. To reduce the computational overhead and make the algorithm scalable, we adapt $k$-means clustering\cite{kmeans} for the target dataset $X$, and replace it with the centroid set $\hat{X}$. Its associated objective function is $\hat{f}(\mathcal{S})$. Even when exploiting only the centroids\,($k=100$ in practice), we show significant performance gains on various benchmarks.

\vspace{-10pt}
\paragraph{Iterative coreset sampling:}
$\hat{f}(\mathcal{S})$ is a monotonically increasing submodular function. One remark is that if there is no constrained budget on $\mathcal{S}$, 
a subset achieving the maximum value of $\hat{f}(\mathcal{S})$ is not unique.
If we denote $\mathcal{S}^*$ as a minimal set with the maximum value, $\mathcal{S}^*$ is obtained when including only the instances closest to each instance of $\hat{X}$ (\ie, $|\mathcal{S}^*|\leq|\hat{X}|$).
Since we want to sample sufficiently large subset, we re-define Eq.\,\ref{eq:submodular} with the selection round $t$: $\hat{f}(\mathcal{S}_t)$ where the candidate set is $\mathcal{U}_t$.
Thus, we iterate the rounds to repeat sampling $\mathcal{S}_t^*$ and excluding them from the candidate set, until we reach the proper budget size (see Algorithm\,\ref{alg:simcore}). We collect all the coreset samples into a set $\mathcal{I}=\mathcal{S}^*_1 \cup \mathcal{S}^*_2 \cup\cdots\cup \mathcal{S}^*_T$ to merge with $X$.

\vspace{-10pt}
\paragraph{Stopping criterion:}
However, we have no knowledge in practice if every sampled instance within the budget is sufficiently close to the target set. This necessitates stopping criterion that blocks the sampling process from continuing when the samples are no longer close to the target set.
To this end, we calculate the ratio  $\hat{f}(\mathcal{S}_t^*)/\hat{f}(\mathcal{S}_1^*)$ at each iteration and stop the sampling process if its value is less than the threshold. This implies that we stop at iteration $t$ if the sampled subset is not as similar as the first subset is to the target set. We filtered-out by using the threshold $\tau=0.95$ throughout experiments.

\setlength{\textfloatsep}{10pt}
\begin{algorithm}[!t]
\caption{Simple coreset sampling from open-set\!\!\!\!}
 \label{alg:simcore}
\SetAlgoLined
{\bf Require:} $E_{\theta}$: encoder pretrained on $X$\;
{\bf Require:} $\mathcal{U}_0$: initial candidate set (open-set)\;
{\bf Require:} $\mathcal{B}$, $\tau$: coreset budget, threshold\;
initialize $\mathcal{I} \leftarrow \emptyset$, $t\leftarrow 0$\;
replace $\hat{X} \leftarrow$ cluster centroids of $X$\;
calculate $z_x, z_u \leftarrow E_{\theta}(x), E_{\theta}(u)$ for $\forall x,u \!\in\! \hat{X}\!\times\mathcal{U}_0$\;
\While{$|\mathcal{I}|<\mathcal{B}$}{
    set $\mathcal{S}_t^*$ as the elements in $\mathcal{U}_t$ that are closest to each element in $\hat{X}$ (Eq.\,\ref{eq:submodular})\;
    $\mathcal{I} \leftarrow \mathcal{I} \cup \mathcal{S}_t^*$,\,\, 
    $\mathcal{U}_{t+1} \leftarrow \mathcal{U}_t \setminus \mathcal{S}_t^*$ \\
    $t \leftarrow t+1$ \\
    \textcolor{gray}{\textit{//\,stopping criterion}}\\
        \If{${\hat{f}(\mathcal{S}_t^*)} < \tau \cdot {\hat{f}(\mathcal{S}_1^*)}$}
        {
            \textit{stop sampling}\;
        }
 }
re-initialize $\theta$ and pretrain $E_{\theta}$ with $X \cup \mathcal{I}$;
\end{algorithm}

\section{Experiments}\label{sec:experiments}

We evaluate the learned representation quality according to each pretraining dataset. We thus demonstrate, through various fine-grained target tasks, that SimCore samples an effective subset that enhances pretraining.
Here, we used eleven target datasets: Aircraft\cite{aircraft}, Cars\cite{cars}, Pet\cite{pet}, Birds\cite{birds}, Dogs\cite{khosla2011novel}, Flowers\cite{nilsback2008automated}, Action\cite{yao2011human}, Indoor\cite{quattoni2009recognizing}, Textures\cite{cimpoi2014describing}, Faces\cite{lee2020maskgan}, and Food\cite{singla2016food}. For the open-set, we mainly used ImageNet-1k\cite{deng2009imagenet}, while we extended to other open-sets in Section\,\ref{subsec:simcore_on_various_opensets}. By default, we used SimCLR\cite{chen2020simple} with ResNet50 architecture as the encoder.
The detailed experimental setups and dataset configurations are in Appendix\,\ref{appx:experimental_settings}.

\subsection{Performance Evaluation on Target Tasks}

\begin{table*}[!t]
    \small
    \centering
    \resizebox{\textwidth}{!}{
    \begin{tabular}{lcccccccccccc}
    \toprule
     & & \multicolumn{10}{c}{Target dataset ($X$) and its number of samples} \\
    \cmidrule(l{2pt}r{2pt}){3-13}
    & & \!\!Aircraft\!\! & Cars & Pet & Birds & Dogs & \!Flowers\!\! & Action & Indoor & \!\!Textures\! & Faces & Food \\
    pretrain & \!\!$p$ & 6,667 & 8,144 & 3,680 & 5,990 & 12,000 & 2,040 & 4,000 & 5,360 & 3,760 & 4,263 & 13,296 \\
    \midrule
    $X$ & \!\!- & 46.56 & 55.42 & 59.23 & 29.27 & 49.88 & 80.14 & 43.76 & 54.10 & 58.78 & 56.63 & 87.99 \\
    \textit{OS} & \!\!- & 41.50 & 41.86 & 67.66 & 33.21 & 49.94 & 85.67 & 60.65 & 64.46 & 67.23 & 52.84 & 86.14 \\
    $X$+\textit{OS} & \!\!- & 39.88 & 42.92 & 68.22 & 32.88 & 50.42 & 85.34 & 60.61 & 63.66 & 67.98 & 52.76 & 85.90 \\
    \hline
    $X$+$\textit{OS}_{\text{rand}}$ & \!\!1\% & 48.24 & 49.26 & 64.27 & 31.90 & 49.62 & 83.17 & 47.25 & 55.37 & 61.33 & 57.37 & 88.08 \\
    $X$+$\textit{OS}_{\text{SimCore}^\dagger}$ & \!\!1\% &  48.06 & 58.56 & 74.82 & 33.37 & 57.42 & 82.12 & 51.37 & 57.84 & 61.76 & 56.95 & 90.35 \\
    \bf $X$+$\textit{OS}_{\text{SimCore}}$ & \!\!1\% &  \bf 48.45 & \underline{59.00} & 77.13 & 36.56 & 59.83 & 86.70 & {52.98} & {59.18} & {63.40} & 58.85 & 89.78 \\
    \hline
    $X$+$\textit{OS}_{\text{rand}}$ & \!\!5\% &  45.75 & 46.03 & 68.38 & 33.63 & 50.24 & 84.52 & 57.27 & 60.71 & 65.80 & 56.05 & 87.75 \\
    $X$+$\textit{OS}_{\text{SimCore}^\dagger}$ & \!\!5\% &  45.57 & 50.75 & \underline{80.20} & 35.56 & 64.62 & 85.11 & 64.53 & 68.13 & 66.22 & 58.93 & 89.87 \\
    \bf $X$+$\textit{OS}_{\text{SimCore}}$ & \!\!5\% &  47.14 & 52.22 & \bf 81.75 & \bf 39.21 & \bf 66.82 & \bf 87.28 & \underline{66.38} & \underline{70.96} & \bf 68.13 &  \bf 59.34 & \underline{90.74} \\
    \hline 
     \multicolumn{2}{l}{\textcolor{gray}{\textit{\,Stopping Criterion}}} & \textcolor{gray}{$1.03\%$} & \textcolor{gray}{$0.95\%$} & \textcolor{gray}{$14.4\%$} & \textcolor{gray}{$13.7\%$} & \textcolor{gray}{$9.72\%$} & \textcolor{gray}{$7.96\%$} & \textcolor{gray}{$15.6\%$} & \textcolor{gray}{$13.5\%$} & \textcolor{gray}{$5.89\%$} & \textcolor{gray}{$0.27\%$} & \textcolor{gray}{$3.86\%$} \\
    \bf $X$+$\textit{OS}_{\text{SimCore}}$ & \!\!- &  \underline{48.27} & \textbf{60.29} & 79.66 & \underline{37.65} & \underline{66.48} & \underline{87.04} & \textbf{67.46} & \textbf{71.95} & \underline{67.66} & \underline{59.01} & \textbf{91.31} \\
    \bottomrule
    \end{tabular} }
    \vspace{-5pt}
    \caption{Linear evaluation performance on eleven fine-grained datasets. We used ImageNet-1k\cite{deng2009imagenet} as an open-set. The ratio $p$ is a fixed budget size to sample from the open-set, either via random sampling\,($\textit{OS}_{\text{rand}}$) or SimCore\,($\textit{OS}_{\text{SimCore}}$). Given that \textit{OS} has 1.3M samples, $p=1\%$ corresponds to 13K samples. $\dagger$ denotes SimCore with $k=1$, a single cluster. \textbf{Bold} and \underline{underline} indicate the best and the second best accuracy for each target dataset, respectively.}
    \label{tab:main_exp}
    \vspace{-10pt}
\end{table*}

\paragraph{Linear evaluation:}

Conventional SSL literature evaluates linear probing performance with frozen pretrained encoder to measure the representation quality\cite{chen2020simple, grill2020bootstrap}.
Table\,\ref{tab:main_exp} summarizes the linear evaluation results on eleven fine-grained datasets. We note two observations from this experiment.
First, we evaluated if the coreset sampled by SimCore is qualitative as pretraining data. To this end, we set the budget size to $p=1\%$ or $p=5\%$ (\ie, sampling $p$-ratio of the open-set) and compared them with $p\%$ random sampling strategy.
In every case, SimCore outperformed the random sampling. This demonstrates that exploiting the coreset is actually crucial compared to na\"ively using random samples.
Second, using a number of cluster centroids of the target dataset is more advantageous than a single cluster, although SimCore with $k=1$ also outperformed the random sampling.

\begin{table}[!t]
    \vspace{5pt}
    \small
    \centering
    \resizebox{\linewidth}{!}{
    \addtolength{\tabcolsep}{-1pt}    
    \renewcommand*{\arraystretch}{0.95}
    \begin{tabular}{lllcccc}
    \toprule
    method & architecture & pretrain &  \!\!Aircraft\!\! & Cars & Pet & Birds \\
    \midrule
    SimCLR & EfficientNet & $X$ & 25.5 & 37.0 & 58.1 & 27.8 \\
    SimCLR & EfficientNet & \textit{OS} & 31.6 & 29.5 & 57.8 &  26.5 \\
    SimCLR & EfficientNet & \bf SimCore & \bf 41.7 & \bf 52.8 & \bf 69.5 & \bf 29.6 \\
    \midrule
    SimCLR & ResNet18 & $X$ & 43.4 & 51.9 & 58.2 & 25.9 \\
    SimCLR & ResNet18 & \textit{OS} & 33.9 & 33.1 & 62.5 & 27.7 \\
    SimCLR & ResNet18 & \bf SimCore & \bf 44.5 & \bf 55.1 & \bf 72.7 & \bf 31.3 \\
    \midrule
    SimCLR & ResNeXt50 & $X$ & 45.9 & 56.5 & 63.4 & 28.6 \\
    SimCLR & ResNeXt50 & \textit{OS} & 39.2 & 39.4 & 68.2 & 32.6 \\
    SimCLR & ResNeXt50 & \bf SimCore & \bf 49.5 & \bf 59.5 & \bf 81.0 & \bf 37.4 \\
    \midrule
    SimCLR & ResNet101 & $X$ & 49.4 & 54.5 & 64.0 & 29.1 \\
    SimCLR & ResNet101 & \textit{OS} & 40.4 & 41.9 & 69.5 & 34.2 \\
    SimCLR & ResNet101 & \bf SimCore & \bf 50.9 & \bf 58.8 & \bf 83.0 & \bf 39.1 \\
    \bottomrule
    \end{tabular}}
    \vspace{-5pt}
    \caption{Linear evaluation performance with different architectures. SimCore corresponds to $X$+$\textit{OS}_{\text{SimCore}}$ with a stopping criterion.}
    \label{tab:different_arch}
\end{table}

Interestingly, the different trend across target datasets gives us a hint about the optimal coreset size based on the level of distribution mismatch to the open-set.
For example, in datasets like Pet and Birds, \textit{OS} pretraining was pretty effective, and in those datasets, SimCore took advantage of the large budget size.
This implies that several target datasets require more coreset samples than do others. 
However, in practice, we cannot pre-define the optimal budget size, since we do not have much knowledge about an uncurated open-set. 
Therefore, we should handle SimCore with a stopping criterion, as we already have proposed in Section\,\ref{subsec:simcore}. 

Surprisingly, SimCore with a stopping criterion highly improves the accuracy by +10.5\% (averaged over 11 datasets), compared to the $X$ pretraining. This is much larger gain compared to the large-scale \textit{OS} pretraining (+2.7\%) and 1\% random sampling (+1.3\%).
This is because SimCore adaptively samples a proper amount of coreset, and this amount differs by each target dataset.
For Aircraft and Cars, SimCore sampled around 1\% of ImageNet. This is a reasonable number, because in the ImageNet dataset\cite{deng2009imagenet}, there are actually 4 aircraft-related and 10 cars-related classes (refer to Appendix\,\ref{appx:details_motivating_experiments}), out of 1,000 classes in total.

\begin{table}[!t]
    \vspace{5pt}
    \small
    \centering
    \resizebox{\linewidth}{!}{
    \addtolength{\tabcolsep}{-1pt}
    \renewcommand*{\arraystretch}{0.924}
    \begin{tabular}{lllcccc}
    \toprule
    method & architecture & pretrain &  \!\!Aircraft\!\! & Cars & Pet & Birds \\
    \midrule
    BYOL & ResNet50 & $X$ & 40.6 & 49.4 & 56.5 & 27.6 \\
    BYOL & ResNet50 & \textit{OS} & 46.1 & 49.6 & 78.4 & 44.7 \\
    BYOL & ResNet50 & \bf SimCore & \bf 46.5 & \bf 50.4 & \bf 85.1 & \bf 47.9 \\
    \midrule
    SwAV & ResNet50 & $X$ & 34.5 & 42.4 & 49.4 & 21.6 \\
    SwAV & ResNet50 & \textit{OS} & 33.8 & 30.0 & 64.2 & 27.3 \\
    SwAV & ResNet50 & \bf SimCore & \bf 45.0 & \bf 45.1 & \bf 80.2 & \bf 36.6 \\
    \midrule
    DINO & ViT-Ti/16 & $X$ & 27.3 & \bf 48.2 & 42.4 & 28.5 \\
    DINO & ViT-Ti/16 & \textit{OS} & 42.0 & 39.1 & 78.4 & 61.2 \\
    DINO & ViT-Ti/16 & \bf SimCore & \bf 43.2 & 47.2 & \bf 83.3 & \bf 72.6 \\
    \midrule
    MAE & ViT-B/16 & $X$ & \bf 55.9 & 44.7 & 56.3 & 32.2 \\
    MAE & ViT-B/16 & \textit{OS} & 39.8 & 37.3 & 68.3 & 31.4 \\
    MAE & ViT-B/16 & \bf SimCore & 48.1 & \bf 52.4 & \bf 77.8 & \bf 42.1 \\
    \bottomrule
    \end{tabular}}
    \vspace{-5pt}
    \caption{Linear evaluation performance with different SSL methods. SimCore corresponds to $X$+$\textit{OS}_{\text{SimCore}}$ with a stopping criterion.}
    \label{tab:different_ssl}
\end{table}

\vspace{-12pt}
\paragraph{Different encoder architectures and SSL methods:} 
In Table\,\ref{tab:different_arch}, we have applied SimCore to different architectures: EfficientNet-B0\cite{tan2019efficientnet}, ResNet18\cite{he2016deep}, ResNeXt50\cite{xie2017aggregated}, and ResNet101\cite{he2016deep}.
Regardless of whether the encoder architecture is much smaller or larger, SimCore greatly improves pretraining on the target dataset.
Moreover, we have experimented with various SSL methods, such as BYOL\cite{grill2020bootstrap}, SwAV\cite{caron2020unsupervised}, DINO\cite{caron2021emerging}, and MAE\cite{he2022masked}, in Table\,\ref{tab:different_ssl}.
SimCore consistently demonstrates the effect of merging the coreset samples, even with the recent autoencoder-based SSL.

\begin{figure}[!t]
\begin{subfigure}[b]{\linewidth}
    \centering
    \includegraphics[width=\textwidth]{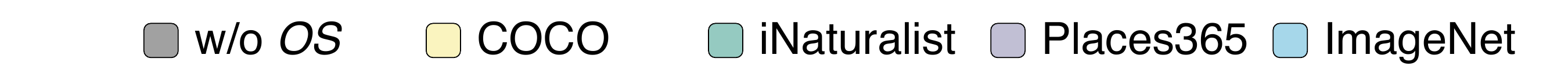}
\end{subfigure}
\hfill
\begin{subfigure}[b]{0.48\linewidth}
    \centering
    \includegraphics[width=\textwidth]{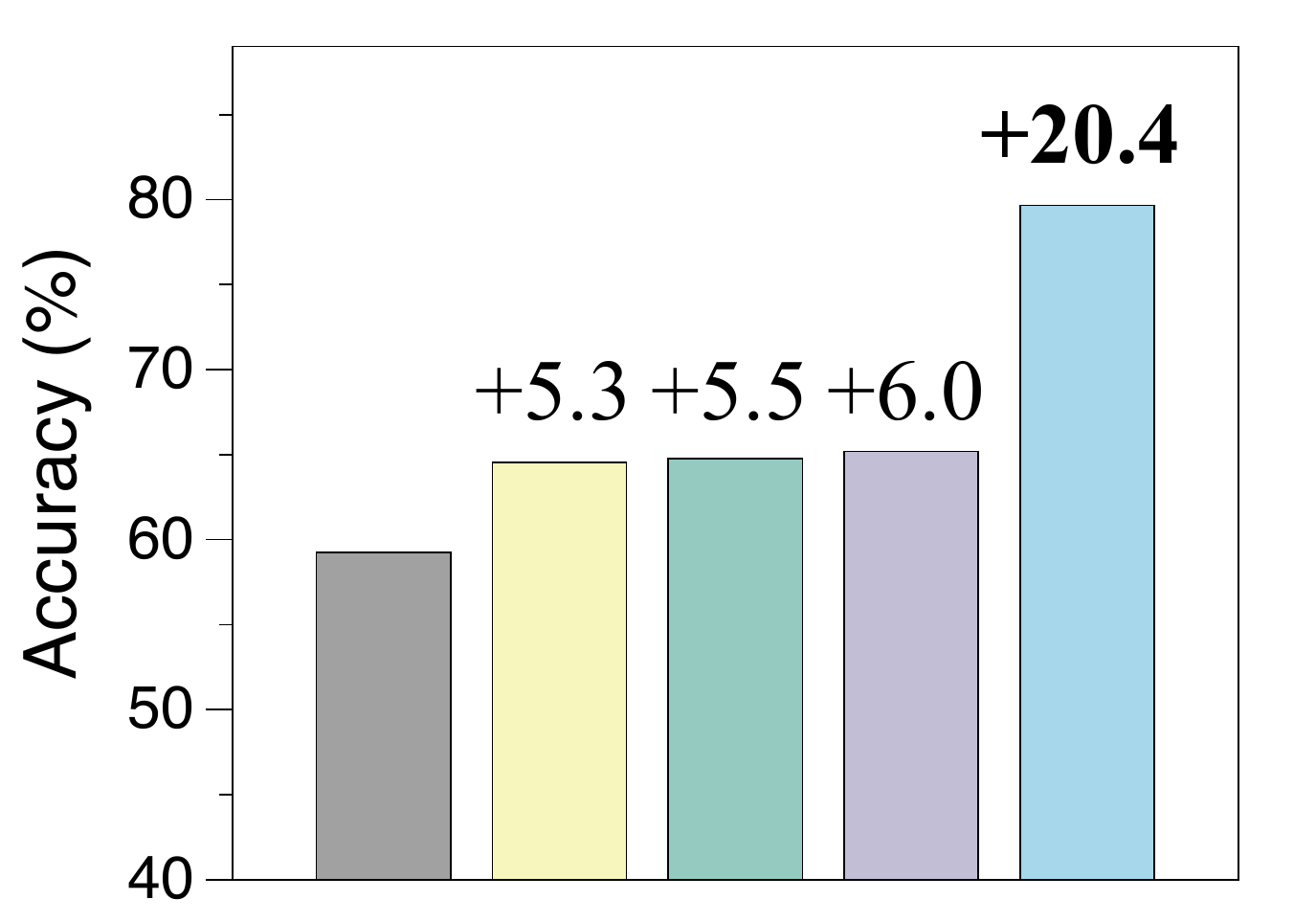}
    \vspace{-15pt}
    \caption{Pet}
\end{subfigure}
\hfill
\begin{subfigure}[b]{0.48\linewidth}
    \centering
    \includegraphics[width=\textwidth]{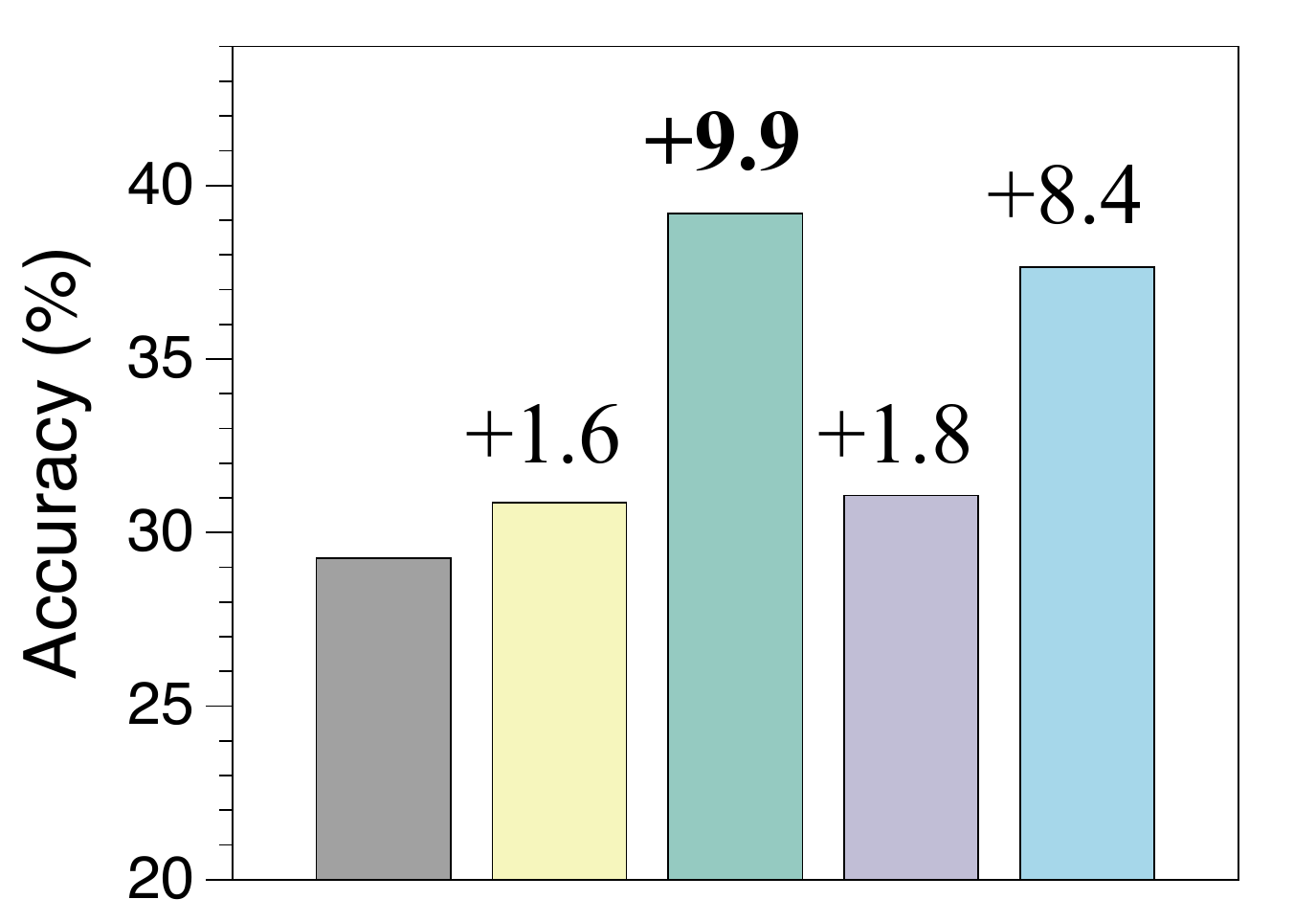}
    \vspace{-15pt}
    \caption{Birds}
\end{subfigure}
\hfill
\begin{subfigure}[b]{0.48\linewidth}
    \centering
    \includegraphics[width=\textwidth]{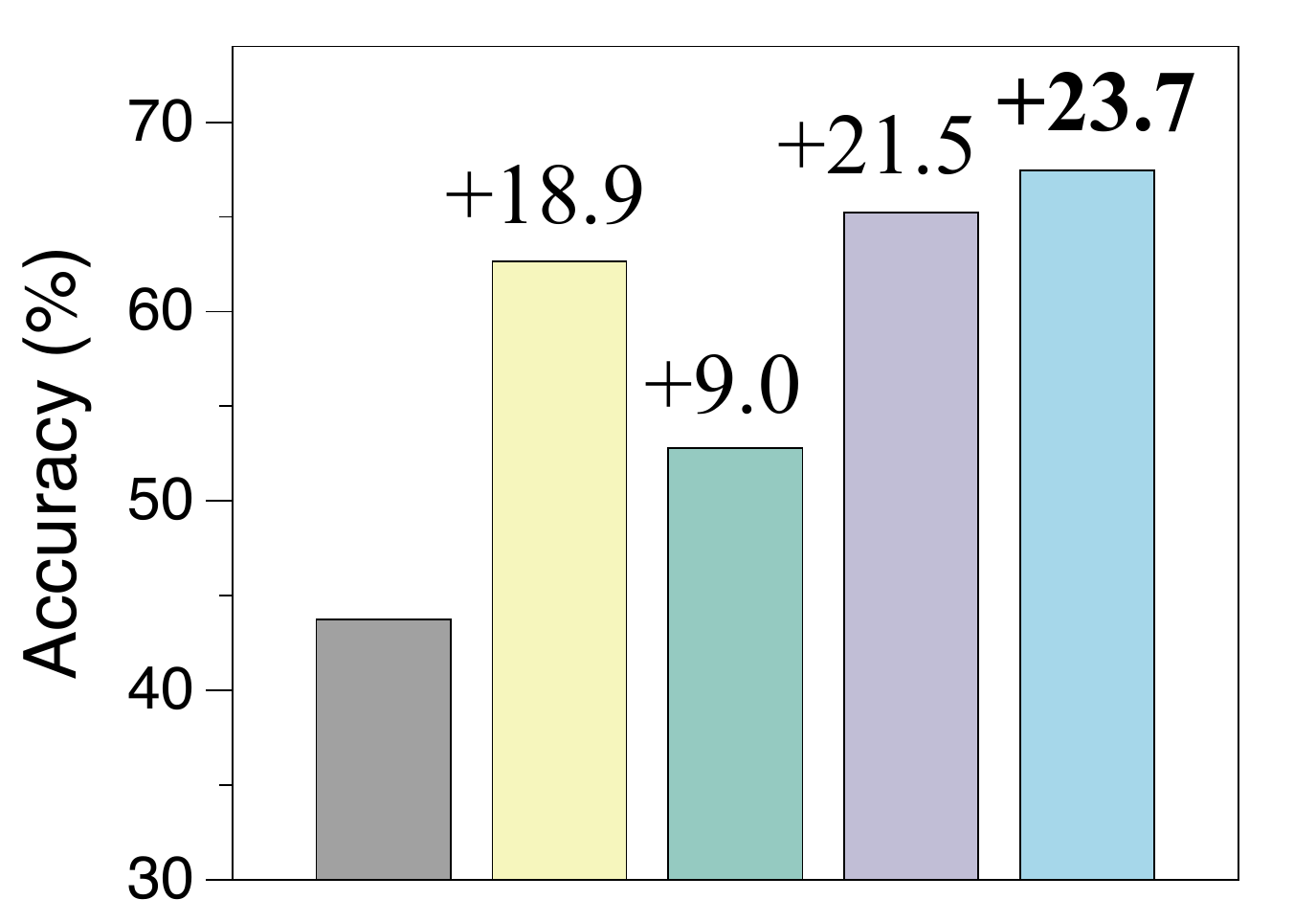}
    \vspace{-15pt}
    \caption{Action}
\end{subfigure}
\hfill
\begin{subfigure}[b]{0.48\linewidth}
    \centering
    \includegraphics[width=\textwidth]{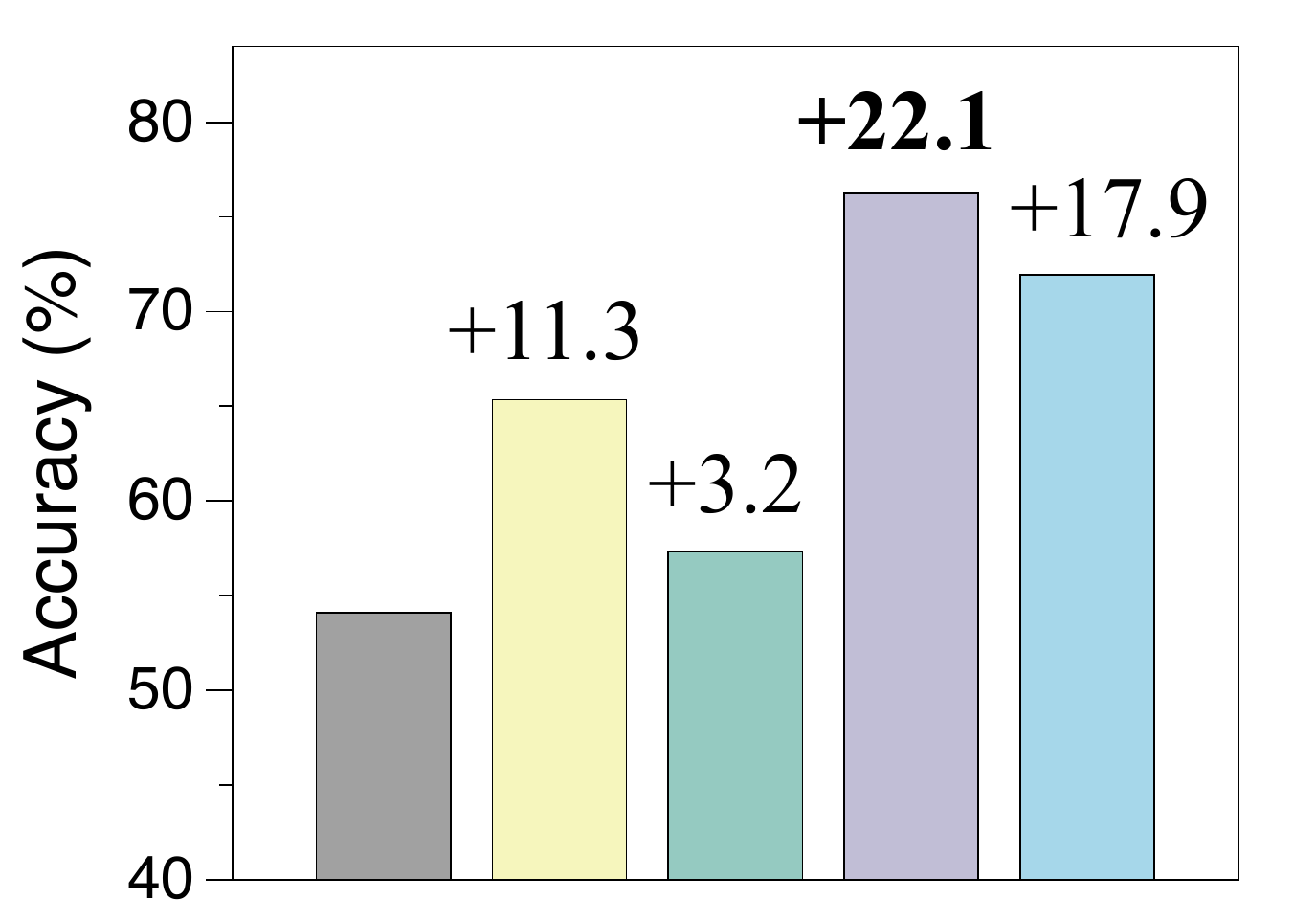}
    \vspace{-15pt}
    \caption{Indoor}
\end{subfigure}
\vspace{-5pt}
\caption{SimCore performances compared to the $X$ pretraining (w/o \textit{OS}). In addition to ImageNet-1k, we included MS COCO, iNaturalist 2021-mini, and Places365 as an open-set. For target datasets, we used Pet and Birds for natural image datasets and Action and Indoor for unnatural image datasets.}
\label{fig:diff_openset}
\end{figure}

\begin{figure}[!t]
    \centering
    \includegraphics[width=\linewidth]{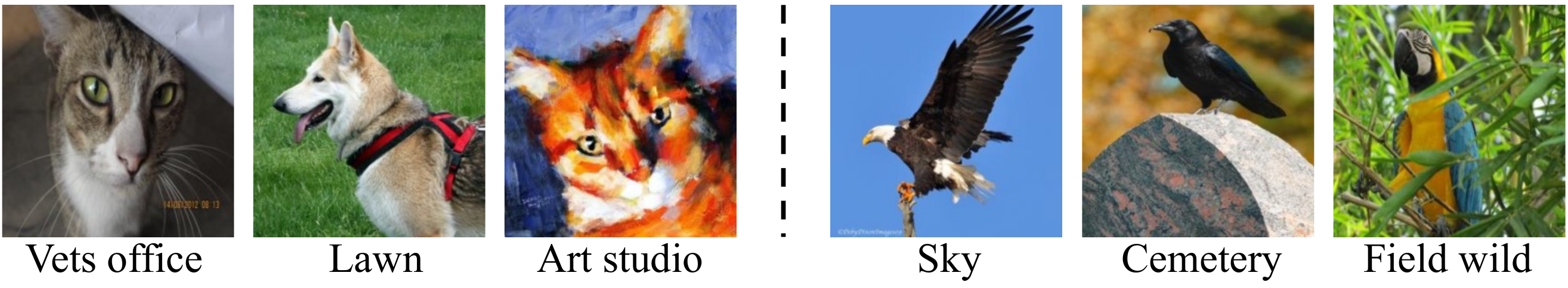} \\
    \vspace{3pt}
    \includegraphics[width=\linewidth]{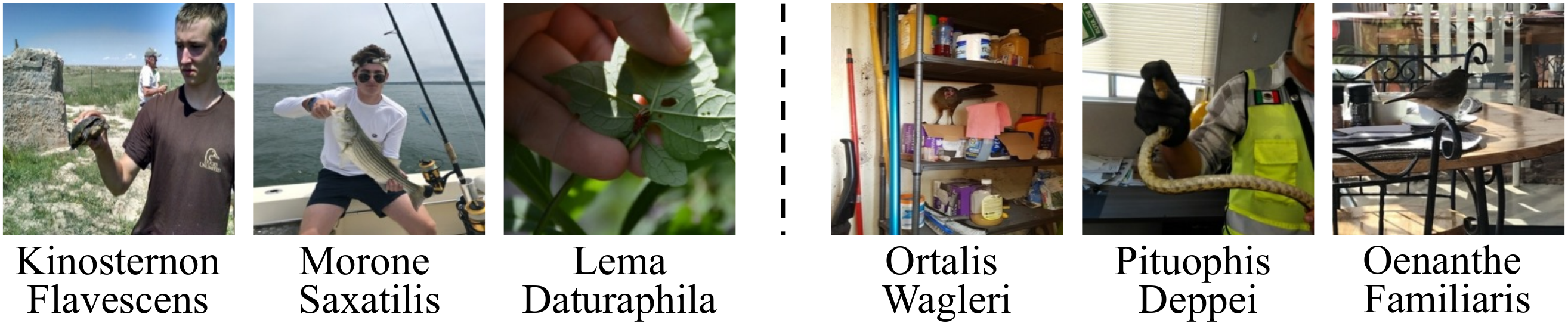}
    \vspace{-17pt}
    \caption{Selected samples of Places365 coreset\,(top) for Pet\,(left) and Birds\,(right) and iNaturalist coreset\,(bottom) for Action\,(left) and Indoor\,(right). Captions are the actual labels in each open-set.}
    \label{fig:places365_coreset}
\end{figure}

\subsection{SimCore on Various Open-Sets}
\label{subsec:simcore_on_various_opensets}

Thus far, we have used ImageNet-1k benchmark\cite{deng2009imagenet} as the open-set, which is a well-curated dataset covering general domains\cite{tian2021divide, goyal2021self}. Drawing a coreset from ImageNet has shown large performance gains in the target tasks. However, in practice, an open-set is far from what we know about; it is rather a grouping of data arbitrarily drawn from the web or database. 
Here, we show that our SimCore with any open-set robustly improves pretraining because it finds a well-matched coreset.
We experimented with three other open-sets: MS COCO\cite{lin2014microsoft}, iNaturalist 2021-mini\cite{van2018inaturalist}, and Places365\cite{zhou2017places}.

Figure\,\ref{fig:diff_openset} shows that SimCore consistently outperforms the pretraining without \textit{OS}, while the performance of SimCore depends on the open-set. 
In unnatural fine-grained datasets like Action and Indoor, using iNaturalist is not as effective as ImageNet, but it offers a good coreset for Birds. As expected, for Indoor, Places365 offers the best coreset among every open-sets because it contains a lot of scenery semantics.

Nevertheless, we have observed the \textit{unexpected} gains, such as Places365 open-set for Pet target dataset. Figure\,\ref{fig:places365_coreset} illustrates a few selected samples of those coresets. It is interesting to see that SimCore has found the animal images, although the actual labels correspond to the locations. Also, the iNaturalist coreset contains natural creatures that are held by humans or located in indoor. This might help in Action target, part of which are humans taking photos, fishing, or gardening, as well as in Indoor scenery target.

\vspace{-12pt}
\paragraph{Uncurated open-sets:}
To demonstrate the effect of SimCore in the more realistic scenario, we have tested SimCore with uncurated open-sets.
First, we used a combined dataset of all four pre-mentioned open-sets\,(\texttt{ALL}), to simulate the more large-scaled and heterogeneous open-set case.
In addition, we further used web-crawled image dataset queried by ImageNet classes\,(WebVision\,\cite{li2017webvision}), and the images queried by Aircraft+Cars+Birds classes\,(WebFG-496\,\cite{sun2021webly}). Interestingly, in the case of WebFG-496 that includes noisy instances, SimCore sampled slightly less of the actual crawled set for each target.
For example, while WebFG-496 contains 13,508 number of queried instances for Aircraft, SimCore sampled a coreset of 8,089 instances.
Table\,\ref{tab:uncurated_openset} demonstrates that these uncurated open-sets are comparable to the curated ones and significantly outperform the pretraining without open-sets. Indeed, SimCore could sample a useful coreset regardless of how uncurated an open-set is.

\begin{table}[!t]
    \small
    \centering
    \resizebox{\linewidth}{!}{
    \addtolength{\tabcolsep}{2.0pt}
    \renewcommand*{\arraystretch}{0.9}
    \begin{tabular}{lcccccc}
        \toprule
        \textit{OS} & \!\!\!Aircraft\!\!\! & Cars & Birds & Pet & \!\!\!Action\!\!\! & \!\!Indoor\!\! \\
        \midrule
        \xmark & 46.6 & 55.4 & 29.3 & 59.2 & 43.8 & 54.1 \\
        ImageNet-1k & 48.3 & 60.3 & 37.7 & 79.7 & \textbf{67.5} & 72.0 \\
        \midrule
        \texttt{ALL} & \bf 58.8 & \bf 64.8 & \textbf{38.0} & 79.5 & 67.2 & \textbf{75.1} \\
        WebVision & 48.2 & 59.1 & 36.7 & \textbf{80.3} & 67.1 & 72.6 \\
        WebFG-496 & 55.4 & 63.1 & 37.7 & - & - & - \\
        \bottomrule
    \end{tabular}
    }
    \vspace{-5pt}
    \caption{Linear evaluation of SimCore with uncurated open-sets.}
    \label{tab:uncurated_openset}
\end{table}

\subsection{Qualitative Evaluation}
\label{subsec:analysis_of_simcore}

\begin{figure}[!t]
    \centering
    \begin{subfigure}[b]{0.48\linewidth}
    \begin{subfigure}[b]{0.32\linewidth}
    \centering
    \includegraphics[width=\linewidth]{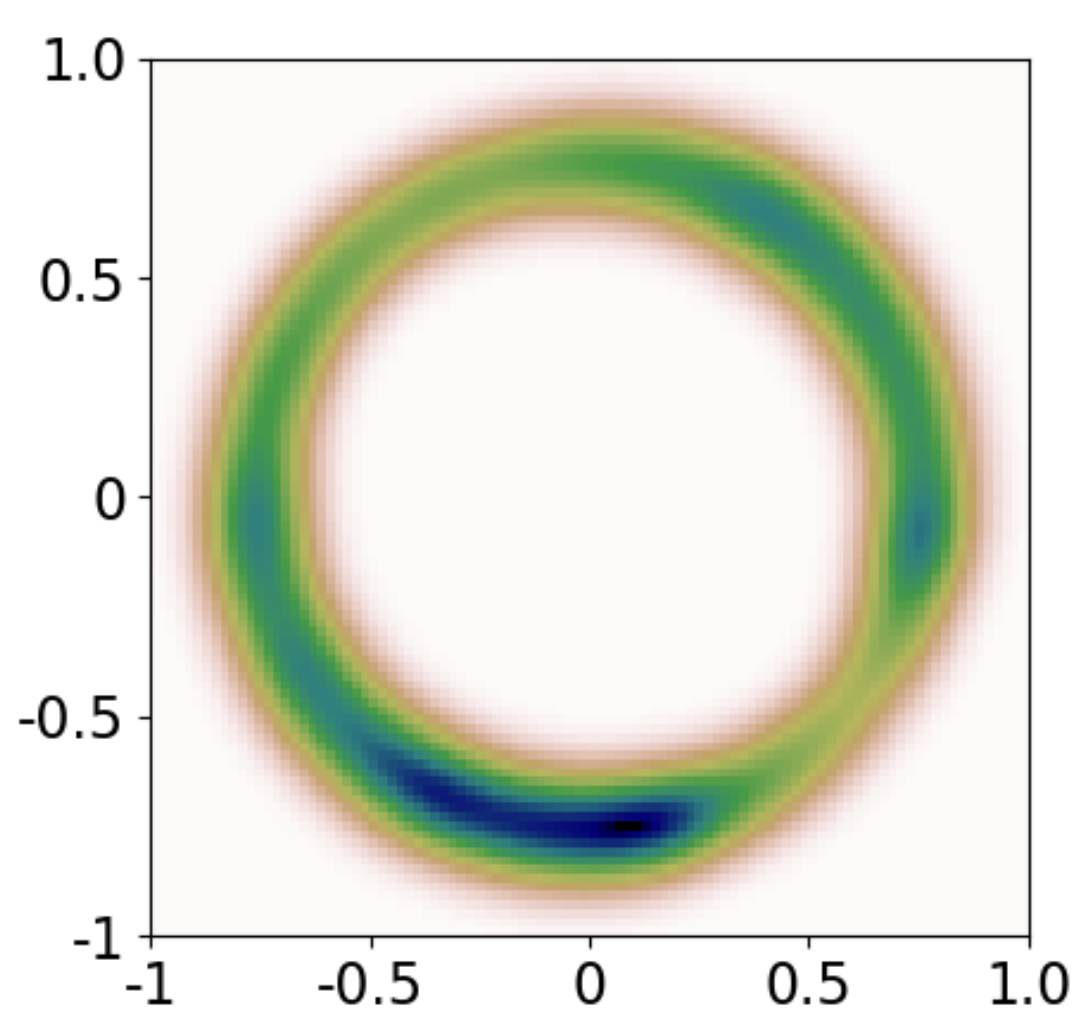}
    \end{subfigure}
    \begin{subfigure}[b]{0.32\linewidth}
    \centering
    \includegraphics[width=\linewidth]{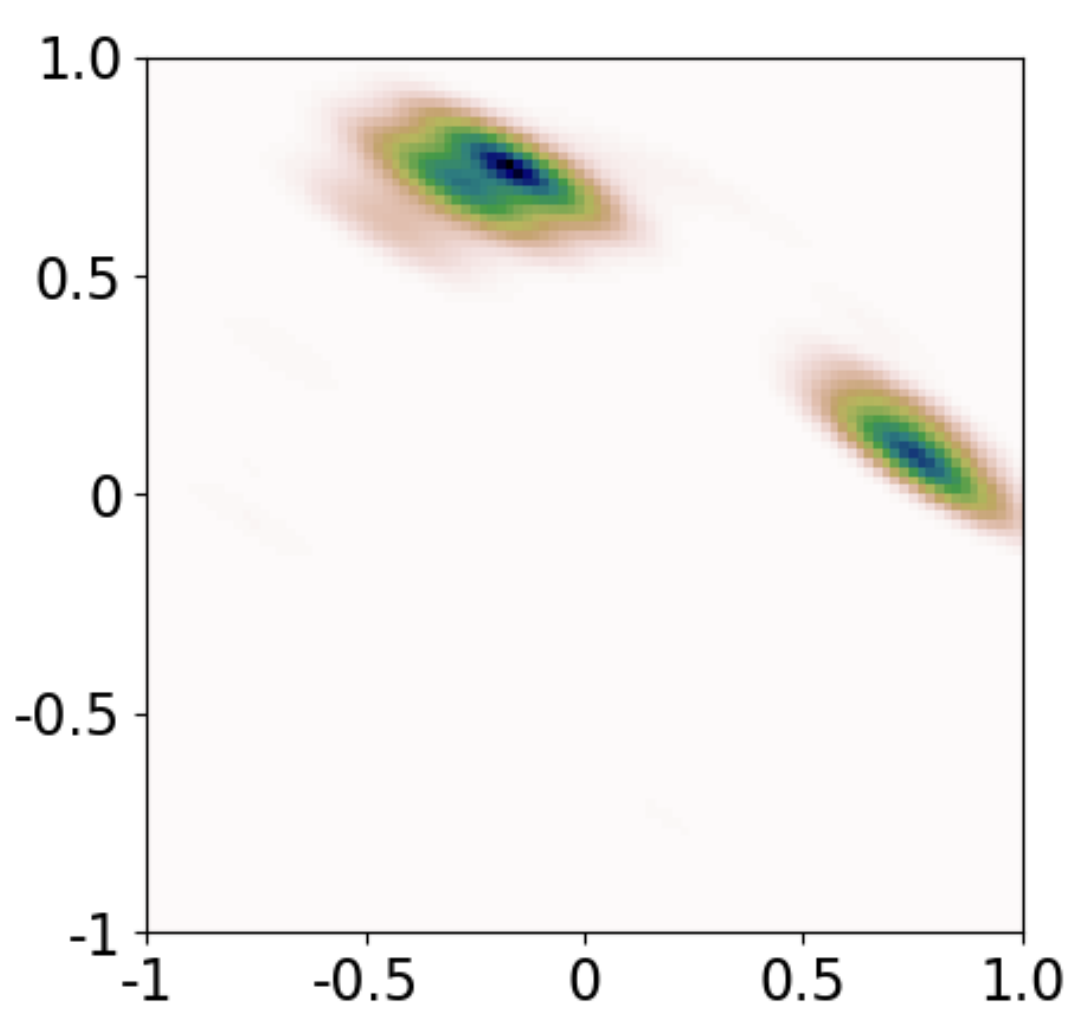}
    \end{subfigure}
    \begin{subfigure}[b]{0.32\linewidth}
    \centering
    \includegraphics[width=\linewidth]{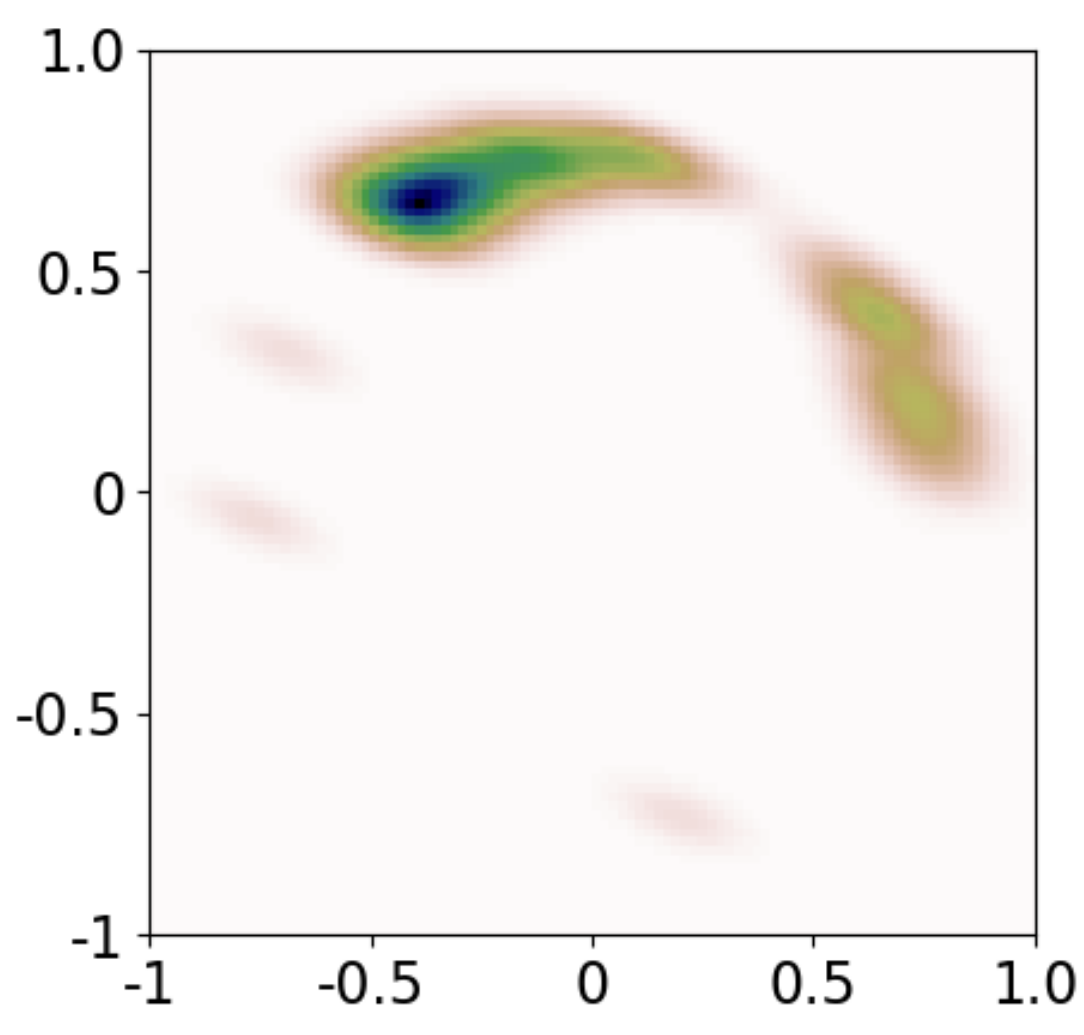}
    \end{subfigure}
    \caption{Aircraft}
    \end{subfigure}
    \hfill
    \begin{subfigure}[b]{0.48\linewidth}
    \begin{subfigure}[b]{0.32\linewidth}
    \centering
    \includegraphics[width=\linewidth]{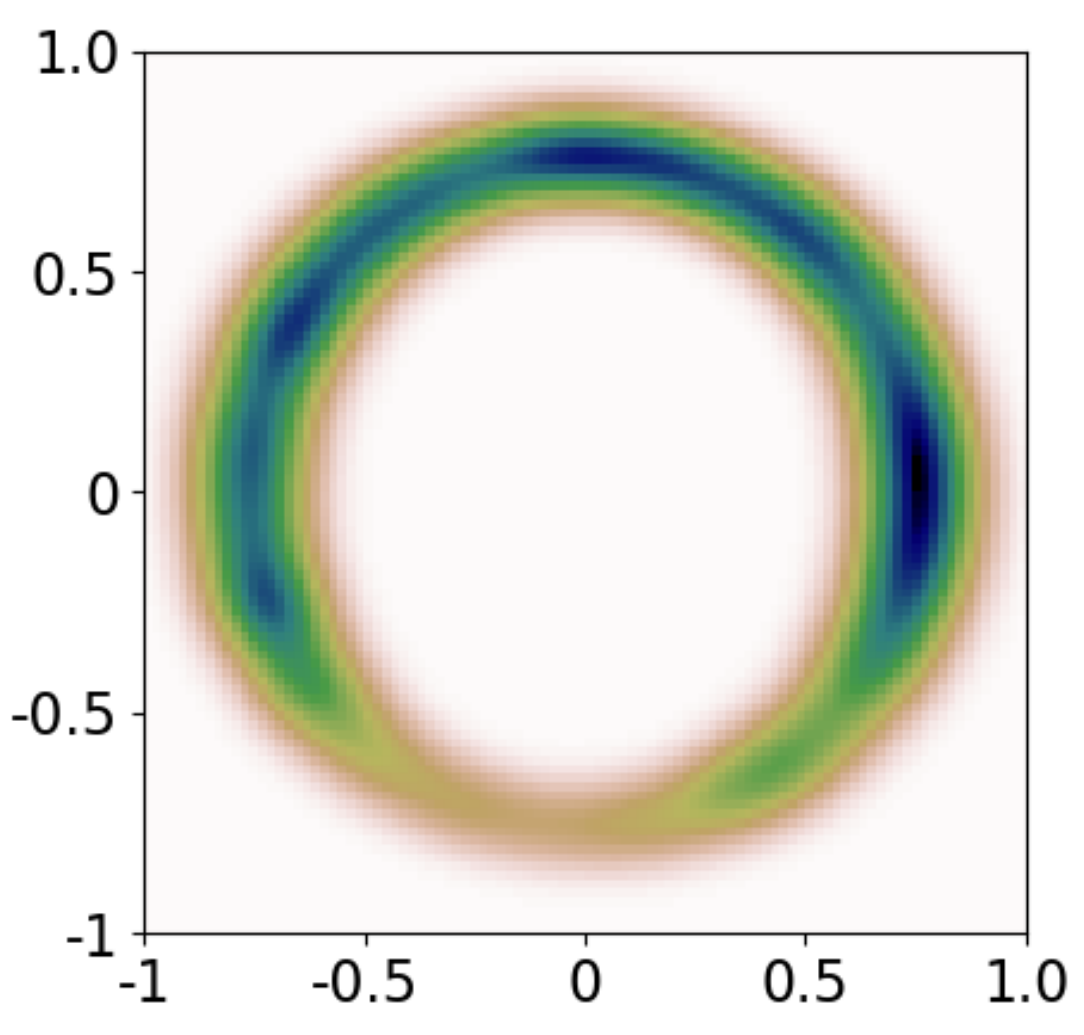}
    \end{subfigure}
    \begin{subfigure}[b]{0.32\linewidth}
    \centering
    \includegraphics[width=\linewidth]{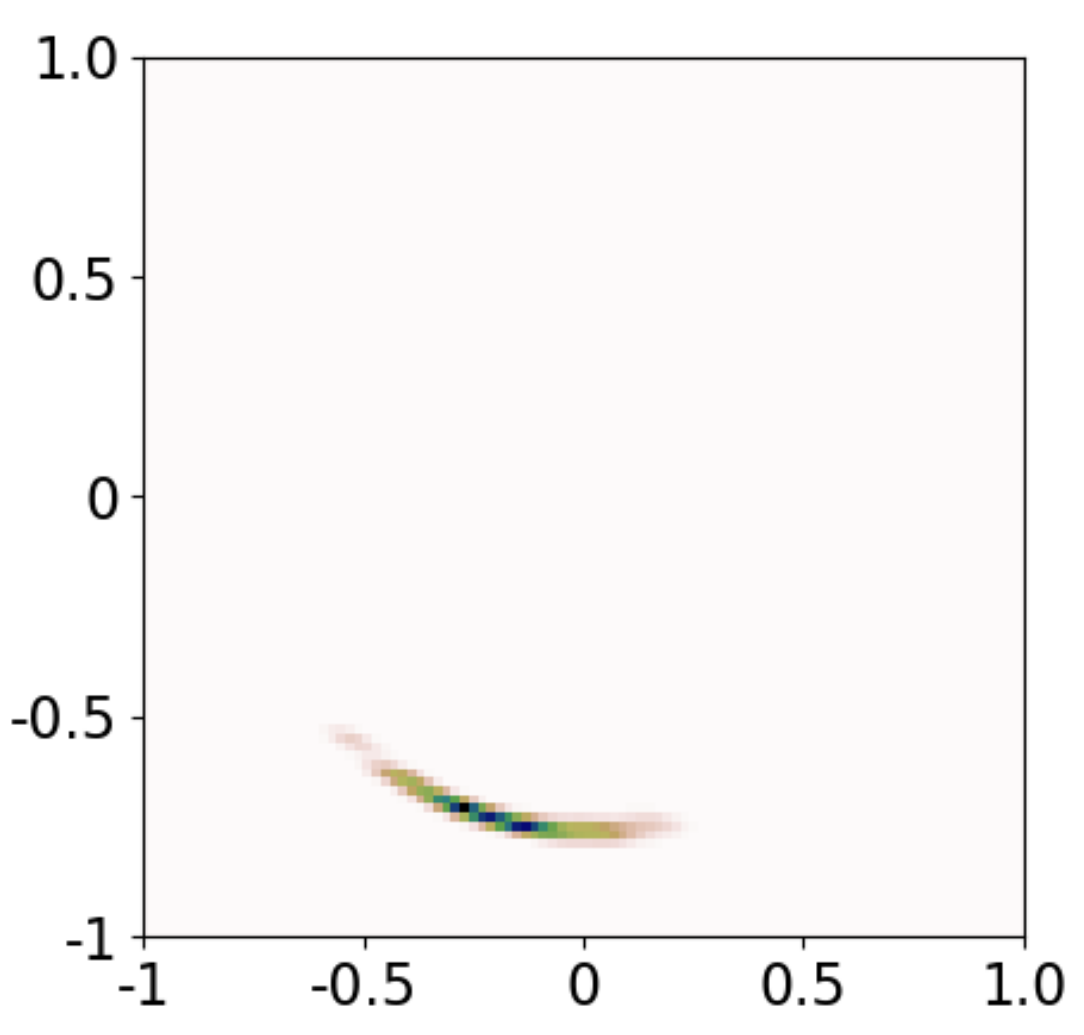}
    \end{subfigure}
    \begin{subfigure}[b]{0.32\linewidth}
    \centering
    \includegraphics[width=\linewidth]{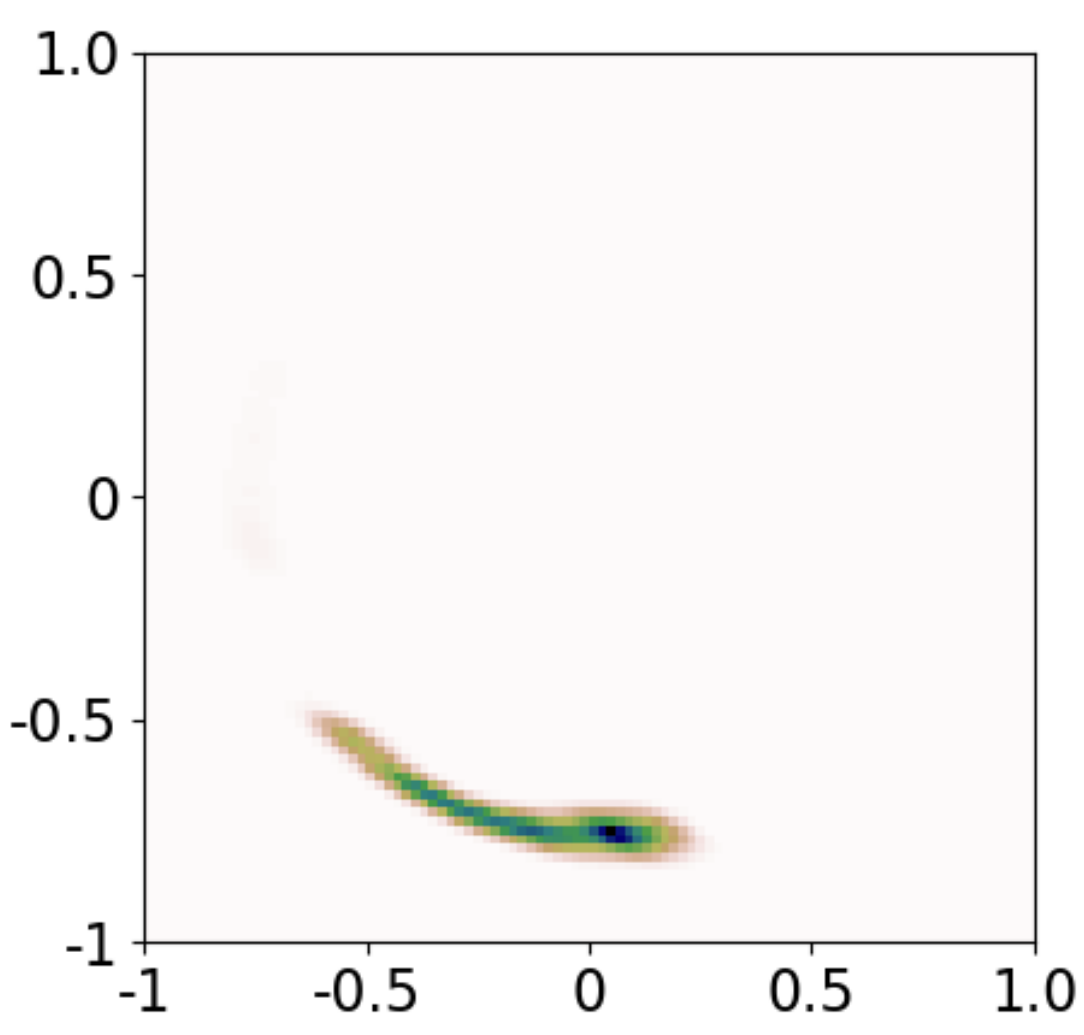}
    \end{subfigure}
    \caption{Cars}
    \end{subfigure}

    \begin{subfigure}[b]{0.48\linewidth}
    \begin{subfigure}[b]{0.32\linewidth}
    \centering
    \includegraphics[width=\linewidth]{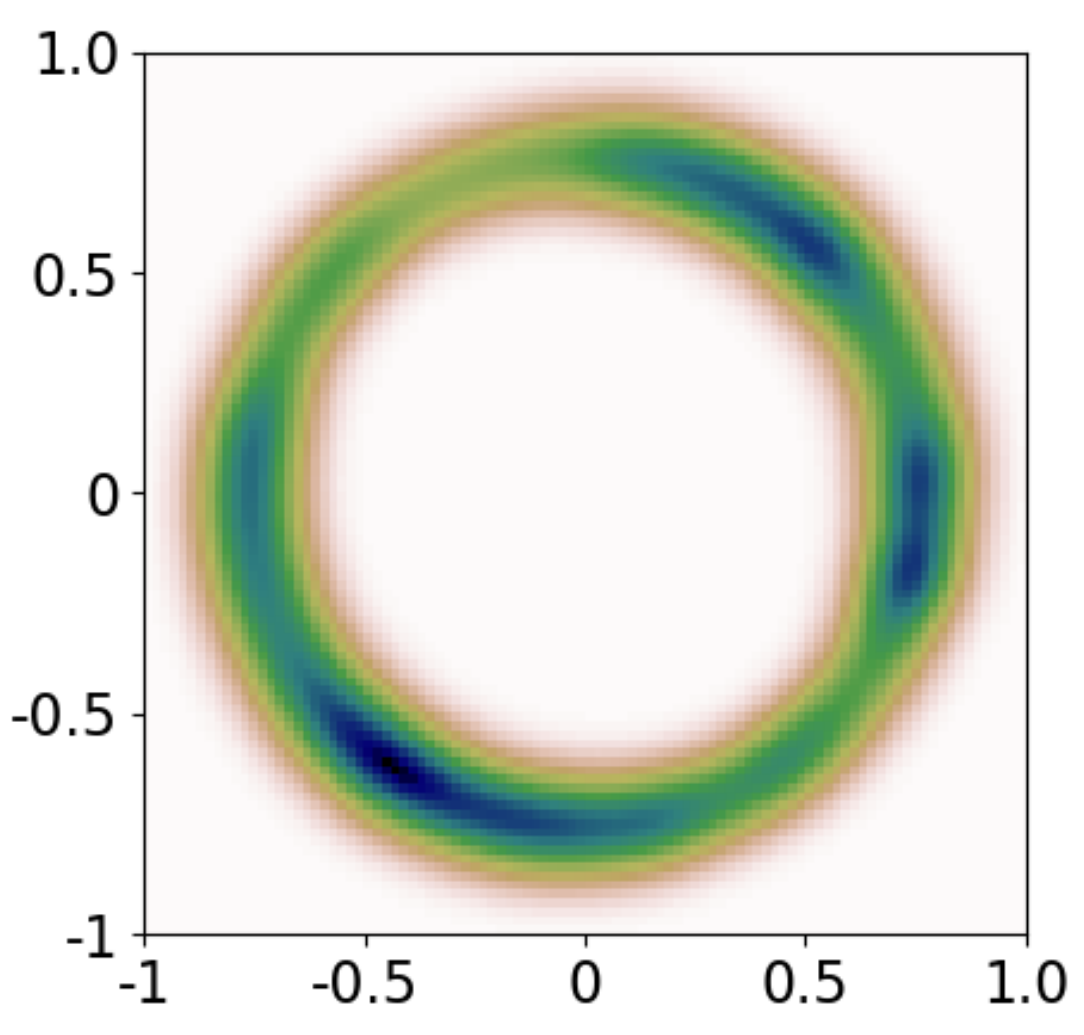}
    \end{subfigure}
    \begin{subfigure}[b]{0.32\linewidth}
    \centering
    \includegraphics[width=\linewidth]{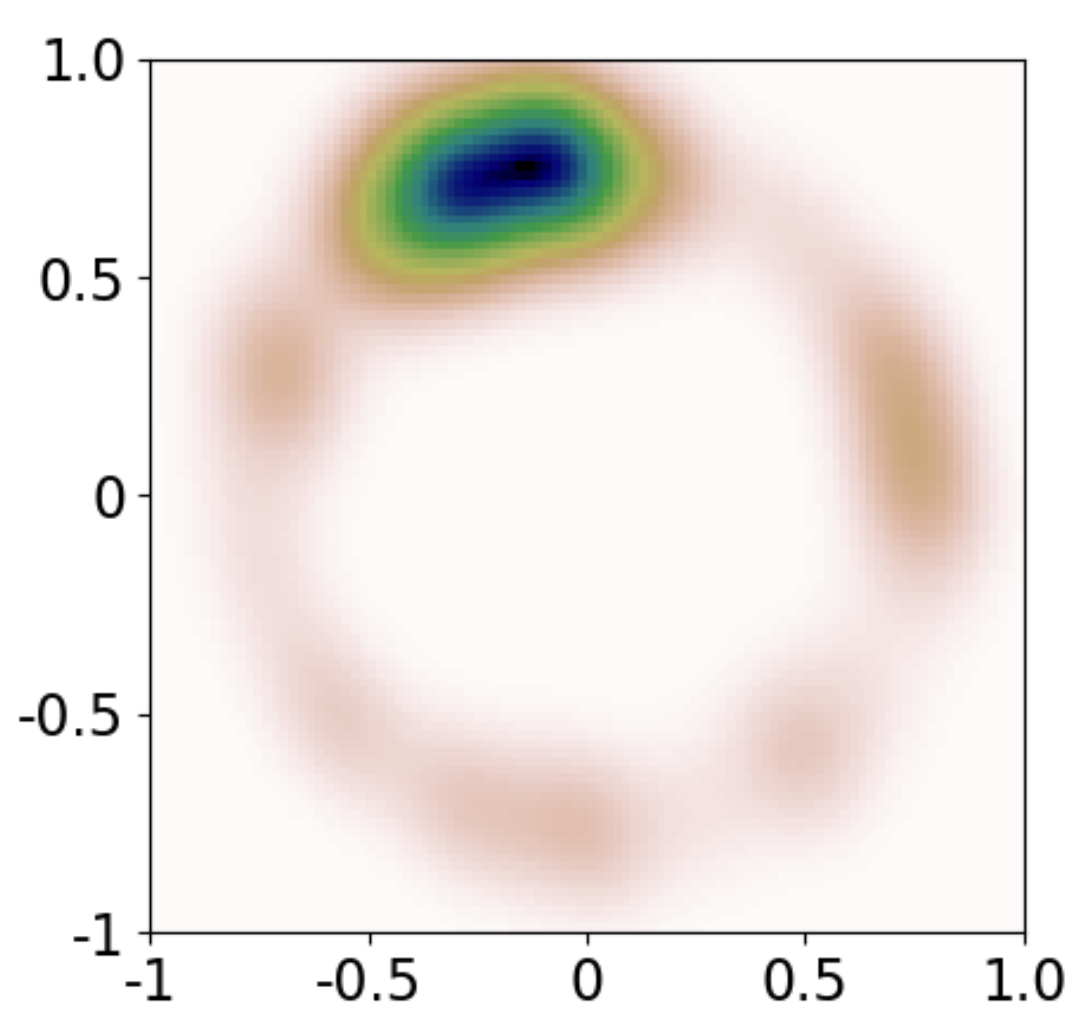}
    \end{subfigure}
    \begin{subfigure}[b]{0.32\linewidth}
    \centering
    \includegraphics[width=\linewidth]{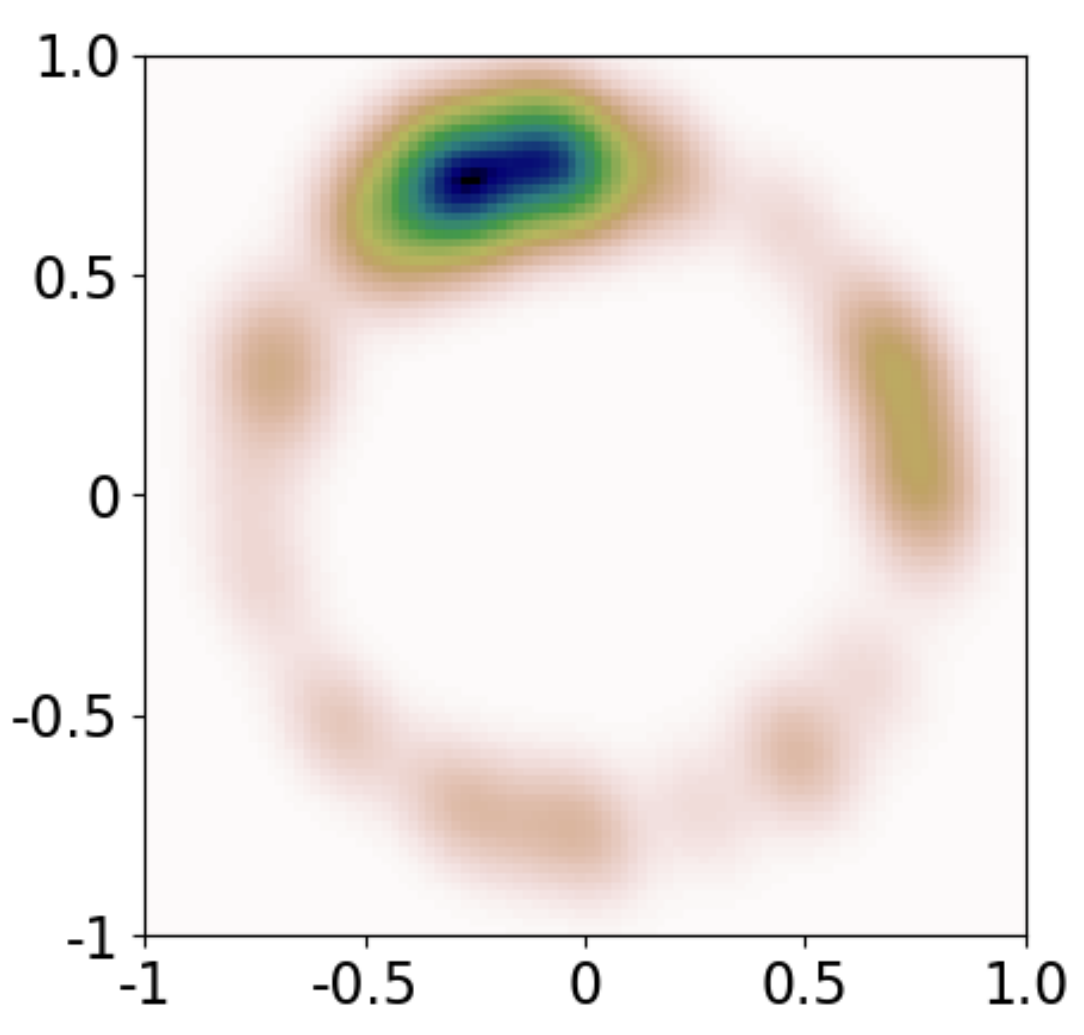}
    \end{subfigure}
    \caption{Pet}
    \end{subfigure}
    \hfill
    \begin{subfigure}[b]{0.48\linewidth}
    \begin{subfigure}[b]{0.32\linewidth}
    \centering
    \includegraphics[width=\linewidth]{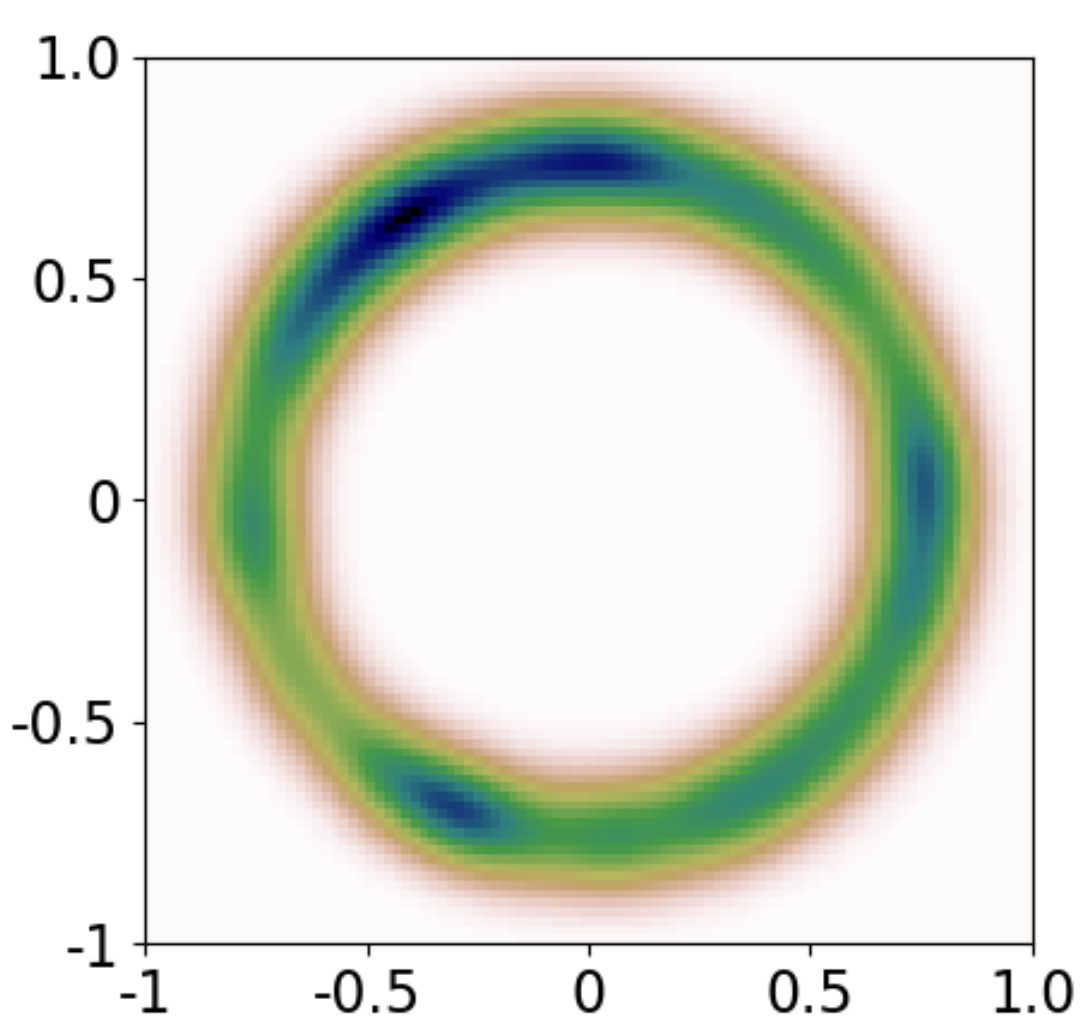}
    \end{subfigure}
    \begin{subfigure}[b]{0.32\linewidth}
    \centering
    \includegraphics[width=\linewidth]{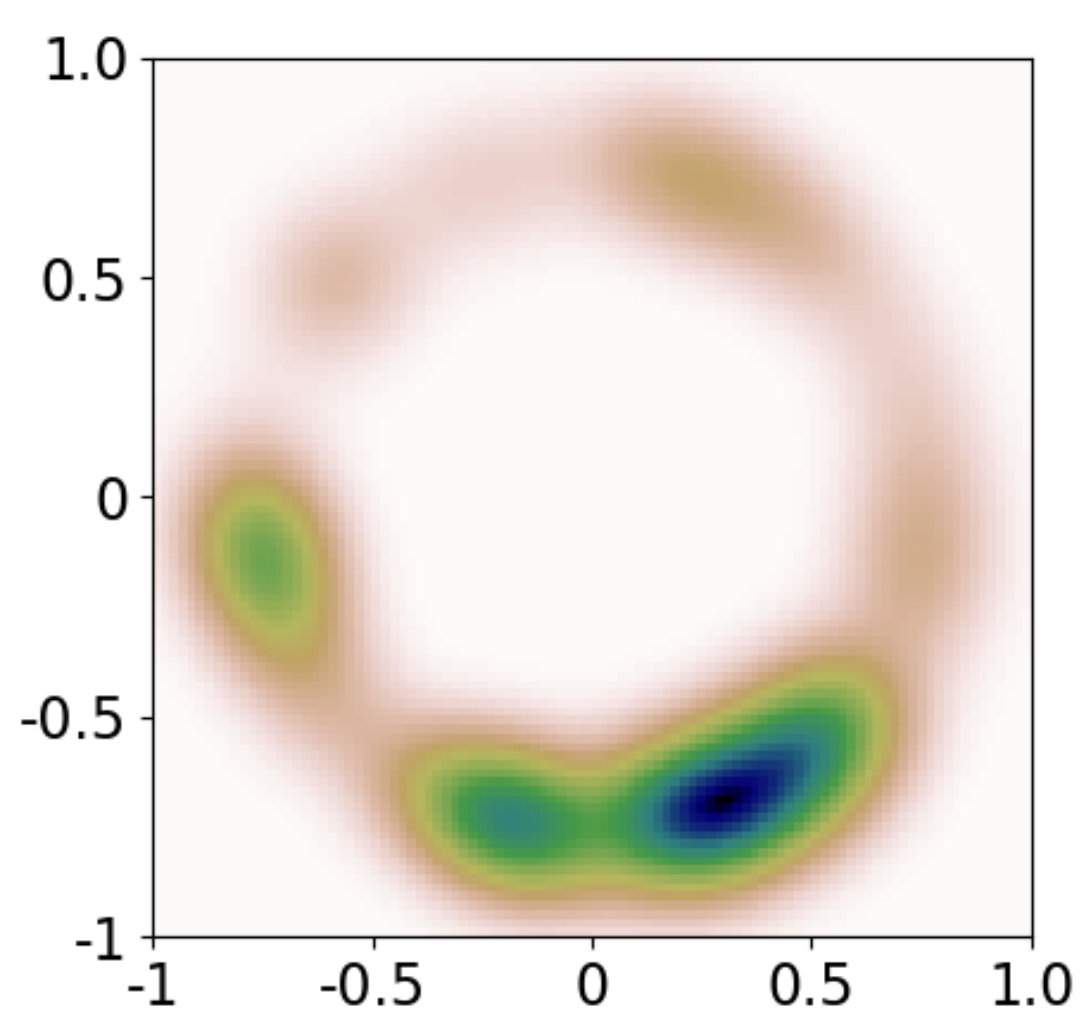}
    \end{subfigure}
    \begin{subfigure}[b]{0.32\linewidth}
    \centering
    \includegraphics[width=\linewidth]{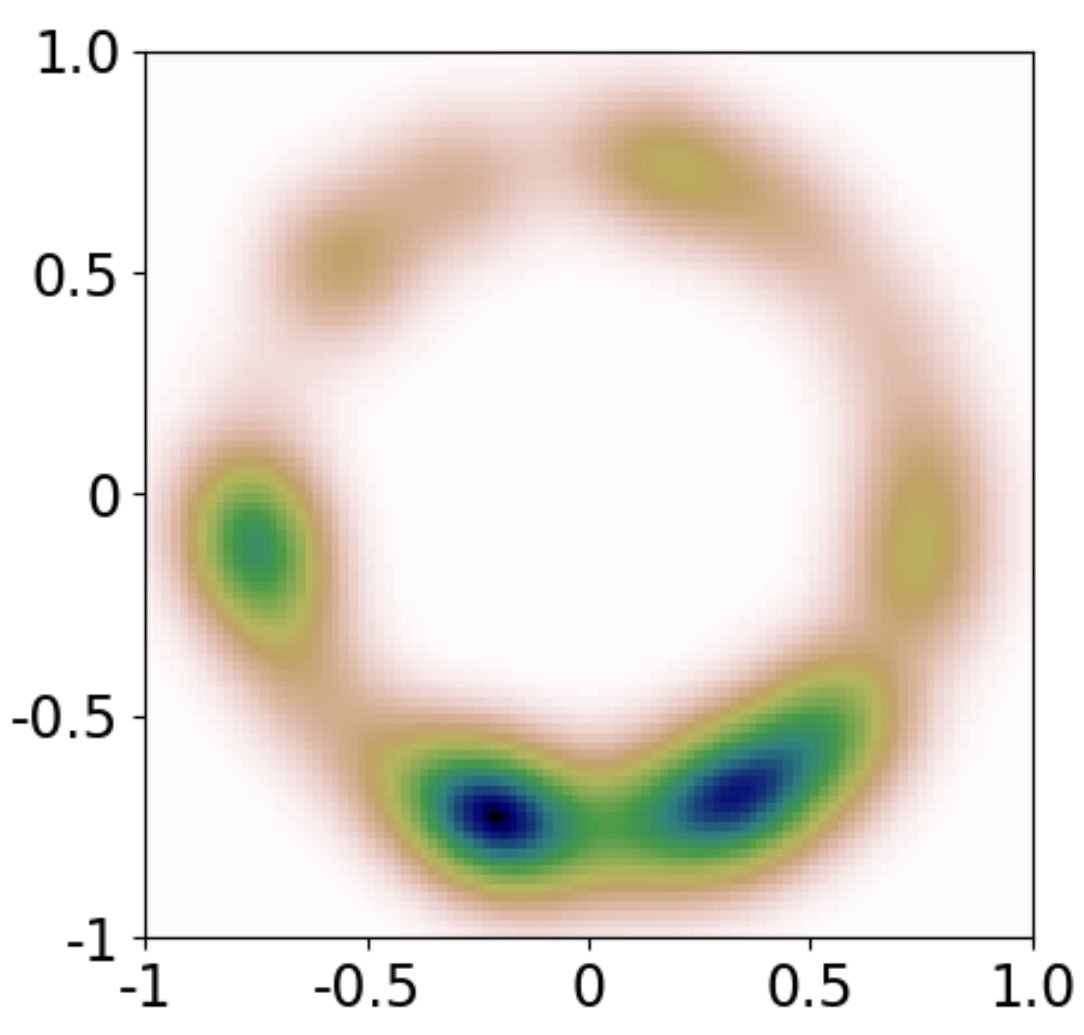}
    \end{subfigure}
    \caption{Birds}
    \end{subfigure}
    \vspace{-5pt}
    \caption{Feature distribution map of \textit{OS}\,(left), $X$\,(middle), and the coreset sampled by SimCore\,(right).}
    \label{fig:feat_dist}
\end{figure}

\begin{figure*}[!t]
\begin{subfigure}[b]{0.49\textwidth}
    \raggedleft
    \includegraphics[width=\textwidth]{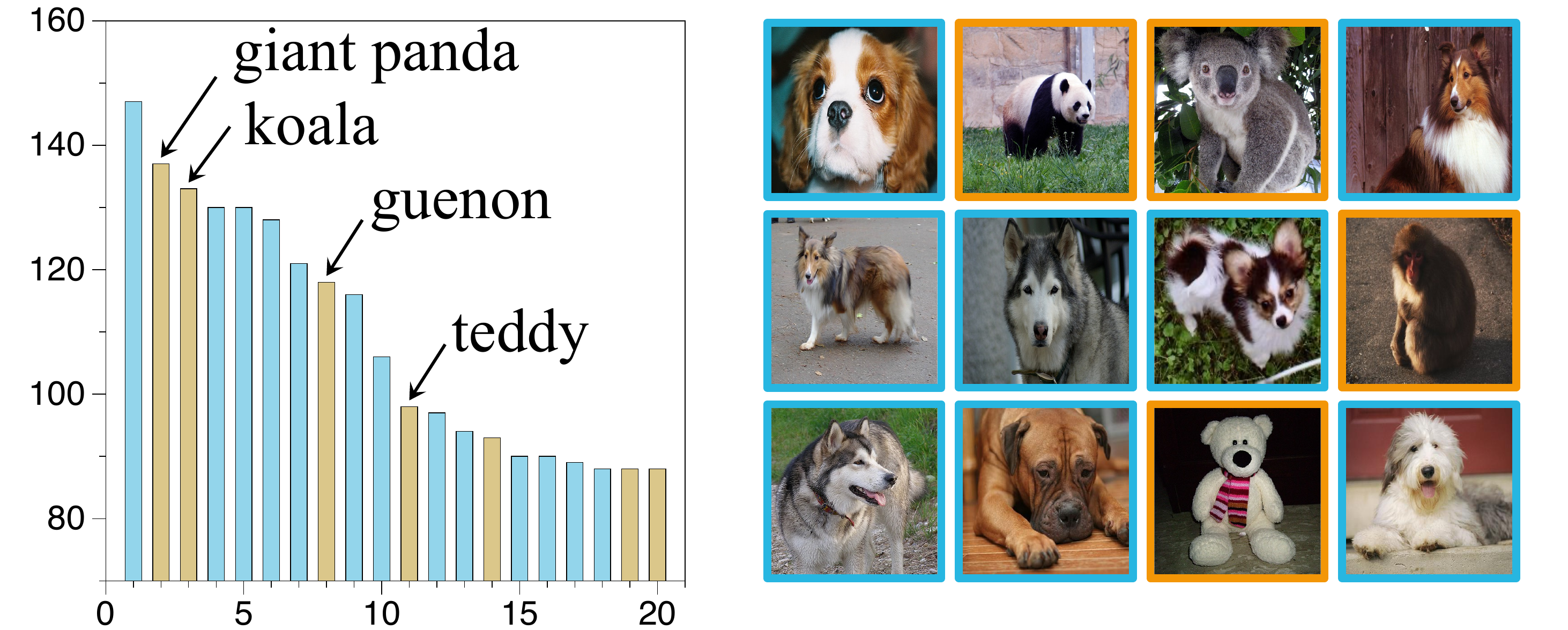}
\end{subfigure}
\tikz{\draw[-,black, densely dashed, thick](0,1.05) -- (0, 4.55);}
\hfill
\begin{subfigure}[b]{0.49\textwidth}
    \raggedright
    \includegraphics[width=\textwidth]{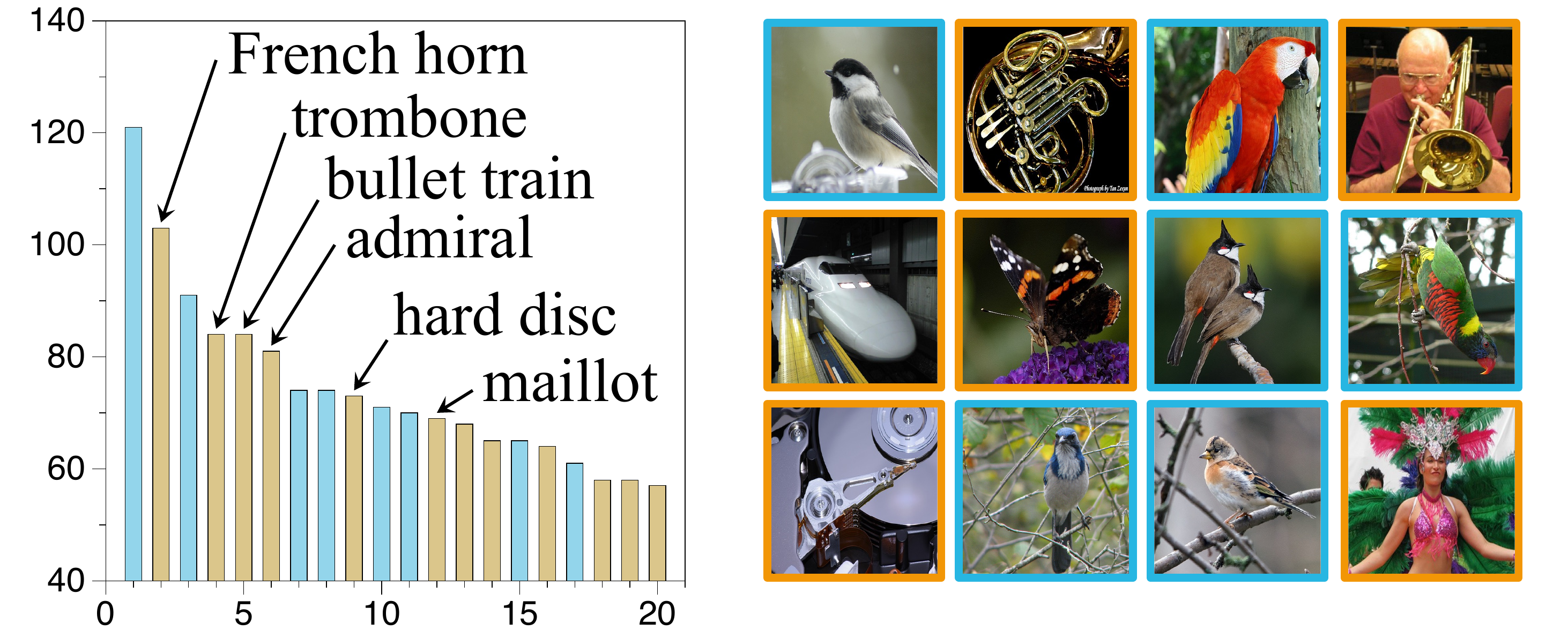}
\end{subfigure}
\hfill
\begin{subfigure}[b]{0.49\textwidth}
    \vspace{5pt}
    \raggedleft
    \includegraphics[width=\textwidth]{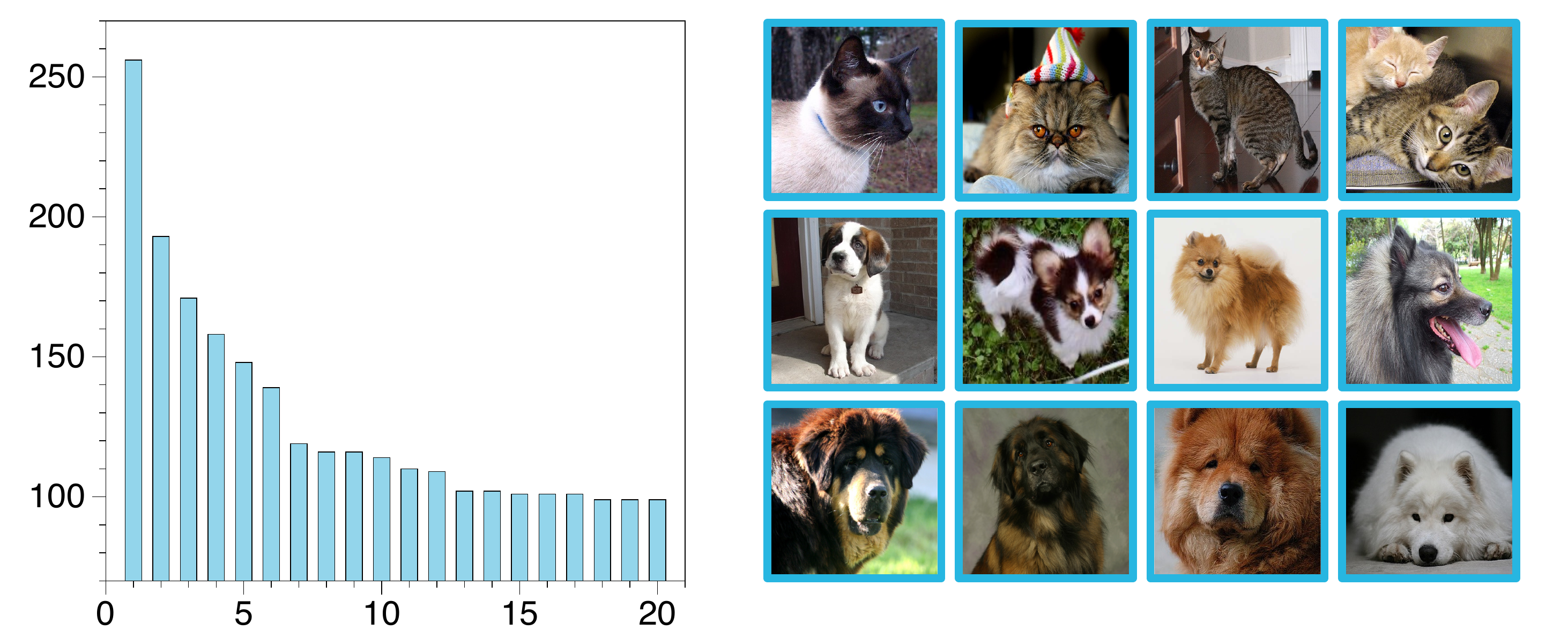}
    \caption{$\textit{OS}_{\text{SimCore}}$ with $X$ = Pet and \{$k$ = 1\,(top), $k$ = 100\,(bottom)\}}
    \label{fig:coreset_vis_a}
\end{subfigure}
\tikz{\draw[-,black, densely dashed, thick](0, 1.55) -- (0, -2.55);}
\hfill
\begin{subfigure}[b]{0.49\textwidth}
    \vspace{5pt}
    \raggedright
    \includegraphics[width=\textwidth]{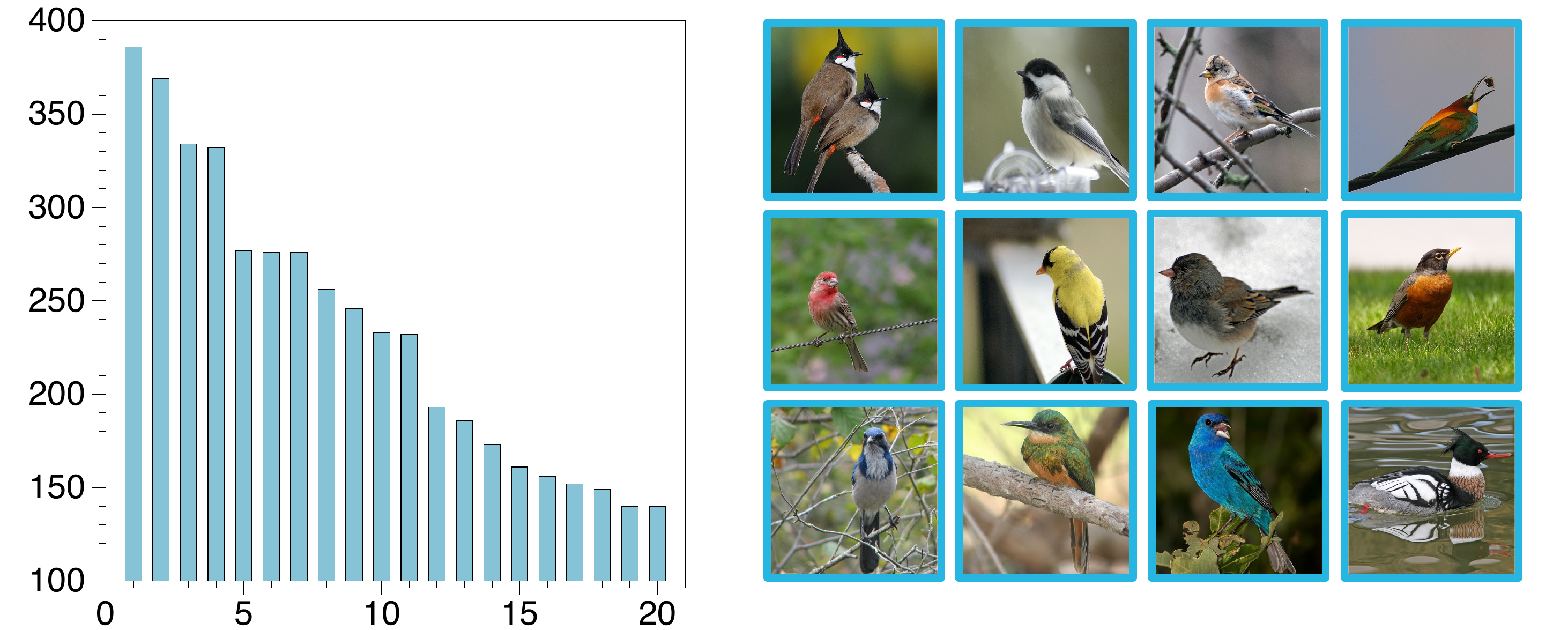}
    \caption{$\textit{OS}_\text{SimCore}$ with $X$ = Birds and \{$k$ = 1\,(top), $k$ = 100\,(bottom)\}}
    \label{fig:coreset_vis_b}
\end{subfigure}
\caption{Visualization of the sampled coreset by SimCore method. We plotted histograms for the top-20 classes with the largest number of samples and visualized one example image per top-12 classes. We highlighted with orange the coreset classes that look dissimilar to the target data. $k$ is the number of centroids in $k$-means clustering to reduce the complexity of SimCore. The x-axis and y-axis of histograms denote the class index and the number of samples, respectively.}
\label{fig:coreset_vis}
\vspace{-5pt}
\end{figure*}

\paragraph{Feature distribution analysis:}
To analyze how the SimCore algorithm samples the coreset in the latent space, we visualized the feature distribution on a unit ring by Gaussian kernel density estimation in $\mathbb{R}^2$\,\cite{wang2020understanding} (implementation details in Appendix\,\ref{appx:implementation_details_feat_dist}).
The results in Figure\,\ref{fig:feat_dist} are interesting, as SimCore actually samples the instances that are closely embedded to the target data. 
In addition, through a comparison of the occupied areas of \textit{OS} and $X$, we confirmed the distribution similarity of the open-set to each target dataset, indicating the sampling ratios by SimCore in Table\,\ref{tab:main_exp} are reasonable.

\vspace{-12pt}
\paragraph{Coreset visualization:}
We visualized which instances from the open-set are actually sampled by our SimCore algorithm. To this end, we displayed in Figure\,\ref{fig:coreset_vis} the ground-truth labels of the coreset samples when the target dataset is Pet or Birds. For comparison, we also displayed the coreset by SimCore with $k=1$, using a single centroid.
We used the open-set as ImageNet, so the ground-truth labels of coreset samples correspond to the ImageNet classes.
Note that Pet contains 12 cat breeds and 25 dog breeds\cite{pet}, and Birds contains 200 bird species\cite{birds}. 

For the Pet dataset (Figure\,\ref{fig:coreset_vis_a}), SimCore with $k=1$ sampled mostly animal images, but included data somewhat irrelevant to cats and dogs. The second most class is giant panda, the third is koala, and the eighth is guenon, a kind of monkey. On the contrary, SimCore with $k=100$ sampled mostly cat or dog images, with up to top-20 classes, each being a breed of either cat or dog. Interestingly, eight out of the top-10 classes were those that overlap with the Pet class labels, such as Persian cat, Saint Bernard, etc. 

For the Birds dataset (Figure\,\ref{fig:coreset_vis_b}), SimCore with $k=1$ sampled a lot of irrelevant images, such as French horn, trombone, bullet train, admiral, hard disc, maillot, etc. On the contrary, SimCore with $k=100$ sampled only the bird species up to the top-20 classes, including bulbul, chickadee, brambling, bee eater, house finch, goldfinch, etc.

\begin{table}[!t]
    \small
    \centering
    \addtolength{\tabcolsep}{-3.5pt}
    \renewcommand*{\arraystretch}{1.1}
    \resizebox{\linewidth}{!}{
    \begin{tabular}{l|lccc|lccc}
    \toprule
    & \multicolumn{4}{c|}{framework: \textit{Open-Set Semi-Sup.}} & \multicolumn{4}{c}{framework: \textit{Webly Sup.}} \\
    \hline
    pretrain & method & \!\!Aircraft\!\! & Cars & Birds & method & \!\!Aircraft\!\! & Cars & Birds \\
    \hline
    SimCore & FT\,(50\%) & 73.5 & 80.1 & 57.4 & FT\,(100\%) & 84.3 & \textbf{89.3} & 70.6 \\
    \hline
    \xmark & SelfTrain & 51.9 & 55.5 & 35.7 & CoTeach & 79.3 & 51.7 & 70.4 \\
    SimCore & SelfTrain & \textbf{78.1} & \textbf{81.3} & \textbf{59.1} & CoTeach & \textbf{89.8} & 57.0 & \textbf{78.9} \\
    \hline
    \xmark & OpenMatch & 70.1 & 70.2 & 52.3 & DivideMix & 82.2 & 54.4 & 74.5 \\
    SimCore & OpenMatch & \textbf{83.5} & \textbf{89.5} & \textbf{66.4} & DivideMix & \textbf{86.5} & 56.5 & \textbf{80.0} \\
    \bottomrule
    \end{tabular}
    }
    \vspace{-5pt}
    \caption{Comparisons with Self-Training\,\cite{su2021realistic} and OpenMatch\,\cite{saito2021openmatch} in the OpenSemi framework, and Co-teaching\,\cite{han2018co} and DivideMix\,\cite{li2020dividemix} in the WeblySup framework.}
    \label{tab:opensemi_weblysup}
\end{table}

\subsection{Comparisons with Open-Set Semi-Supervised and Webly Supervised Learning}

\begin{table*}[!t]
    \begin{subfigure}[b]{0.43\textwidth}
    \centering
    \small
    \addtolength{\tabcolsep}{-1.4pt}
    \resizebox{\textwidth}{!}{
    \begin{tabular}{lcccccccc}
    \toprule
         & \multicolumn{2}{c}{Aircraft} & \multicolumn{2}{c}{Cars} & \multicolumn{2}{c}{Pet} & \multicolumn{2}{c}{Birds} \\
         \cmidrule(l{2pt}r{2pt}){2-3} \cmidrule(l{2pt}r{2pt}){4-5} \cmidrule(l{2pt}r{2pt}){6-7} \cmidrule(l{2pt}r{2pt}){8-9}
        pretrain & \!20\! & \!\!200\!\! & \!20\! & \!\!200\!\! & \!20\! & \!\!200\!\! & \!20\! & \!\!200\!\! \\
        \midrule
        $X$ & 36.1 & 36.7 & 33.1 & 34.3 & 52.0 & 51.8 & \bf 20.7 & \bf 21.8 \\
        \textit{OS} & 19.3 & 17.7 & 11.4 & 10.9 & 50.4 & 49.0 & 13.9 & 15.1 \\
        \bf SimCore & \bf 40.7 & \bf 41.4 & \bf 33.8 & \bf 34.6 & \bf 61.4 & \bf 61.4 & 18.3 & 19.2 \\
    \bottomrule
    \end{tabular}}
    \caption{kNN classification}
    \label{tab:knn_classification}
    \end{subfigure}
    \hfill
    \begin{subfigure}[b]{0.55\textwidth}
    \centering
    \small
    \addtolength{\tabcolsep}{-3pt}
    \resizebox{\textwidth}{!}{
    \begin{tabular}{lcccccccccccc}
    \toprule
         & \multicolumn{3}{c}{Aircraft} & \multicolumn{3}{c}{Cars} & \multicolumn{3}{c}{Pet} & \multicolumn{3}{c}{Birds} \\
         \cmidrule(l{2pt}r{2pt}){2-4} \cmidrule(l{2pt}r{2pt}){5-7} \cmidrule(l{2pt}r{2pt}){8-10} \cmidrule(l{2pt}r{2pt}){11-13}
        pretrain & 10\% & 20\% & 50\% & 10\% & 20\% & 50\% & 10\% & 20\% & 50\% & 10\% & 20\% & 50\% \\
        \midrule
        $X$ & 29.0 & 47.6 & 64.6 & 25.1 & 53.5 & 80.2 & 47.2 & 58.7 & 71.4 & \bf 13.3 & 25.2 & 51.2 \\
        \textit{OS} & 19.6 & 34.1 & 43.9 & 10.8 & 35.7 & 74.1 & 35.7 & 62.3 & 76.9 & 10.1 & 21.0 & 51.2 \\
        \bf SimCore & \bf 33.9 & \bf 51.3 & \bf 65.6 & \bf 25.4 & \bf 55.3 & \bf 81.6 & \bf 50.9 & \bf 70.4 & \bf 79.7 & 11.9 & \bf 25.5 & \bf 55.7 \\
    \bottomrule
    \end{tabular}}
    \caption{Semi-supervised learning}
    \label{tab:semi-supervised_learning}
    \end{subfigure}
    \begin{subfigure}[b]{0.49\textwidth}
    \small
    \addtolength{\tabcolsep}{-2pt}
    \resizebox{\textwidth}{!}{
    \begin{tabular}{lcccccccc}
    \toprule
        & \multicolumn{2}{c}{Aircraft} & \multicolumn{2}{c}{Cars} & \multicolumn{2}{c}{Pet} & \multicolumn{2}{c}{Birds} \\
        \cmidrule(l{2pt}r{2pt}){2-3} \cmidrule(l{2pt}r{2pt}){4-5} \cmidrule(l{2pt}r{2pt}){6-7} \cmidrule(l{2pt}r{2pt}){8-9}
        pretrain & mAP & $\text{mAP}_{\text{50}}$ & mAP & $\text{mAP}_{\text{50}}$ & $\text{IoU}_{\text{fg}}$ & $\text{IoU}_{\text{bg}}$ & $\text{IoU}_{\text{fg}}$ & $\text{IoU}_{\text{bg}}$ \\
        \midrule
        $X$ & 10.8 & 12.7 & 34.7 & 40.0 & 79.1 & 82.0 & 65.3 & 92.6  \\
        \textit{OS} & 23.7 & 27.0 & 20.8 & 23.6 & 79.8 & 82.8 & 67.9 & 93.3 \\
        \bf SimCore & \bf 29.6 & \bf 36.8 & \bf 37.6 & \bf 43.2 & \bf 80.0 & \bf 83.1 & \bf 68.4 & \bf 93.4  \\
    \bottomrule
    \end{tabular}}
    \caption{Object detection and pixel-wise segmentation}
    \label{tab:detection_segmentation}
    \end{subfigure}
    \hfill
    \begin{subfigure}[b]{0.49\textwidth}
    \centering
    \small
    \addtolength{\tabcolsep}{-1pt}
    \resizebox{\textwidth}{!}{
    \begin{tabular}{lcccccccc}
    \toprule
        & \multicolumn{2}{c}{Aircraft} & \multicolumn{2}{c}{Cars} & \multicolumn{4}{c}{Faces} \\
        \cmidrule(l{2pt}r{2pt}){2-3} \cmidrule(l{2pt}r{2pt}){4-5} \cmidrule(l{2pt}r{2pt}){6-9}
        pretrain & mfr. & \!family\! & brand & type & pointy\!\! & oval & \!young\!\! & \!\!smiling\!\!  \\
        \midrule
        $X$ & 21.1 & 40.4 & 67.3 & 78.0 & 66.8 & 83.4 & 93.1 & 93.4 \\
        \textit{OS} & 17.9 & 35.2 & 49.3 & 61.3 & 64.9 & 81.9 & 92.9 & 86.3 \\
        \bf SimCore & \bf 21.9 & \bf 41.9 & \bf 70.7 & \bf 80.1 & \bf 67.5 & \bf 83.9 & \bf 93.6 & \bf 93.7 \\
    \bottomrule
    \end{tabular}}
    \caption{Multi-attribute classification}
    \label{tab:multi-attribute_classification}
    \end{subfigure}
    \vspace{-5pt}
    \caption{Various downstream task performances. For the kNN classifier, we experimented with 20-nearest and 200-nearest neighbors to classify the query images. In the semi-supervised learning setup, we used the label ratio of 10\%, 20\%, and 50\%. mAP is evaluated in the object detection, and IoU of foreground and background are evaluated in the semantic segmentation. In the multi-attribute classification, mfr., pointy, and oval denote manufacturer, pointy nose, and oval face, respectively.}
    \label{tab:downstream_tasks}
    \vspace{-5pt}
\end{table*}

Prior literature has similarly utilized unlabeled or noisy-labeled open-sets, such as open-set semi-supervised learning (OpenSemi)\,\cite{su2021realistic, saito2021openmatch} and webly supervised learning (WeblySup)\,\cite{chen2015webly, sun2021webly}.
OpenSemi and WeblySup work with predefined labels and co-train with the entire huge-scale open-set. 
Our OpenSSL, on the other hand, is a valuable problem itself in that a model can be pretrained without any label information and with efficient subset sampling.
We thus design an experiment under each framework, setting ($X$\,/\,\textit{OS}) as the following example: ($\text{Birds}^{\text{50\%}}$\,/\,$\text{Birds}^{\text{50\%}}$+$\text{WebFG}$) and ($\text{Birds}^{\text{100\%}}$\,/\,$\text{WebFG}^\text{Birds}$), respectively.

Table\,\ref{tab:opensemi_weblysup} summarizes the comparisons with the representative methods of each learning framework.
We observed two findings from Table\,\ref{tab:opensemi_weblysup}. (1)\,When the SimCore model, using WebFG-496 as an open-set, is simply fine-tuned on each target, it outperforms OpenSemi and WeblySup methods. (2)\,SimCore can synergize with both frameworks, serving as an effective initialization.
These results are in line with Su \etal\cite{su2021realistic}, where Self-Training with SSL pretrained model was most preferred.

\subsection{Different Downstream Tasks}

\paragraph{kNN classifier:}
We have shown the linear evaluation performance on each fine-grained dataset to evaluate the quality of learned representation.
Here, we evaluate on another downstream task, nearest neighbor\,(NN) classification.
The NN classification is a nonparametric way to classify test images, making the prediction via weighted voting of nearest neighbors\cite{wu2018unsupervised, caron2020unsupervised}.
Table\,\ref{tab:knn_classification} summarizes the 20 NN and 200 NN classifier results, presenting that SimCore shows the best accuracy, except for Birds where every accuracy was low.

\vspace{-12pt}
\paragraph{Semi-supervised learning:}
In addition, when part of the datasets becomes labeled by expert annotators, we can use those labels to further fine-tune the entire network. Here, we followed the semi-supervised learning protocol in \cite{chen2020simple, grill2020bootstrap}, where the randomly selected samples are annotated and used in the fine-tuning. In Table\,\ref{tab:semi-supervised_learning}, we show the results with three label ratios, each trained with 100 epochs.  
SimCore again showed the best results overall, by particularly large margins in Aircraft and Pet datasets. 
For other semi-supervised learning algorithms, refer to Appendix\,\ref{subsec:additional_finetuning}.

\vspace{-12pt}
\paragraph{Object detection and pixel-wise segmentation:}
For object detection, we used RetinaNet detector\cite{lin2017focal} with ResNet50 encoder and trained for 100 epochs. The metric is mAP evaluated as in\cite{lin2014microsoft}, averaging over 10 IoU thresholds from 0.5 to 0.95 with a step size 0.05. $\text{mAP}_\text{50}$ is the mAP when IoU threshold is 0.5. For pixel-wise segmentation, we used DeepLabV3+ segmentation model\cite{chen2018encoder} and trained for 30 epochs. In Table\,\ref{tab:detection_segmentation}, our SimCore-pretrained encoder could more effectively identify the vehicle objects and segregate the foreground part of pet and bird images.

\vspace{-12pt}
\paragraph{Multi-attribute classification:}
Fine-grained images are often distinguished by multiple attributes. For example, one might ask if the aircraft images could be classified by manufacturers\,(\eg, Boeing), families\,(\eg, Boeing 737), or variants\,(\eg, Boeing 737-700). 
The Cars dataset can also be identified by brands\,(\eg, Audi) or types\,(\eg, Coupe), in addition to the models\,(\eg, Audi TTS Coupe 2012) we used in linear evaluation. 
To this end, we evaluated our pretrained models on multi-attribute classification tasks of three fine-grained datasets, and SimCore excelled the baselines in every task (see Table\,\ref{tab:multi-attribute_classification}).

\section{Conclusion}

In this work, we present a novel OpenSSL problem, where a large-scale open-set can be utilized in self-supervised learning on a fine-grained target dataset. 
Thereafter, we propose SimCore algorithm, which samples the effective coreset from an open-set with semantics similar to the target dataset.
We demonstrate that the coreset samples enhance representation learning for any fine-grained downstream tasks.
We believe that our approach will stimulate more investigation into fine-grained SSL with the open-set in the future.

\vspace{-12pt}
\paragraph{Acknowledgement.}
This work was supported by Institute of Information \& communications Technology Planning \& Evaluation (IITP) grant funded by the Korea government (MSIT) (No. 2019-0-00075, Artificial Intelligence Graduate School Program (KAIST) and No. 2022-0-00871, Development of AI Autonomy and Knowledge Enhancement for AI Agent Collaboration).

{\small
\bibliographystyle{ieee_fullname}
\bibliography{cvpr}
}

\clearpage
\appendix
\onecolumn
\section{Details of Motivating Experiment in Figure \ref{fig:motivation}}
\label{appx:details_motivating_experiments}

\begin{figure}[h]
    \vspace{-0.2in}
    \centering
    \includegraphics[width=0.75\textwidth]{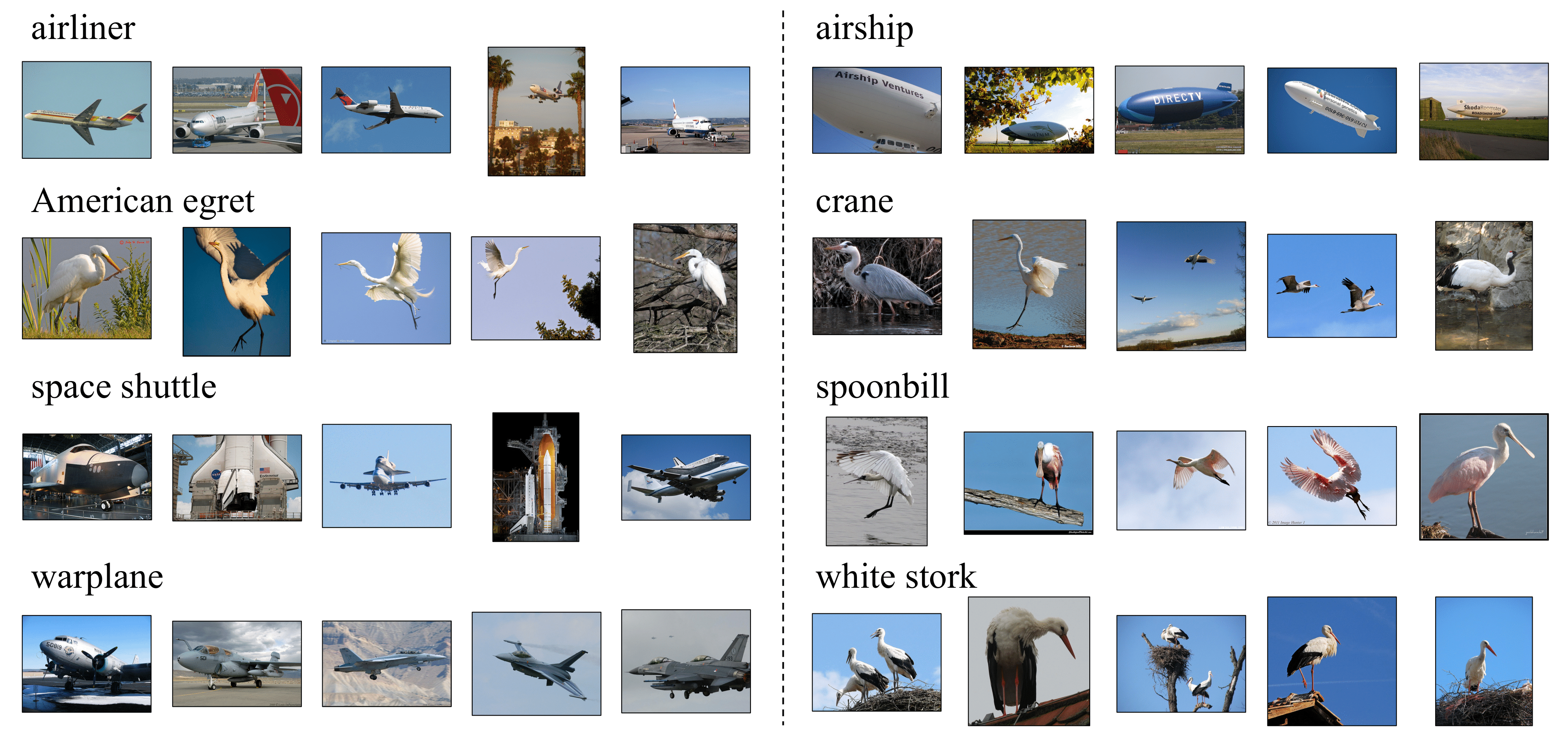}
    \vspace{-5pt}
    \caption{Visualization of examples whose classes belong to $\textit{OS}_\text{oracle}$ ($X$ = FGVC-Aircraft).}
    \label{fig:aircraft_oracle}
\end{figure}

\begin{figure}[h]
    \vspace{-0.1in}
    \centering
    \includegraphics[width=0.75\textwidth]{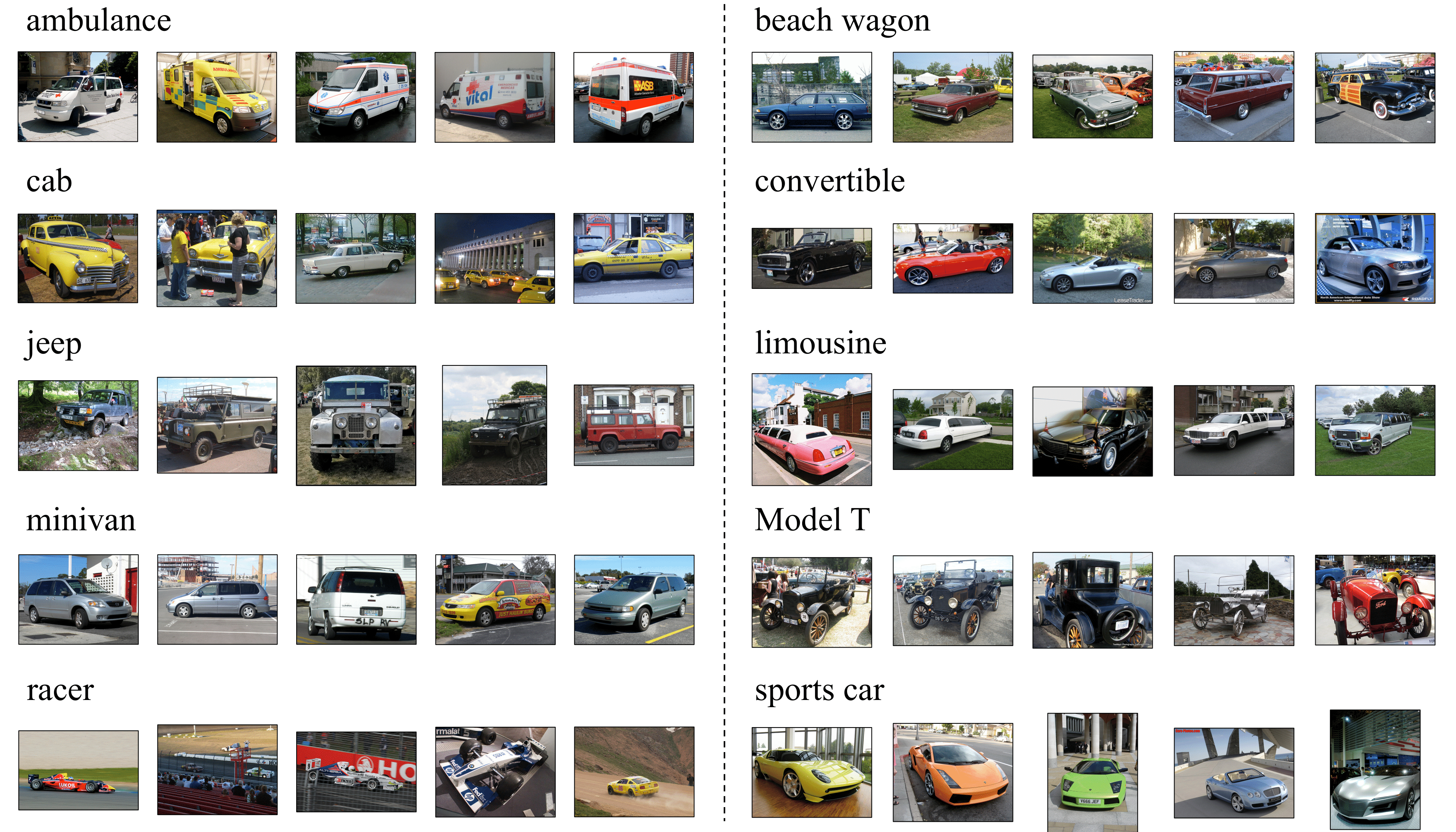}
    \vspace{-5pt}
    \caption{Visualization of examples whose classes belong to $\textit{OS}_\text{oracle}$ ($X$ = Stanford Cars).}
    \label{fig:cars_oracle}
\end{figure}

\begin{figure}[h]
    \centering
    \vspace{30pt}
    \includegraphics[width=.75\textwidth]{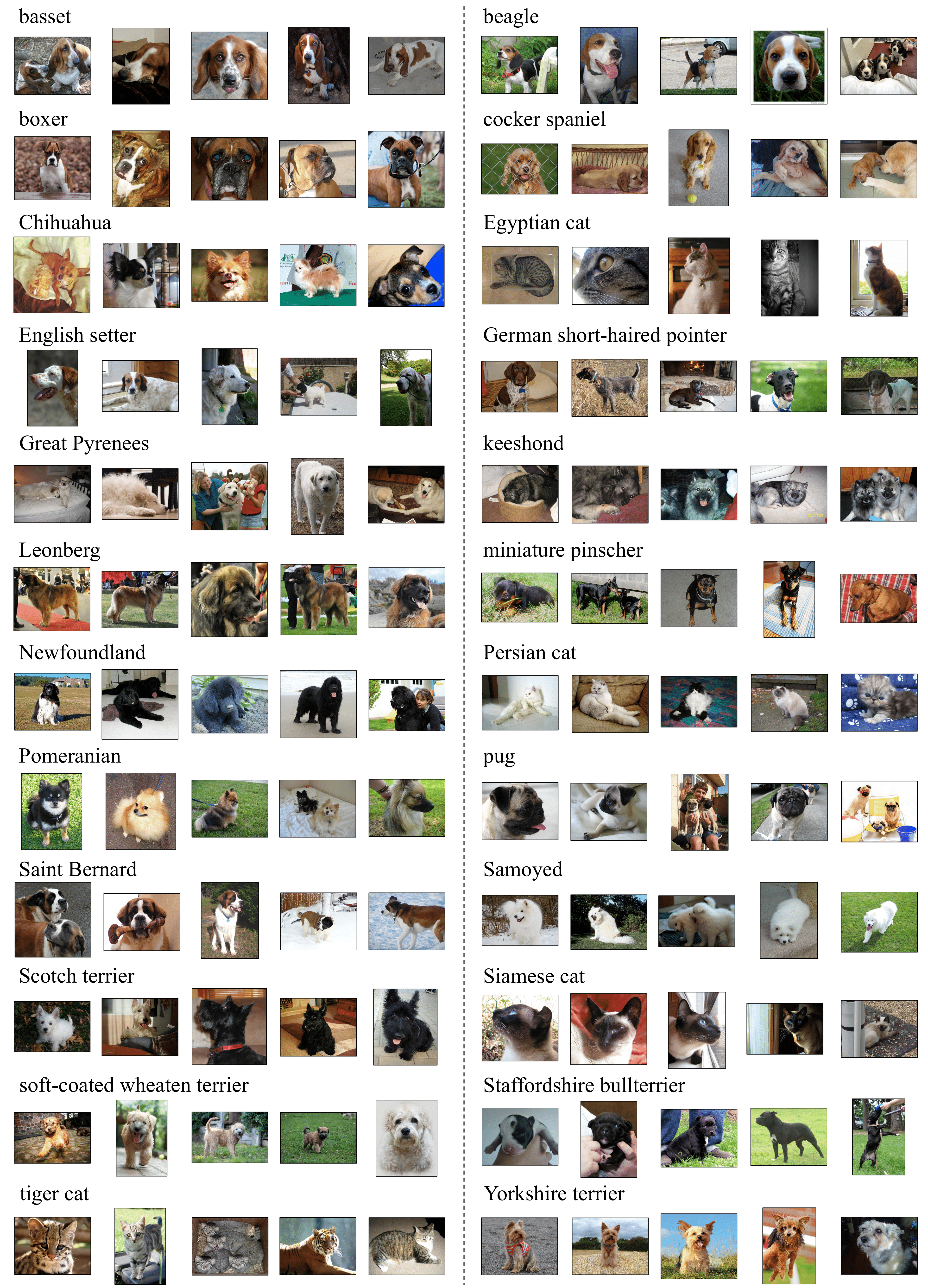}
    \caption{Visualization of examples whose classes belong to $\textit{OS}_\text{oracle}$ ($X$ = Oxford-IIIT Pet).}
    \label{fig:pet_oracle}
\end{figure}

\begin{figure}[h]
    \centering
    \includegraphics[width=.75\textwidth]{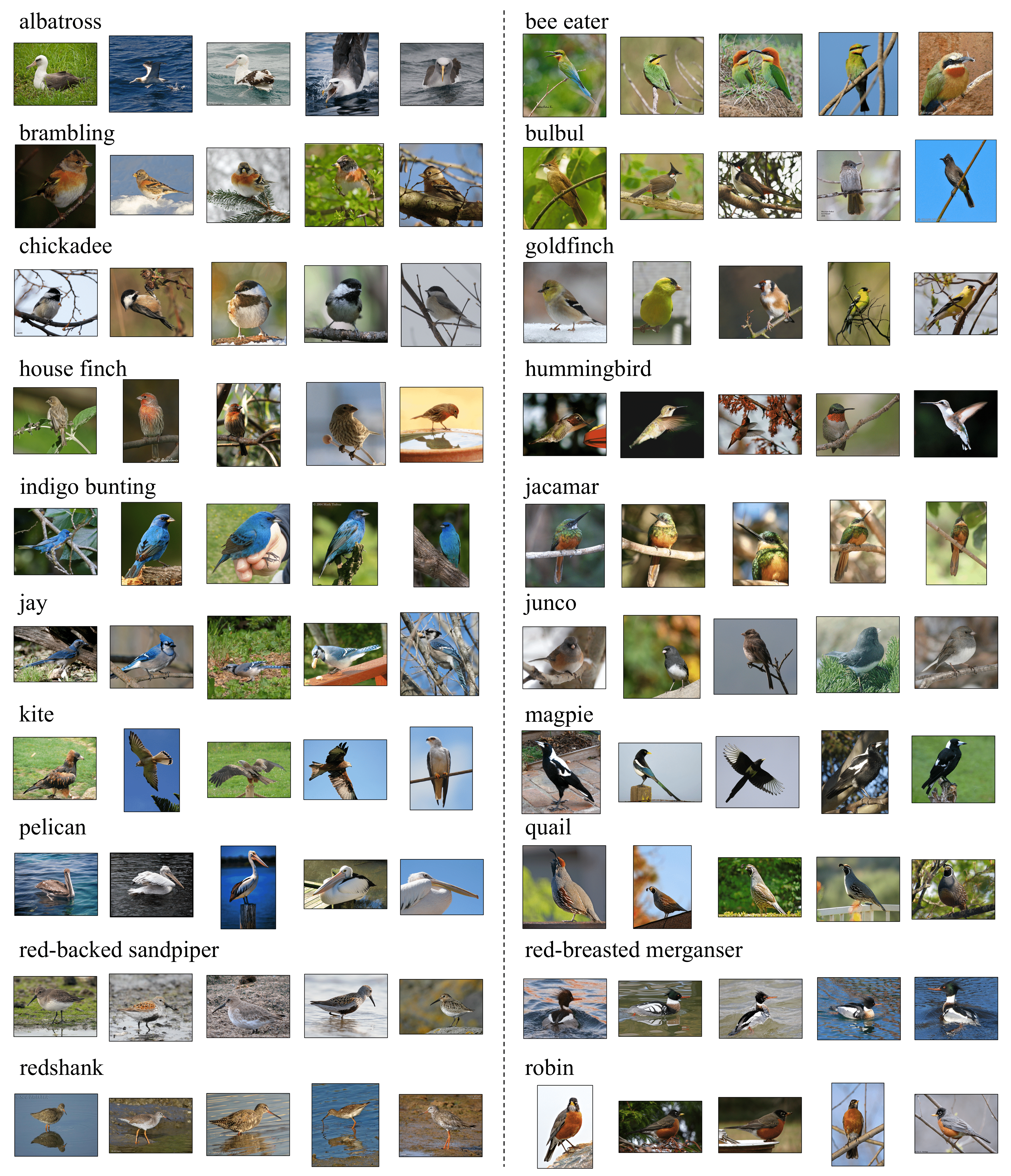}
    \caption{Visualization of examples whose classes belong to $\textit{OS}_\text{oracle}$ ($X$ = Caltech-UCSD Birds).}
    \label{fig:cub_oracle}
\end{figure}

In Table\,\ref{tab:os_oracle}, we described all the selected classes from ImageNet to construct $\textit{OS}_\text{oracle}$, for each target dataset. For clarity, we also visualized examples for each selected category in Figures\,\ref{fig:aircraft_oracle}--\ref{fig:cub_oracle}. Also, Table\,\ref{tab:exact_motivation} summarizes the exact numbers of the motivating experiment results.

\begin{table}[h]
    \small
    \centering
    \resizebox{\textwidth}{!}{
    \renewcommand{\arraystretch}{1.1}
    \begin{tabular}{|l|l|}
        \Xhline{2\arrayrulewidth}
        Target ($X$) & Selected Classes for $\textit{OS}_\text{oracle}$ \\
        \Xhline{2\arrayrulewidth}
        Aircraft & airliner, airship, American egret, crane, space shuttle, spoonbill, warplane, white stork \\
        \hline
        Cars & ambulance, beach wagon, cab, convertible, jeep, limousine, minivan, Model T, racer, sports car \\
        \hline
        \multirow{3}{*}{Pet} &  basset, beagle, boxer, cocker spaniel, Chihuahua, Egyptian cat, English setter, German short-haired pointer, Great Pyrenees, \\
        &  keeshond, Leonberg, miniature pinscher, Newfoundland, Persian cat, Pomeranian, pug, Saint Bernard, Samoyed, \\
        & Scotch terrier, Siamese cat, soft-coated wheaten terrier, Staffordshire bullterrier, tiger cat, Yorkshire terrier \\
        \hline
        \multirow{2}{*}{Birds} & albatross, bee eater, brambling, bulbul, chickadee, goldfinch, house finch, hummingbird, indigo bunting, jacamar, \\
        &  jay, junco, kite, magpie, pelican, quail, red-backed sandpiper, red-breasted merganser, redshank, robin \\
        \Xhline{2\arrayrulewidth}
    \end{tabular}}
    \vspace{-5pt}
    \caption{The class list of $\textit{OS}_\text{oracle}$ according to each target dataset\,($X$).}
    \label{tab:os_oracle}
\end{table}

\clearpage

\begin{table}[!ht]
    \small
    \centering
    \vspace{-5pt}
    \begin{tabular}{lp{20pt}cp{20pt}cp{20pt}cp{20pt}c}
    \toprule
     & \multicolumn{8}{c}{Target dataset ($X$) and its number of samples}  \\
    \cmidrule(l{2pt}r{2pt}){2-9}
    pretraining & \multicolumn{2}{c}{FGVC-Aircraft} & \multicolumn{2}{c}{Stanford Cars} & \multicolumn{2}{c}{Oxford-IIIT Pet} & \multicolumn{2}{c}{\!\!\!\!Caltech-UCSD Birds\!\!} \\
    \midrule
    $X$ & 46.56\% & (6.7K) & 55.42\% & (8.1K) & 59.23\% & (3.7K) & 29.27\% & (6.0K) \\
    \textit{OS} & 41.50\% & (1.3M) & 41.86\% & (1.3M) & 67.66\% & (1.3M) & 33.21\% & (1.3M) \\
    $X$+\textit{OS} & 39.88\% & (1.3M) & 42.92\% & (1.3M) & 68.22\% & (1.3M) & 32.88\% & (1.3M) \\
    \midrule
    $X$+$\textit{OS}_{\text{rand}}$ & 48.24\% & (19.5K) & 49.26\% & (20.9K) & 64.27\% & (16.5K) & 31.90\% & (18.8K) \\
    \bf $X$+$\textit{OS}_{\text{oracle}}$ & \bf 48.78\% & (17.1K) & \bf 57.56\% & (21.1K) & \bf 80.73\% & (34.9K) & \bf 38.26\% & (32.0K) \\
    \bottomrule
    \end{tabular}
    \caption{Linear evaluation performance with sample size for each pretraining dataset. ImageNet is used as the open-set, and the random sampling portion is 1\% of $|$\textit{OS}$|$.}\label{tab:exact_motivation}
\end{table}

\newpage

\section{Experimental Settings}\label{appx:experimental_settings}

We used eleven fine-grained datasets and four open-sets in the main experiments. We summarized the dataset configurations in the following Table\,\ref{tab:exp_setup}.
We followed the linear evaluation protocol used in recent SSL literature\cite{chen2020simple, grill2020bootstrap}: pretraining the encoder and fine-tuning only the classifier with the frozen encoder part. 

\subsection{Datasets}
Since the SSL pretraining does not rely on any label information, we incorporate the train and validation set of the original benchmarks, such as FGVC-Aircraft, Oxford 102 Flower, Describable Textures, and Food 11, to enlarge the number of training samples. Besides, in the case of CelebAMask-HQ, we exclude identities that contain less than 15 images. The original number of 30,000 images thus have reduced to 4,263, and the number of identities have reduced from 6,217 to 307. 

\begin{table}[!h]
    \centering
    \small
    \begin{tabular}{l|ccc}
        \toprule
        Dataset & Train\,\# & Test\,\# & Class\,\#  \\
        \midrule
        ImageNet\cite{deng2009imagenet} & 1,281,167 & 50,000 & 1,000  \\ 
        Microsoft COCO\cite{lin2014microsoft} & 118,287 & 5,000 & 80 \\
        iNaturalist 2021-mini\cite{van2018inaturalist} & 500,000 & 100,000 & 10,000 \\
        Places365\cite{zhou2017places} & 8,026,628 & 328,500 & 365 \\
        \hline
        FGVC-Aircraft\cite{aircraft} & 6,667 & 3,333 & 100 \\ 
        Stanford Cars\cite{cars} & 8,144 & 8,041 & 196  \\ 
        Oxford-IIIT Pet\cite{pet} & 3,680 & 3,669 & 37  \\ 
        Caltech-UCSD Birds\cite{birds} & 5,990 & 5,790 & 200  \\ 
        Stanford Dogs\cite{khosla2011novel} & 12,000 & 8,580 & 120  \\ 
        Oxford 102 Flower\cite{nilsback2008automated} & 2,040 & 6,149 & 102 \\
        Stanford 40 Actions\cite{yao2011human} & 4,000 & 5,532 & 40 \\ 
        MIT-67 Indoor Scene Recognition\cite{quattoni2009recognizing} & 5,360 & 1,340 & 67 \\ 
        Describable Textures\,(DTD)\cite{cimpoi2014describing} & 3,760 & 1,880 & 47 \\
        CelebAMask-HQ\cite{lee2020maskgan} & 4,263 & 1,215 & 307 \\
        Food 11\cite{singla2016food} & 13,296 & 3,347 & 11 \\
        \bottomrule
    \end{tabular}
    \caption{Datasets we used and their configurations. The upper part corresponds to open-sets and the lower part corresponds to fine-grained target datasets. Although every open-set contains its test set and designated classes, we have only used an unlabeled train set.}
    \label{tab:exp_setup}
\end{table}

\vspace{-10pt}
\subsection{Encoder Pretraining}

\paragraph{Basic settings:}
We set the cost of pretraining to 8 GPU days (4 GPUs $\times$ 2 days) since the train set sizes vary depending on the pretraining dataset.
For example, we pretrained the model for longer epochs of 5K on small-scale fine-grained datasets, because one epoch is a few hundred times shorter in iteration than an epoch of \textit{OS}.
Meanwhile, although it took more than 8 GPU days, we trained 200 epochs for ImageNet open-set\,(\textit{OS}) and $X$+\textit{OS} experiments for full pretraining.

For SimCore pretraining, every 1\% sampling last 5K epochs, but every 5\% sampling last 2K epochs due to the excess of 8 GPU days.
Since SimCore with a stopping criterion had different sampling ratios according to the target dataset, in this case, we trained the model for $\min(\text{5K}, \text{10K}/p)$ epochs. If the open-set changes, we increased or decreased the epochs by the ratio between the open-set size and ImageNet size, \ie, $X$+$\text{Places365}_{\text{SimCore}}$ uses $\times 6.3$ shorter epochs than $X$+$\text{ImageNet}_{\text{SimCore}}$.
Also, when using SimCore with a stopping criterion, a budget limit is not necessarily needed. Nonetheless, in case there are sufficiently large number of core-samples in the open-set, we set the budget to $\times$50 of the target dataset size to see the effect of a small coreset.
Note that we used the stopping threshold $\tau$ as 0.95 in default.

\vspace{-12pt}
\paragraph{SimCLR method:}
When pretraining the encoder with the SimCLR method, SGD optimizer is used with an initial learning rate of 0.1 and $\ell_2$ regularization parameter 1e-4, with 512 batch size. The learning rate was scheduled by cosine annealing\cite{cosine_schedule}, decayed to zero on the last epoch.
We followed the same hyperparameters, the temperature of 0.07 and the projection dimension of 128, as in the original paper\cite{chen2020simple}. For data augmentation, we used a random resized crop, horizontal flip, color jitter, and random gray-scale.

\vspace{-12pt}
\paragraph{BYOL method:}
We used an Adam optimizer\cite{kingma2014adam} with a learning rate of 1e-3 and $\ell_2$ regularization parameter 1e-6. BYOL method is highly sensitive to the exponential moving average\,(EMA) value of a momentum encoder\cite{he2020momentum}. Therefore, we followed the same EMA scheduler as in the original implementation\cite{grill2020bootstrap}, of which the momentum starts from 0.996 to 1. We should also note that we did not use the EMA scheduler for the experiments with short epochs.

\vspace{-12pt}
\paragraph{SwAV method:}
We mostly followed the same settings as the SimCLR method, such as an SGD optimizer with a learning rate of 0.1 and weight decay of 1e-4. In the case of pretraining on ImageNet, we used the temperature value of 0.1 and an epsilon value of 0.02, while we froze the 3,000-dimensional prototypes for one epoch. For every other cases, we froze the 100-dimensional prototypes for ten epochs.

\vspace{-12pt}
\paragraph{DINO method:}
We mostly followed the pretraining setups as in the original paper\cite{caron2021emerging}. We used an AdamW optimizer\,\cite{loshchilov2017decoupled} with a learning rate of 1e-3. Cosine annealing and warmup were used during the 2\% of total epochs.
We scheduled an $\ell_2$ regularization parameter from 0.04 to 0.4, and the EMA momentum was increased from 0.996 to 1.0. Note that the $\ell_2$ regularization had not been applied to any biases and normalization parameters.
Besides, we scheduled the teacher temperature from 0.04 to 0.07 for the 40\% of total epochs and set the student temperature as 0.1.
The center momentum was 0.9, and we set the dimension of projector as 65,536. As a data augmentation technique, we used additional four local croppings\cite{caron2020unsupervised} (96$\times$96). The encoder's last layer was frozen during the first 10 epochs.
These numerous hyperparameters were highly fit to the ImageNet pretraining, whereas they did not seem effective in several fine-grained datasets.
In practice, the stopping threshold $\tau$ of 0.6 was optimal for the DINO method, which can be attributed to the large dimension of the projector.

\vspace{-12pt}
\paragraph{MAE method:}
We used an AdamW optimizer with a learning rate of 3e-4, cosine annealing, and warmup epochs of 100. Furthermore, we used the constant $\ell_2$ regularization parameter of 0.05. As in the original paper\cite{he2022masked}, we set the mask ratio as 75\%, and we normalized the pixel values when calculating the reconstruction loss.

\subsection{Fine-Tuning for Linear Evaluation}
We fine-tuned a linear classifier for 100 epochs and searched the optimal learning rate among five logarithmically spaced values from $1$ to $10^2$. We decayed the learning rate by 0.1 at 60 and 80 epochs, and any regularization techniques, such as weight decay, were not used.

\subsection{Open-Set Semi-Supervised and Webly Supervised Learning}

\paragraph{SimCore fine-tuning:}
For SimCore fine-tuning baselines, we fine-tuned the SimCore pretrained models for 200 epochs with the batch size of 256. We used an SGD optimizer with $\ell_2$ regularization parameter of 1e-4. The learning rate started from the value of 3e-2 and was decayed by a cosine annealing scheduler.

\vspace{-12pt}
\paragraph{OpenSemi framework:}
Under the open-set semi-supervised learning (OpenSemi) framework, we used an SGD optimizer with a learning rate of 3e-2, $\ell_2$ regularization parameter of 1e-4, and cosine annealing.
In default, we trained the random initialized model for 128 epochs with iteration steps of 512 and the batch size of 64. For the models pretrained by SimCore, we used the shorter epochs of 32.

For the Self-Training algorithm\,\cite{su2021realistic}, we used the temperature of 1.0 and set the coefficient of the unlabeled loss as 0.5.
We trained a teacher network from scratch for 500 epochs and fine-tuned the SimCore initialized model for 100 epochs.

For the OpenMatch algorithm\,\cite{saito2021openmatch}, we set the coefficient of open-set entropy minimization (OEM) loss and soft open-set consistency regularization (SOCR) loss as 0.1 and 0.5, respectively.
Besides, we set the epochs to start FixMatch training as 20 for random initialized models and 5 for SimCore pretrained models.

\vspace{-12pt}
\paragraph{WeblySup framework:}
We used an SGD optimizer with a learning rate of 3e-2, $\ell_2$ regularization parameter of 1e-4, and cosine annealing.
For the Co-Teaching algorithm\,\cite{han2018co}, we used the batch size of 256 and the forget ratio of 0.2. In addition, we followed the values of other hyperparameters as in Han \etal. We set the epochs as 1,000 for training from scratch, and 200 for the experiments of SimCore initialization. Note that we modified the learning rate and forget ratio to 0.1 for Cars dataset.

For the DivideMix method\,\cite{li2020dividemix}, we trained the model with 400 epochs and the batch size of 128. We set the probability threshold as 0.2 and warmup epochs as 30 for every dataset. Since the DivideMix algorithm includes the training schemes of MixMatch\cite{berthelot2019mixmatch}, we used the default hyperparameters of the MixMatch algorithm, such as the unlabeled loss coefficient of 75 and alpha value of 0.75.

\subsection{Downstream Tasks}
For the object detection task, we used an Adam optimizer with a learning rate of 1e-4 and the batch size of 16. Besides, we set the total epochs and IoU threshold of non-maximum suppression as 30 and 0.2, respectively. We should note that the backbone network of DeepLabV3+ is frozen during fine-tuning, and we trained an FPN network from scratch.

For the semantic segmentation task, we have also frozen the SSL pretrained ResNet-50 encoder of RetinaNet. We used an SGD optimizer with $\ell_2$ regularization parameter of 1e-4 and the Nesterov momentum. The optimal learning rate is chosen among five logarithmically spaced values from 1e-1 to 1e-3.

\newpage

\section{Sampling Ratios of SimCore Experiments}

The coreset denotes a subset of the open-set that is semantically similar to the target dataset.
Since we measure this similarity on the feature space generated by a retrieval model, the sampling ratio with a stopping criterion could vary depending on several factors, such as model architectures, SSL losses, open-set, or target dataset.
That is, an image pair can be recognized as similar or dissimilar according to how they are represented.

The exact values of sampling ratios in our experiments are summarized in Tables\,\ref{tab:sampling_ratio}--\ref{tab:sampling_ratio_openset}.
In effect, SimCLR with any architecture samples less than 1\% as the coreset of Cars, but BYOL with ResNet50 samples over 30\% of the open-set.
Although the proper sampling ratios were different across each method, SimCore has consistently improved performances with the selected samples. 

\begin{table}[!h]
    \small
    \centering
    \begin{tabular}{llcccc}
    \toprule
    method & architecture & \!\!Aircraft\!\! & Cars & Pet & Birds \\
    \midrule
    SimCLR & EfficientNet & 0.59\% & 0.37\% & 2.16\% & 1.16\% \\
    SimCLR & ResNet18 & 0.31\% & 0.35\% & 0.96\% & 1.09\% \\
    SimCLR & ResNeXt50 & 0.79\% & 0.96\% & 14.36\% & 13.17\% \\
    SimCLR & ResNet101 & 0.64\% & 0.65\% & 8.51\% & 11.37\% \\
    \midrule
    BYOL & ResNet50 & 4.78\% & 31.78\% & 14.36\% & 23.38\% \\
    SwAV & ResNet50 & 26.02\% & 31.78\% & 14.36\% & 23.38\% \\
    DINO & ViT-Ti/16 & 12.19\% & 2.51\% & 14.36\% & 21.41\% \\
    MAE & ViT-B/16 & 3.23\% & 1.11\% & 8.47\% & 5.69\% \\
    \bottomrule
    \end{tabular}
    \vspace{-5pt}
    \caption{Sampling ratios of the SimCore experiments in Table\,\ref{tab:different_arch} and Table\,\ref{tab:different_ssl}.}
    \label{tab:sampling_ratio}
\end{table}

Furthermore, SimCore samples the reasonable amount with any open-set samples, \eg, sampling number of iNaturalist\,(32K) vs. Places365\,(268K) in Indoor. Along with Figure\,\ref{fig:diff_openset} and Table\,\ref{tab:uncurated_openset}, the performance gain of SimCore is correlated with the semantic similarity between $X$ and \textit{OS}, suggesting the benefit of suitable open-sets.

\begin{table}[!h]
    \small
    \centering
    \begin{tabular}{lcccccc}
    \toprule
    \textit{OS} for SimCore & Pet & Birds & Action & Indoor & Aircraft & Cars \\
    \midrule
    COCO (0.12M) & 29.07\% & 29.60\% & 59.88\% & 37.49\% & - & - \\
    iNaturalist (0.5M) & 22.05\% & 13.03\% & 21.87\% & 6.31\% & - & - \\
    Places365 (8M) & 2.29\% & 3.73\% & 2.49\% & 3.34\% & - & - \\
    ImageNet (1.3M) & 14.36\% & 13.68\% & 15.61\% & 13.46\% & 1.03\% & 0.95\%  \\
    \midrule
    \texttt{ALL} (9.9M) & 1.85\% & 3.02\% & 2.01\% & 2.70\% & 0.20\% & 0.43\% \\
    WebVision (2.4M) & 7.52\% & 12.24\% & 8.18\% & 10.96\% & 0.39\% & 0.76\% \\
    WebFG-496 (0.05M) & - & 34.47\% & - & - & 25.32\% & 40.21\% \\
    \bottomrule
    \end{tabular}
    \vspace{-5pt}
    \caption{Sampling ratios of the SimCore experiments in Figure\,\ref{fig:diff_openset} and Table\,\ref{tab:uncurated_openset}. The ratios are calculated based on the size of each open-set.}
    \label{tab:sampling_ratio_openset}
\end{table}

\section{Implementation Details of Feature Distribution Analysis}
\label{appx:implementation_details_feat_dist}

Figure\,\ref{fig:feat_dist} in Section \ref{subsec:analysis_of_simcore} visualizes feature distribution of each pretraining dataset. We extracted the feature embeddings with the retrieval model (\ie, pretrained encoder on the target dataset for short epochs). Then, we reduced all representations to two-dimensional vectors via t-SNE\cite{van2008visualizing}. These representation vectors are distributed on a unit ring by Gaussian kernel density estimation, as introduced in \cite{wang2020understanding}. 

In detail, we visualized the feature distributions of \textit{OS}, $X$, and the coreset sampled from \textit{OS}. For the visualization of \textit{OS}, ImageNet in this case, we randomly selected 10\% of \textit{OS} for faster convergence of t-SNE. For the coreset, SimCore samples 1\% of \textit{OS}.
Unlike in Aircraft and Cars, in Pet and Birds, the distribution of $X$ features are more overlapped with that of \textit{OS} features. This implies high semantic similarity between the open-set\,(ImageNet) and the target dataset\,(Pet or Birds). 

\newpage
\section{Sensitivity Study}

\subsection{Stopping Threshold}

The stopping criterion in SimCore algorithm requires a threshold value $\tau$. In default, we used $\tau=0.95$ throughout the experiments.
Here, we investigate the sensitivity of SimCore performance to its stopping threshold value. 
In Table\,\ref{tab:sensitivity_threshold}, we summarized both sampling ratios and classification accuracies by varying the stopping threshold.
We could observe that the $\tau$ value of $0.95$ generally performs well in both SimCLR and MAE.

\begin{table}[!h]
    \small
    \centering
    \addtolength{\tabcolsep}{-1pt}
    \begin{tabular}{lcccccccccccc}
    \toprule
    & \multicolumn{6}{c}{SimCLR with ResNet50} & \multicolumn{6}{c}{MAE with ViT-B/16} \\
    \cmidrule(l{2pt}r{4pt}){2-7}\cmidrule(l{4pt}r{2pt}){8-13}
    target & \multicolumn{2}{c}{$\tau=0.99$} & \multicolumn{2}{c}{$\mathbf{\boldsymbol{\tau}=0.95}$} & \multicolumn{2}{c}{$\tau=0.9$} & \multicolumn{2}{c}{${\tau=0.99}$} & \multicolumn{2}{c}{$\mathbf{\boldsymbol{\tau}=0.95}$} & \multicolumn{2}{c}{$\tau=0.9$} \\
    \midrule
    Aircraft & \textcolor{gray}{0.21\%} & 46.7 & \textcolor{gray}{1.03\%} & 48.3 & \textcolor{gray}{5.27\%} & 47.6 & \textcolor{gray}{0.26\%} & 49.5 & \textcolor{gray}{3.23\%} & 48.1 & \textcolor{gray}{17.18\%} & 52.9 \\
    Cars & \textcolor{gray}{0.24\%} & 56.5 & \textcolor{gray}{0.95\%} & 60.3 & \textcolor{gray}{4.52\%} & 52.5 & \textcolor{gray}{0.29\%} & 62.1 & \textcolor{gray}{1.11\%} & 52.4 & \textcolor{gray}{5.22\%} & 62.9 \\
    Pet & \textcolor{gray}{1.96\%} & 79.8 & \textcolor{gray}{14.36\%} & 79.7 & \textcolor{gray}{14.36\%} & 79.7 & \textcolor{gray}{0.45\%} & 52.5 & \textcolor{gray}{8.47\%} & 77.8 & \textcolor{gray}{14.36\%} & 75.8 \\
    Birds & \textcolor{gray}{0.71\%} & 36.2 & \textcolor{gray}{13.68\%} & 37.7 & \textcolor{gray}{23.38\%} & 37.7 & \textcolor{gray}{0.43\%} & 31.1 & \textcolor{gray}{5.69\%} & 42.1 & \textcolor{gray}{23.38\%} & 35.3 \\
    \bottomrule
    \end{tabular}
    \vspace{-5pt}
    \caption{Sensitivity study to the stopping threshold of SimCore. For each threshold value, \textcolor{gray}{first column} indicates the sampling ratio according to the threshold, and second column indicates the corresponding linear evaluation accuracy.}
    \label{tab:sensitivity_threshold}
\end{table}

\subsection{Retrieval Model}
\begin{figure}[!h]
\vspace{-3pt}
\centering
\begin{subfigure}[b]{0.4\linewidth}
    \centering
    \includegraphics[width=\linewidth]{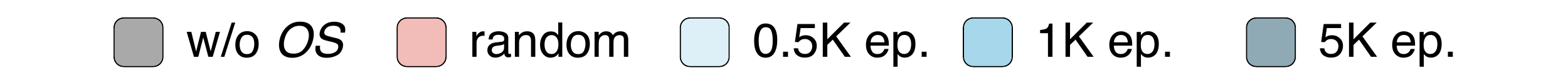}
    \vspace{-12pt}
\end{subfigure}

\begin{subfigure}[b]{0.22\linewidth}
    \centering
    \includegraphics[width=\textwidth]{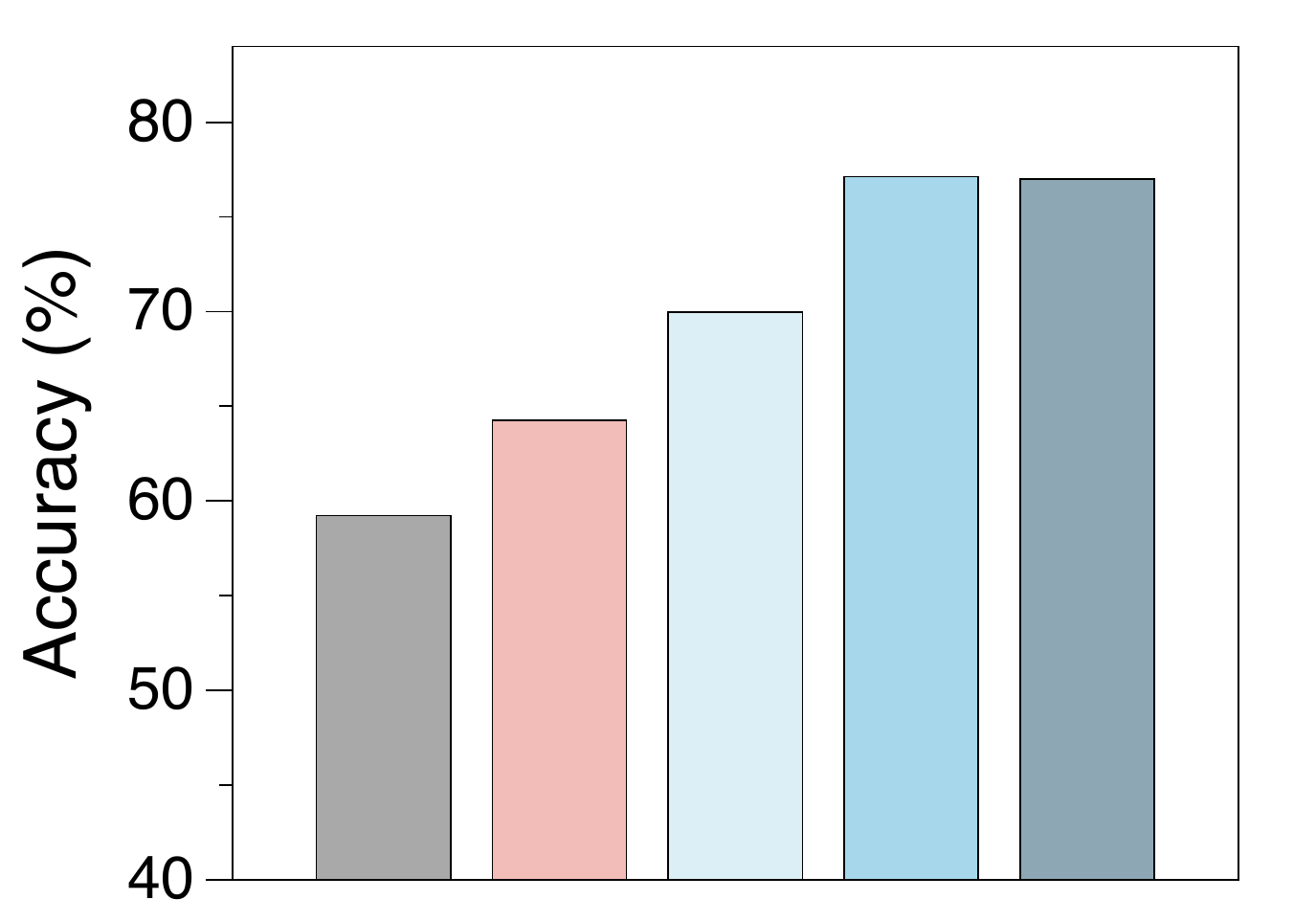}
    \vspace{-15pt}
    \caption{Pet}
\end{subfigure}
\begin{subfigure}[b]{0.22\linewidth}
    \centering
    \includegraphics[width=\textwidth]{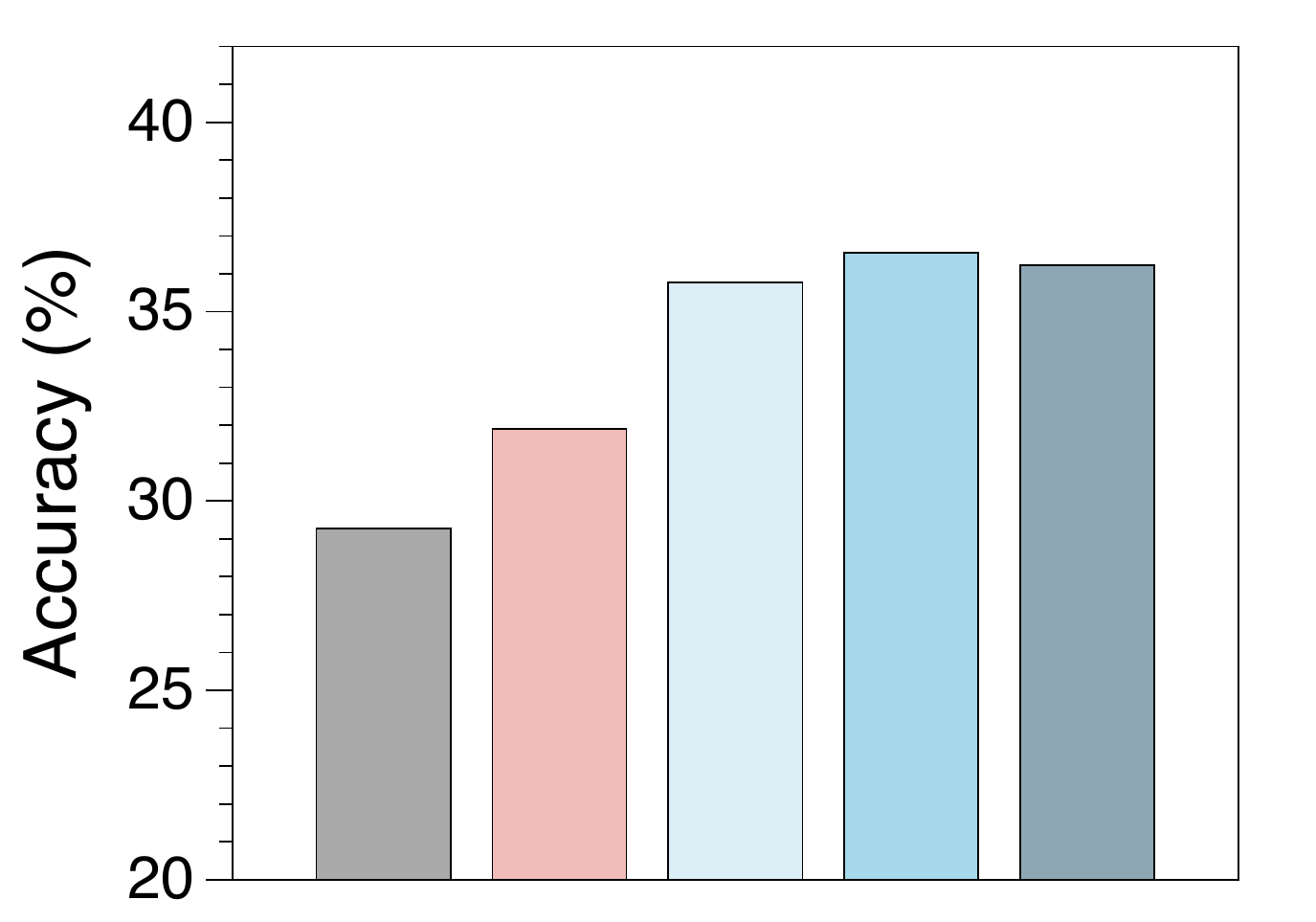}
    \vspace{-15pt}
    \caption{Birds}
\end{subfigure}
\vspace{-7pt}
\caption{Performance comparisons with different pretraining epochs of retrieval models. ``w/o OS'': training on $X$ for 5K epochs. ``random'': random sampling followed by 5K epochs training on $X$+$\textit{OS}_\text{rand}$. ``$N$ ep.'': $N$ epochs pretraining on $X$ before the coreset sampling, followed by 5K epochs training on $X$+$\textit{OS}_\text{SimCore}$.}
\label{fig:pretrain_epoch}
\end{figure}

Prior to the coreset sampling, we pretrain the encoder on a target dataset for short epochs, 1K in our experiments. We refer this model, used in the coreset selection, to a \emph{retrieval model}. This is necessary because we should measure the similarity between the features from the target dataset and from the open-set. Figure\,\ref{fig:pretrain_epoch} presents SimCore performances, given different pretraining epochs of retrieval models. For comparison, the sampling ratio is set to $p=1\%$, and the random sampling strategy is also included.

As a result, 0.5K ep. model performed marginally worse than other SimCore models, whereas 1K ep. was comparable to 5K ep. model. Still, every SimCore models outperformed random sampling or without \textit{OS}. We should note that 1K ep. pretraining takes up a small portion of the entire training process. On the basis of Pet, 1K pretraining of the retrieval model only costs 4.3\% of the total iterations.

However, one can still have a concern on the training cost for the retrieval model. To address this, we conducted an additional experiment, where the models are pretrained on only target datasets for the same number of iterations as our SimCore with the sampling ratio of 1\%.
For example, we trained the models on Pet for 18,925 epochs to exactly match the iterations to SimCore 1\% experiment.
Interestingly, we obtained the results as follows, compared to SimCore: 52.26\% in Aircraft ($+$3.81\%), 52.97\% in Cars ($-$6.03\%), 60.80\% in Pet ($-$16.33\%), and 30.59\% in Birds ($-$5.97\%).
Based on these results, we have confirmed the efficacy of the sampled coreset, as simply increasing the pretraining epochs does not result in the performance improvement.

\newpage
\subsection{Number of Clusters}
\begin{figure}[!h]
\vspace{-3pt}
    \centering
    \begin{subfigure}[b]{0.22\linewidth}
        \centering
        \includegraphics[width=\linewidth]{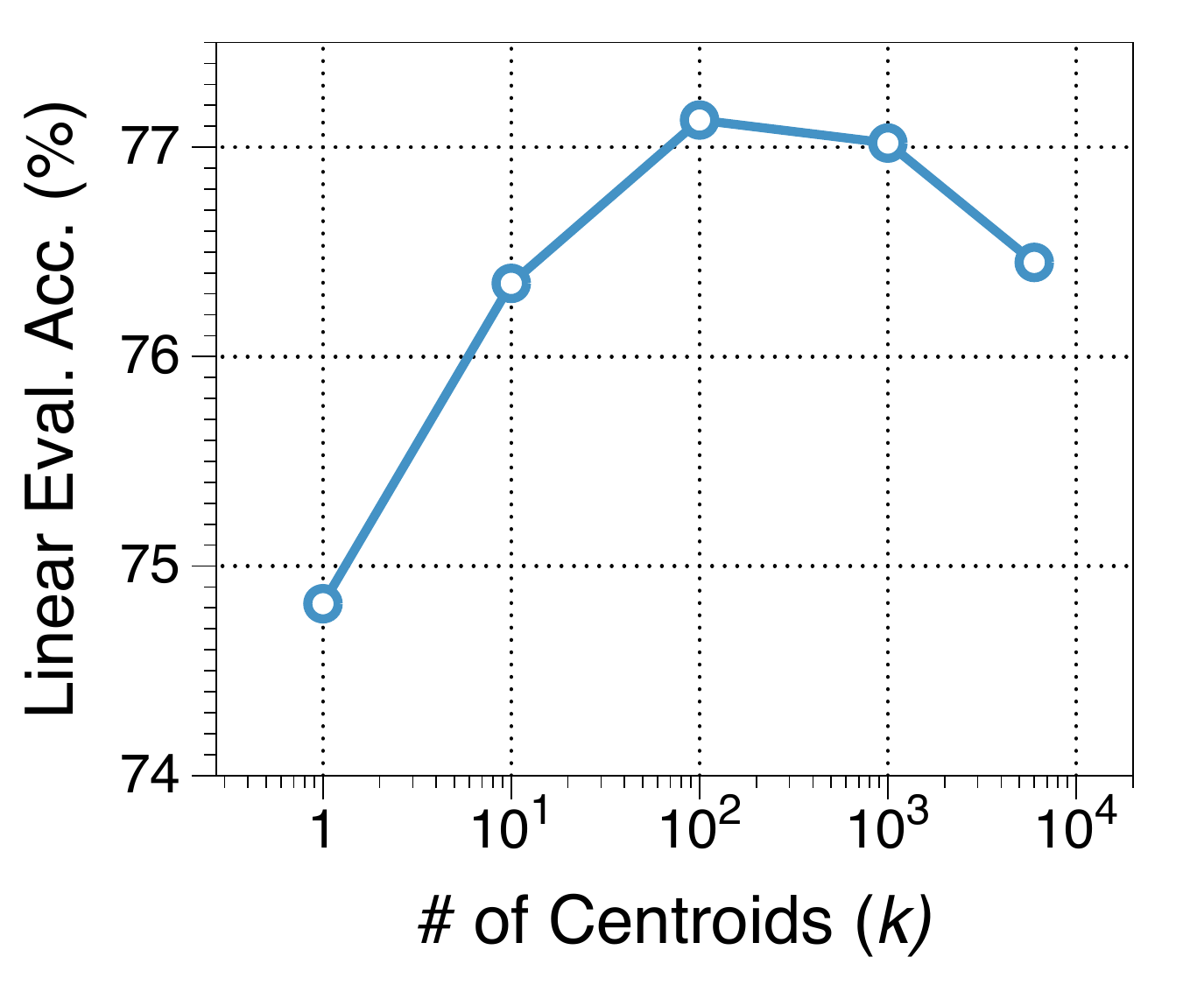}
        \caption{Pet: $|X|$ = 3,680}
        \vspace{-5pt}
        \label{fig:k_sensitivity_pet}
    \end{subfigure}
    \begin{subfigure}[b]{0.22\linewidth}
        \centering
        \includegraphics[width=\linewidth]{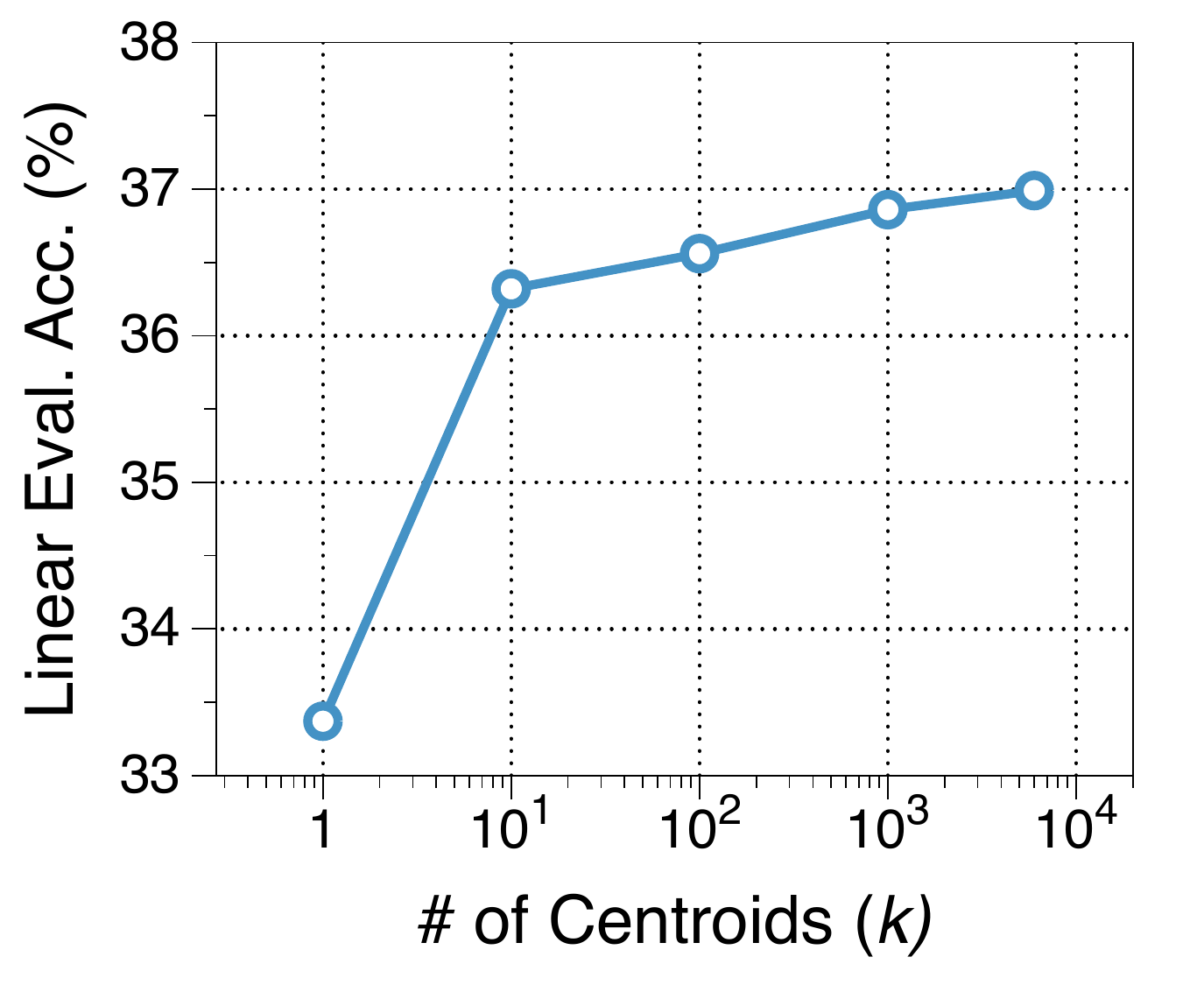}
        \caption{Birds: $|X|$ = 5,990}
        \vspace{-5pt}
        \label{fig:k_sensitivity_birds}
    \end{subfigure}
    \caption{Sensitivity study of the SimCore performance according to the number of centroids. The maximum feasible centroid number is 3,680 and 5,990 for Pet and Birds, respectively.}
    \label{fig:k_sensitivity}
\end{figure}

To reduce the complexity, we have used 100 centroids to calculate the $\hat{f}(\mathcal{S})$ value after $k$-means clustering. Here, we show that SimCore is robust to the number of $k$. Figure\,\ref{fig:k_sensitivity} shows SimCore performance results according to different $k$ values, \{$1$, $10$, $10^2$, $10^3$, $|X|$\}, with 1\% coreset sampling from ImageNet. Except for the single centroid case, SimCore exhibited overall high accuracies.

\section{Additional Experiments}
\subsection{Two-Stage SSL Pretraining}

If a good initialization pretrained on \textit{OS} is available, we could further pretrain this model on either $X$\,($\textit{OS} \rightarrow X$) or the union of $X$ and the coreset\,($\textit{OS} \rightarrow X+\textit{OS}_\text{SimCore}$).
Therefore, we additionally experimented with the checkpoint of the official code of SimCLR\,\cite{chen2020simple}, which is already pretrained on the ImageNet dataset. Here, we used an Adam optimizer with a learning rate of 1e-3 and reduced the epochs to 20\% compared to training from scratch.
Table\,\ref{tab:two_stage_ssl} summarizes the evaluation results of linear probing for three pretraining schemes.
In practice, our SimCore outperforms baselines in 10 out of 11 fine-grained datasets.
This demonstrates that a well-pretrained model can also be effectively exploited through two-stage pretraining schemes with the SimCore algorithm.

\begin{table*}[!h]
    \small
    \centering
    \begin{tabular}{lccccccccccc}
    \toprule
    pretrain & \!\!Aircraft\!\! & Cars & Pet & Birds & Dogs & \!Flowers\!\! & Action & Indoor & \!\!Textures\! & Faces & Food \\
    \midrule
    \textit{OS} & 41.90 & 45.79 & 78.35 & 41.95 & 61.32 & 89.67 & 68.67 & 68.04 & 71.81 & 53.91 & 88.05 \\
    $\textit{OS}\,\rightarrow\,X$ & 55.77 & 53.07 & 77.46 & 39.97 & 61.53 & 91.29 & 66.25 & 72.02 & 72.66 & \textbf{61.48} & 92.41 \\
    $\textit{OS}\,\rightarrow\,X+\textit{OS}_\text{SimCore}$ & \textbf{55.83} & \textbf{57.03} & \textbf{84.10} & \textbf{44.99} & \textbf{68.44} & \textbf{91.45} & \textbf{72.31} & \textbf{75.20} & \textbf{73.19} & 59.09 & \textbf{92.68} \\
    \bottomrule
    \end{tabular}
    \vspace{-5pt}
    \caption{Comparisons between different pretraining schemes on eleven fine-grained datasets. The models are all initialized from the SimCLR's checkpoint that is already pretrained on ImageNet.}
    \label{tab:two_stage_ssl}
\end{table*}

\subsection{Sampling Strategy}
In our default SimCore sampling, we sampled the coreset only once after the training of the retrieval model. In practice, the sampling procedure only takes $\sim$20 minutes, mostly for computing the pairwise similarity (note that SSL pretraining takes $\sim$2 days). There are several available variations on this sampling strategy. One na\"ive way is to sample a coreset several times during the pretraining, since the model evolves and thus retrieves different coresets throughout the training.

In this sense, we resampled the coreset ($p=1\%$) three times during 5K epochs training, while maintaining the entire cost for pretraining. 
As a result, this sampling strategy yielded the performances of 49.99\% in Aircraft ($+$1.54\%), 57.63\% in Cars ($-$1.37\%), 79.75\% in Pet ($+$2.62\%), and 38.08\% in Birds ($+$1.52\%). Thus, exploring the coreset sampling strategies may be a future work that potentially improves the performance of SimCore.

\newpage
\subsection{Additional Fine-Tuning Schemes}
\label{subsec:additional_finetuning}

\paragraph{Semi-supervised learning:}

Table\,\ref{tab:semisup_finetuning} presents the experimental results on fine-tuning using various semi-supervised learning methods, including MixMatch\cite{berthelot2019mixmatch}, ReMixMatch\cite{berthelotremixmatch}, FixMatch\cite{sohn2020fixmatch}, and FlexMatch\cite{zhang2021flexmatch}.
We would note that the semi-supervised learning approach in Table\,\ref{tab:semi-supervised_learning} has simply fine-tuned models with a small portion of labeled data, following the learning protocols described in \cite{chen2020simple, grill2020bootstrap}.
In terms of implementation details, we used the default hyperparameter settings for each method and fine-tuned models for 32 epochs, with iteration steps of 512. In most cases, we found that our SimCore pretraining approach showed the significant performance improvements, regardless of semi-supervised learning methods.
This suggests that a self-supervised pretraining with coreset selection, followed by the use of semi-supervised algorithms, can be an effective approach when only a small amount of labeled data is available.

\begin{table}[!h]
    \vspace{5pt}
    \small
    \centering
    \begin{tabular}{llcccc|llcccc}
    \toprule
    method & pretrain & \!\!Aircraft\!\! & Cars & Pet & Birds & method & pretrain & \!\!Aircraft\!\! & Cars & Pet & Birds \\
    \midrule
    MixMatch & $X$ & 37.2 & 38.1 & 62.0 & 21.9 & FixMatch & $X$ & 30.4 & \textbf{33.8} & 52.8 & 13.7 \\
    MixMatch & \textit{OS} & 44.3 & 30.5 & 73.7 & 24.4 & FixMatch &  \textit{OS} & 25.1 & 21.4 & 63.8 & 13.3 \\
    MixMatch & \bf SimCore &  \textbf{45.1} & \textbf{38.6} & \textbf{74.3} & \textbf{26.0} & FixMatch &  \bf SimCore & \textbf{31.2} & 30.5 & \textbf{67.3} & \textbf{13.8} \\
    \midrule
    ReMixMatch & $X$ & 48.6 & 55.5 & 76.8 & 34.5 & FlexMatch & $X$ & 36.9 & \textbf{43.2} & 54.9 & \textbf{16.7} \\
    ReMixMatch & \textit{OS}  & \textbf{57.2} & 44.5 & 79.9 & 33.1 & FlexMatch &  \textit{OS} & 24.2 & 6.8 & 65.4 & 13.4 \\
    ReMixMatch & \bf SimCore &  56.0 & \textbf{56.4} & \textbf{82.6} & \textbf{36.7} & FlexMatch &  \bf SimCore & \textbf{41.9} & 38.0 & \textbf{71.8} & 10.7 \\
    \bottomrule
    \end{tabular}
    \vspace{-5pt}
    \caption{Fine-tuning performances with various semi-supervised learning methods. We assumed that only 10\% of the data were annotated for semi-supervised fine-tuning.}
    \label{tab:semisup_finetuning}
\end{table}

\vspace{-12pt}
\paragraph{Active learning:}
We conducted additional experiments on various active learning methods for fine-tuning the SSL-pretrained models. Table\,\ref{tab:active_learning_finetuning} presents the experimental results with four standard active selection methods, such as Random, Entropy\cite{confidence_sampling}, Coreset\cite{coreset}, and BADGE\cite{badge}.
We randomly selected the first 10\% of the data and sequentially queried 10\% ratio using each active learning algorithm. After annotating the queried samples, we fine-tuned models for 100 epochs, starting from the checkpoint of the previous active learning round.
As an effective initialization, our SimCore pretraining approach outperformed both pretraining baselines using either $X$ or \textit{OS}, in the active learning fine-tuning schemes as well.

\begin{table}[!h]
    \vspace{5pt}
    \small
    \centering
    \addtolength{\tabcolsep}{-1pt}
    \begin{tabular}{lcccccccccccccccc}
    \toprule
    & \multicolumn{4}{c}{Aircraft} & \multicolumn{4}{c}{Cars}  & \multicolumn{4}{c}{Pet}  & \multicolumn{4}{c}{Birds} \\
    \cmidrule(l{2pt}r{2pt}){2-5} \cmidrule(l{2pt}r{2pt}){6-9} \cmidrule(l{2pt}r{2pt}){10-13} \cmidrule(l{2pt}r{2pt}){14-17}
    method & 10\% & 20\% & 30\% & 40\% & 10\% & 20\% & 30\% & 40\% & 10\% & 20\% & 30\% & 40\% & 10\% & 20\% & 30\% & 40\% \\
    \midrule
    \multicolumn{17}{c}{pretrain: $X$} \\
    \hline
    Random & 28.8 & 45.2 & 50.0 & 55.3 & 26.2 & 52.8 & 63.4 & 71.0 & 47.2 & 56.8 & 59.4 & 64.8 & 13.5 & 24.8 & 31.9 & 41.4 \\
    Entropy & - & 38.9 & 46.2 & 53.8 & - & 47.3 & 62.9 & 72.4 & - & 57.6 & 60.4 & 66.2 & - & 25.5 & 32.6 & 40.5 \\
    Coreset & - & 45.2 & 51.4 & 56.3 & - & 53.3 & 66.3 & 74.3 & - & 57.6 & 62.7 & 66.5 & - & 24.2 & 31.7 & 40.2 \\
    BADGE & - & 45.3 & 51.3 & 56.1 & - & 51.9 & \textbf{66.6} & 73.7 & - & 58.8 & 61.4 & 66.9 & - & 25.7 & 31.9 & 40.4 \\
    \midrule
    \multicolumn{17}{c}{pretrain: \textit{OS}} \\
    \hline
    Random & 20.7 & 36.0 & 45.5 & 54.8 & 10.9 & 32.5 & 47.1 & 59.5 & 35.0 & 62.0 & 68.3 & 72.8 & 11.2 & 20.7 & 20.6 & 30.1 \\
    Entropy & - & 34.1 & 43.3 & 52.0 & - & 29.0 & 45.7 & 58.6 & - & 61.9 & 69.5 & 75.1 & - & 21.8 & 30.3 & 39.3 \\
    Coreset & - & 35.3 & 44.5 & 51.8 & - & 27.2 & 43.4 & 56.1 & - & 60.3 & 71.1 & 75.4 & - & 19.9 & 28.1 & 38.4 \\
    BADGE & - & 35.9 & 46.5 & 53.6 & - & 29.5 & 45.0 & 58.3 & - & 62.4 & 70.4 & 76.0 & - & 20.8 & 30.0 & 39.7 \\
    \midrule
    \multicolumn{17}{c}{pretrain: \textbf{SimCore}} \\
    \hline
    Random & 33.7 & 49.6 & 57.6 & 62.8 & 25.2 & 54.3 & 63.9 & 71.5 & 50.5 & 66.2 & 71.6 & 75.5 & 8.1 & 24.7 & 32.1 & \textbf{43.0} \\
    Entropy & - & 43.4 & 53.4 & 62.7 & - & 49.0 & 64.2 & 73.2 & - & 65.5 & 71.0 & 76.4 & - & \textbf{25.8} & 32.7 & 42.8 \\
    Coreset & - & 50.1 & 57.5 & 63.5 & - & 52.1 & 64.9 & 73.9 & - & 67.7 & 72.9 & \textbf{78.2} & - & 23.5 & 31.3 & 41.0 \\
    BADGE & - & \textbf{50.9} & \textbf{58.7} & \textbf{65.0} & - & \textbf{53.4} & 66.5 & \textbf{74.5} & - & \textbf{70.0} & \textbf{75.2} & 77.8 & - & 24.8 & \textbf{32.9} & 42.4 \\
    \bottomrule
    \end{tabular}
    \vspace{-5pt}
    \caption{Fine-tuning performances with various active learning methods.}
    \label{tab:active_learning_finetuning}
\end{table}

\newpage
\section{Comparisons with Hard Negative Mining}
\label{subsec:comparisons_with_hnm}

SimCore can be thought of as hard sample mining from the open-set. In SSL literature, hard negative mining\,(HNM) is a well-known technique for leveraging the informative sample features. Robinson \etal\cite{robinson2020contrastive} proposed an implicit method for mining the negatives that are similar to an anchor\,($z_i$ in Eq.\,\ref{eq:contrastive}), while Wang \etal\cite{wang2021understanding} suggested explicit sampling to inflict large gradient penalties.

Table\,\ref{tab:hnm} compares SimCore with the existing HNM approaches. Hard negative sampling\,(HNS) and explicit HNS\,(EHNS) slightly improve the performance; however, both still use all open-set samples, some of which may not be relevant to the target. In other words, they employ hard negatives for every open-set anchors, which may reduce the performance due to the distribution mismatch to the target dataset.

SimCore, on the other hand, retrieves only the coreset, reducing the need for an additional negative mining in the loss function.
Nonetheless, SimCore can be applied to any SSL method including HNM-based losses. Thus, we have tried using HNS loss after the explicit coreset sampling, achieving a small gain in the Birds dataset.

\begin{table}[!h]
    \small
    \centering
    \addtolength{\tabcolsep}{-1pt}
    \begin{tabular}{l|lcc|cc}
    \toprule
    method & pretrain data & \!\!\!hard\,(exp.)\!\!\! & hard\,(imp.)\! & Pet & Birds \\
    \midrule
    SimCLR\cite{chen2020simple} & $X$ & \xmark & \xmark & 58.2 & 25.9 \\
    HNS\cite{robinson2020contrastive} & $X$+\textit{OS} & \xmark & \cmark & 63.6 & 30.0 \\
    EHNS\cite{wang2021understanding} & $X$+\textit{OS} & \cmark & \xmark & 62.9 & 27.9 \\
    EHNS-m\cite{wang2021understanding} & $X$+\textit{OS}\,(mem.) & \cmark & \xmark & \textcolor{gray}{53.4} & \textcolor{gray}{18.7} \\
    \textbf{SimCore (ours)} & $X$+{coreset} & \cmark & \xmark & \bf 72.7 & \bf 31.3 \\
    \textbf{SimCore}\,\textcolor[HTML]{6BAEDE}{+\,HNS} & $X$+{coreset} & \cmark & \cmark & 69.3 & \bf \textcolor[HTML]{6BAEDE}{33.1} \\
    \bottomrule
    \end{tabular}
    \vspace{-2pt}
    \caption{Comparisons with hard negative mining techniques. We compared each method by the usage of explicit or implicit hard negative sampling and the pretraining data. ResNet18 is used for the above experiments.}
    \label{tab:hnm}
\end{table}

\vspace{-12pt}
\paragraph{Explicit HNS with memory bank:}
We have compared SimCore with previous hard negative mining techniques: HNS\cite{robinson2020contrastive} and EHNS\cite{wang2021understanding}. However, they only slightly improved the performance because they still use all open-set samples as an anchor. To this end, we modified the sampling strategy of EHNS to better align with our conception.

Specifically, because EHNS is an explicit sampling method based on the similarities with an anchor, we can only use target data samples as the anchor and utilize hard negatives in the open-set. 
There are two ways to make it available:
\vspace{-2pt}
\begin{enumerate}
    \item We can sample a minibatch from $X$+\textit{OS} as before, but only calculates the EHNS loss with the target anchors.
    \vspace{-4pt}
    \item We can sample a minibatch only from $X$. The negative pairs are from other target sample's features and open-set features, where the open-set features are saved in a memory bank\cite{he2020momentum}.
\end{enumerate}
\vspace{-2pt}

\noindent For the first strategy, the size difference between \textit{OS} and $X$ is so large that there are not enough target samples in a minibatch, which prevented the model from being converged. The second strategy is more reasonable in that a minibatch is comprised of only $X$ samples, while the explicit hard negative sampling is done with other target features and open-set features. 

Table\,\ref{tab:hnm} presents the result of EHNS with memory bank (EHNS-m), showing that EHNS-m is not much effective. This is because negative pairs from the memory bank do not impose any gradients--only the target samples are directly involved in learning. To make the pretraining effective, therefore, it is crucial to sample a coreset and employ the entire coreset samples into the training.

\end{document}